\documentclass[runningheads]{llncs}
\usepackage{graphicx}
\usepackage{tikz}
\usepackage{comment} 
\usepackage{amsmath,amssymb} % define this before the line numbering.
\usepackage{color}
\usepackage[caption=false]{subfig}
\usepackage{booktabs}
\usepackage{multirow}
\usepackage[export]{adjustbox}
\usepackage{enumitem}
\usepackage{siunitx}
\usepackage{bm}
\usepackage[width=122mm,left=12mm,paperwidth=146mm,height=193mm,top=12mm,paperheight=217mm]{geometry}

\begin{document}
% \linenumbers
\pagestyle{headings}
\mainmatter
\def\ECCVSubNumber{1097}  % Insert your submission number here

\title{Increasing the Robustness of Semantic Segmentation Models with Painting-by-Numbers} 

\begin{comment}
\titlerunning{ECCV-20 submission ID \ECCVSubNumber} 
\authorrunning{ECCV-20 submission ID \ECCVSubNumber} 
\author{Anonymous ECCV submission}
\institute{Paper ID \ECCVSubNumber}
\end{comment}
%******************

%\begin{comment}
\titlerunning{Increasing Robustness with Painting-by-Numbers}
\author{Christoph Kamann\inst{1,3}
\and
Burkhard G{\"u}ssefeld\inst{1}
\and
Robin Hutmacher\inst{2}
\and
Jan Hendrik Metzen\inst{2}
\and
Carsten Rother\inst{3}
}
\authorrunning{C. Kamann et al.}
\institute{
Corporate Research, Robert Bosch GmbH, Renningen, Germany \\
\email{\{christoph.kamann, burkhard.guessefeld\}@bosch.com}
\and
Bosch Center for Artificial Intelligence, Renningen, Germany \\
\email{\{robin.hutmacher, janhendrik.metzen\}@bosch.com}
\and
Visual Learning Lab, Heidelberg University (HCI/IWR), Heidelberg, Germany \\
\email{carsten.rother@iwr.uni-heidelberg.de} \\
\url{http://vislearn.de}
}
%\end{comment}
%******************
\maketitle

\graphicspath{{imgs/}}
\begin{abstract}
% text after this line
For safety-critical applications such as autonomous driving, CNNs have to be robust with respect to unavoidable image corruptions, such as image noise. 
While previous works addressed the task of robust prediction in the context of full-image classification, we consider it for dense semantic segmentation.
We build upon an insight from image classification that output robustness can be improved by increasing the network-bias towards object shapes. 
We present a new training schema that increases this shape bias. 
Our basic idea is to alpha-blend a portion of the RGB training images with faked images, where each class-label is given a fixed, randomly chosen color that is not likely to appear in real imagery. 
This forces the network to rely more strongly on shape cues. 
We call this data augmentation technique ``Painting-by-Numbers''. 
We demonstrate the effectiveness of our training schema for DeepLabv3$+$ with various network backbones, MobileNet-V2, ResNets, and Xception, and evaluate it on the Cityscapes dataset. 
With respect to our 16 different types of image corruptions and 5 different network backbones, we are in \SI{74}{\%} better than training with clean data. 
For cases where we are worse than a model trained without our training schema, it is mostly only marginally worse. 
However, for some image corruptions such as images with noise, we see a considerable performance gain of up to \SI{25}{\%}.
\\
\\
The final authenticated publication is available online at\\ \url{https://doi.org/10.1007/978-3-030-58607-2_22}
\keywords{Semantic segmentation, shape-bias, corruption robustness}
\end{abstract}

\section{Introduction}
Convolutional Neural Networks (CNNs) have set the state-of-the-art for many computer vision tasks~\cite{krizhevsky_imagenet_2012,he_deep_2016,simonyan_very_2015,szegedy_going_2015,lecun_gradient-based_1998,redmon_you_2016,chen_semantic_2015,goodfellow_deep_2016,he_delving_2015,lecun_deep_2015,noh_learning_2015,long_fully_2015}.
The benchmark datasets which are used to measure performance often consist of clean and undistorted images~\cite{cordts_cityscapes_2016}.
When networks are trained on clean image data and tested on real-world image corruptions, such as image noise or blur, the performance can decrease drastically~\cite{geirhos_generalisation_2018,hendrycks_benchmarking_2019,dodge_understanding_2016,azulay_why_2019,kamann_benchmarking_2020}.

Common image corruptions cannot be avoided in safety-critical applications:
Environmental influences, such as adverse weather conditions, may corrupt the image quality significantly. 
Foggy weather decreases the image contrast, and low-light scenarios may exhibit image noise.
Fast-moving objects or camera motion cause image blur.
Such influences cannot be fully suppressed by sensing technology, and it is hence essential that CNNs are robust against common image corruptions.
Obviously a CNN should also be robust towards adversarial perturbations (e.g.,~\cite{szegedy_intriguing_2013,huang_safety_2017,cisse_parseval_2017,gu_towards_2014,carlini_towards_2017,metzen_detecting_2017,carlini_adversarial_2017}).

% Pull Fig 
\begin{figure}[h]
\centering
    \subfloat[Corrupted validation image]{%
      \includegraphics[width=0.455\linewidth]{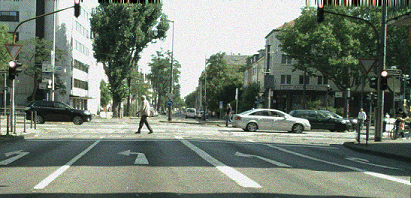}%
    }\hfil
    \subfloat[Ground truth]{%
      \includegraphics[width=0.455\linewidth]{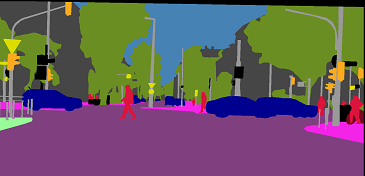}%
    }
    
    \subfloat[Prediction with our training schema]{%
    \includegraphics[width=0.455\linewidth]{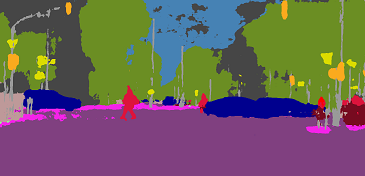}%
    }\hfil
    \subfloat[Prediction with standard training schema]{%
      \includegraphics[width=0.455\linewidth]{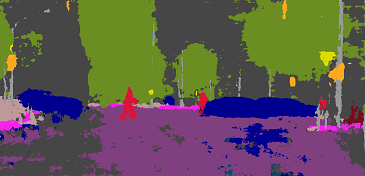}%
    }

    \caption{Results of a semantic segmentation model that is trained with our data augmentation schema. 
    (a) An image crop of the Cityscapes validation set is corrupted by severe image noise. 
    (b) Corresponding ground-truth. 
    (c) Prediction of a model that is trained with our schema. 
    (d) Prediction of the same model with reference training schema, where training images are not augmented with noise.
    The prediction with our training schema (c) is clearly superior to the prediction of the reference training schema (d), though our model is not trained with image noise.
    In particular the classes \textit{road}, \textit{traffic signs}, \textit{cars}, \textit{persons} and \textit{poles}, are more accurately predicted
}
    \label{fig:fig1}
\end{figure}
Training CNNs directly on image corruptions is generally a possibility to increase the performance on the respective type of image corruption, however, this approach comes at the cost of increased training time.
It is also possible that a CNN overfits to a specific type of image corruption trained on~\cite{geirhos_generalisation_2018,vasiljevic_examining_2016}.

Recent work deals with the robustness against common image corruptions for the task of full-image classification, and less effort has been dedicated to semantic segmentation.
Whereas other work utilizes, e.g., a set of data augmentation operations~\cite{hendrycks_augmix:_2019} we propose a new, robustness increasing, data augmentation schema (see Fig.~\ref{fig:fig1}) that does: a) not require any additional image data, and b) is easy to implement and c) can be used within any supervised semantic segmentation network, and d) is robust against many common image corruptions.

\begin{figure}[h]
\centering
    \subfloat[Original data]{%
      \includegraphics[width=0.24\linewidth]{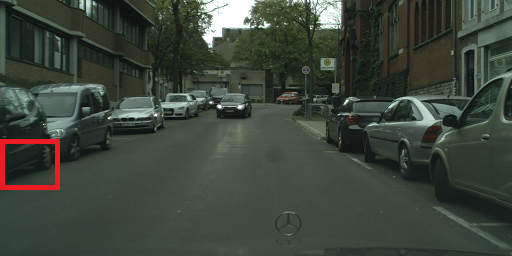}%
    }\hfil
    \subfloat[Stylized data]{%
      \includegraphics[width=0.24\linewidth]{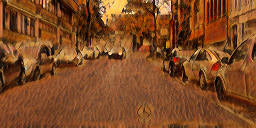}%
    }\hfil
    \subfloat[Zoom of (a)]{%
    \includegraphics[width=0.24\linewidth]{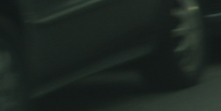}%
    }\hfil
    \subfloat[Zoom of (b)]{%
      \includegraphics[width=0.24\linewidth]{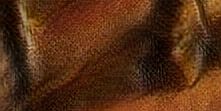}%
    }
\caption{
Illustration of the style transfer technique of~\cite{geirhos_imagenet-trained_2019}.
An original training image (a) of the Cityscapes dataset is stylized by a painting (b).
(c) and (d) show the image content of the red rectangle of (a), where (d) is clearly noisier compared to the original data (c)
}
    \label{fig:stylized}
\end{figure}

For this, we build upon the work of Geirhos et al.~\cite{geirhos_imagenet-trained_2019}, where it has been shown that increasing the network bias towards the shape of objects does make the task of full-image classification more robust with respect to common image corruptions.
We applied the style-transfer technique of Geirhos et al. to Cityscapes, but found the resulting images to be quite noisy (see Fig.~\ref{fig:stylized}). 
Training on such data might, therefore, increase robustness not solely due to an increased shape bias, but rather due to increased image corruption. 
Our aim is to find a training schema that does not have any type of image corruption added. 

Whereas the method of~\cite{geirhos_imagenet-trained_2019} delivers high-quality results, their approach requires a computationally intensive style transfer technique and additional image data. 
We propose a simple, yet effective, data augmentation scheme, that decreases the amount of texture in the training data, and does not need additional data. 
The basic idea is to alpha-blend some training images with a texture-free representation, as illustrated in Fig.~\ref{fig:paintingbynumbers} (b).
By doing so, the texture-based appearance of a training image is less reliable, forcing the network to develop additional shape-based cues for the segmentation.
In this way, our schema does not require additional training data, as we directly use the available semantic segmentation ground-truth.
It can be easily utilized for training any supervised semantic segmentation model.
We demonstrate our data augmentation scheme's effectiveness on a broad range of common image corruptions, evaluated on the Cityscapes dataset.

% Summary of contributions
In summary, we give the following contributions:
\begin{itemize}[noitemsep,nolistsep]
    \item[--] We propose a simple, yet effective, data augmentation scheme that increases the robustness of well-established semantic segmentation models for a broad range of image corruptions, through increasing the model's shape-bias. 
    Our new training schema requires no additional data, can be utilized in any supervised semantic segmentation model, and is computationally efficient. 
    \item[--] 
    We validate our training schema through a series of validation experiments. 
    With respect to our 16 different types of image corruptions and five different network backbones, we are in \SI{74}{\%} better than training with clean data. 
    We are able to increase the mean IoU by up to \SI{25}{\%}.
\end{itemize}

\section{Related Work}
Recent work has dealt with the robustness of CNNs for common image corruptions. 
We discuss the most recent work in the following.

\textbf{Benchmarking robustness with respect to common corruptions.}
The work in~\cite{azulay_why_2019,engstrom_exploring_2019} demonstrates that shifting input pixels can change the outcome significantly.
Dodge and Karam~\cite{dodge_understanding_2016} show that CNNs are prone to common corruptions, such as blur, noise, and contrast variations, for the task of full-image classification. 
The authors further show in~\cite{dodge_study_2017} that the CNN performance of classifying corrupted images is significantly lower than human performance. 
Zhou~\cite{zhou_classification_2017} et al. find similar results.
Geirhos et al.~\cite{geirhos_generalisation_2018} show that established models~\cite{szegedy_going_2015,he_deep_2016,simonyan_very_2015} for image classification trained on one type of image noise can struggle to generalize well to other noise types.
Vasiljevic et al.~\cite{vasiljevic_examining_2016} find a similar result w.r.t image blur, and further, a reduced performance for clean data.

Hendrycks and Dietterich~\cite{hendrycks_benchmarking_2019} corrupt the ImageNet dataset~\cite{deng_imagenet:_2009} by many common image corruptions and image perturbations. 
In this work, we apply the proposed image corruptions to the Cityscapes dataset. 
Michaelis et al.~\cite{michaelis_benchmarking_2019} benchmark the robustness in object detection and find a significant performance drop for corrupted input data.

For the task of semantic segmentation, Vasiljevic et al.~\cite{vasiljevic_examining_2016} find that model performance of a VGG-16~\cite{simonyan_very_2015} is decreasing for an increasing amount of blur in the test data.
Kamann and Rother~\cite{kamann_benchmarking_2020,kamann2020benchmarking} ablate the state-of-the-art semantic segmentation DeepLabv3$+$ architecture and show that established architectural design choices affect model robustness with respect to common image corruptions.
Other work deals with robustness towards adverse weather conditions~\cite{sakaridis_guided_2019,sakaridis_semantic_2018,volk_towards_2019}, night scenes~\cite{dai_dark_2018}, or geometric transformations~\cite{fawzi_manitest:_2015,ruderman_pooling_2018}.

\textbf{Increasing robustness with respect to common corruptions.}
The research interest in increasing the robustness of CNN models with respect to common image corruptions grows.
Most methods have been proposed for the task of full-image classification.
Mahajan et al.~\cite{mahajan_exploring_2018} and Xie et al.~\cite{xie_self-training_2019} show that using more training data increases the robustness.
The same result is found when more complex network backbones are used, also for object detection~\cite{michaelis_benchmarking_2019} and semantic segmentation~\cite{kamann_benchmarking_2020}.
Hendrycks et al.~\cite{hendrycks_benchmarking_2019} show that adversarial logit pairing~\cite{kannan_adversarial_2018} increases the robustness for adversarial and common perturbations.
The authors of~\cite{zheng_improving_2016,laermann_achieving_2019} increase model robustness through stability training methods.

Several other works apply data augmentation techniques to increase generalization performance.
Whereas some work 
occludes parts of images~\cite{zhong_random_2017,devries_improved_2017}, 
crops, replaces and mixes several images~\cite{yun_cutmix:_2019,zhang_mixup:_2017,takahashi_data_2019},
or applies various (learned) sets of distortions~\cite{hendrycks_augmix:_2019,cubuk_autoaugment:_2019}, other methods augment with artificial noise to increase robustness~\cite{gilmer_adversarial_2019,lopes_improving_2019,rusak_increasing_2020}.

Geirhos et al.~\cite{geirhos_imagenet-trained_2019} demonstrate that classifiers trained on ImageNet tend to classify images based on an image's texture.
They further show that increasing the shape-bias of a classifier (through style transfer~\cite{gatys_neural_2015}), also increases the robustness for common image corruptions.
This work builds upon this finding to increase the shape-bias of semantic segmentation models and, thus, the robustness for common image corruptions.

\section{Training Schema: Painting-by-Numbers}
\label{sec:method}
\begin{figure}[htbp]
    \centering
    \includegraphics[width=0.8\textwidth]{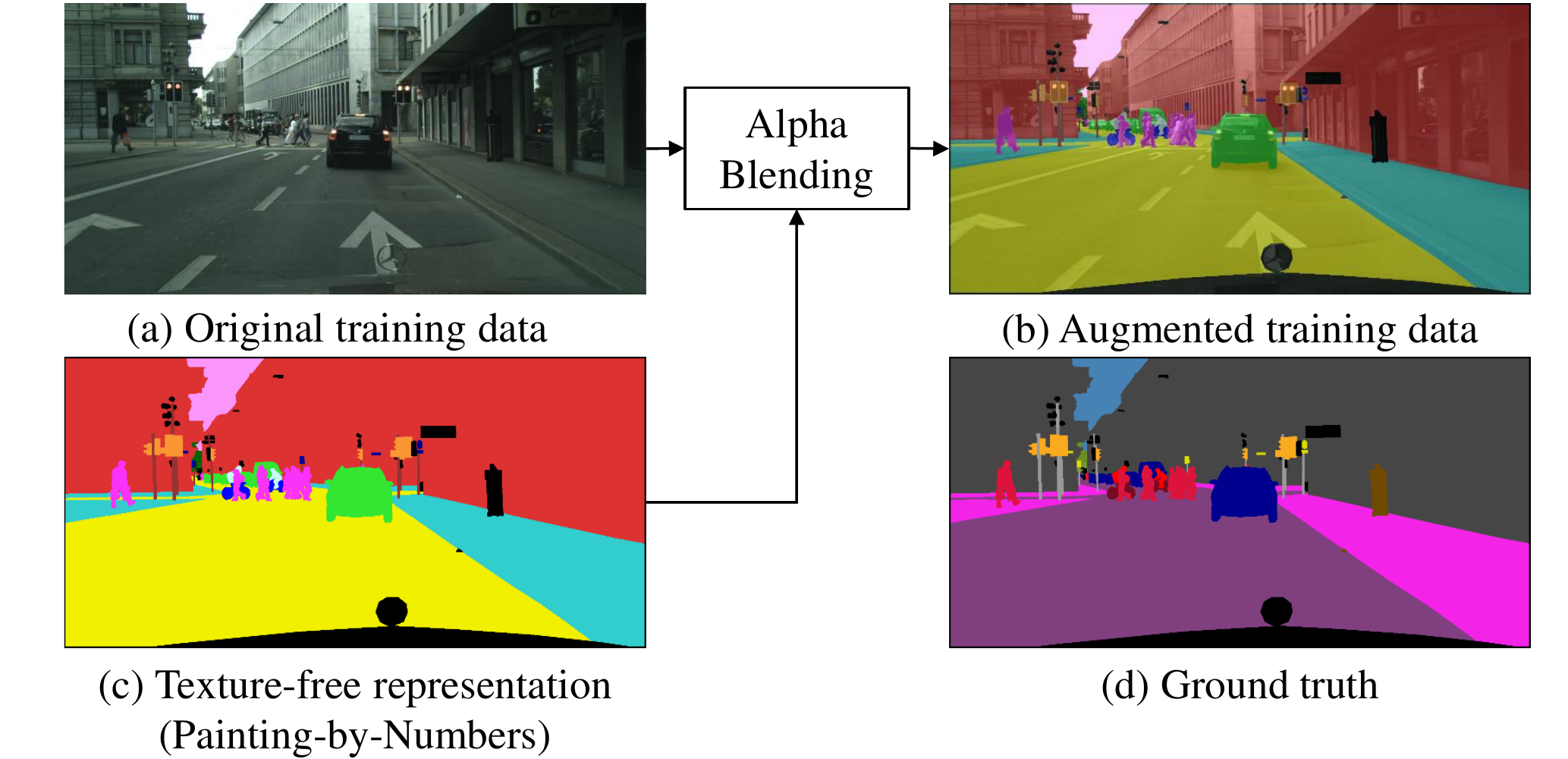}
    \caption{
    Overview of our proposed training schema, which we refer to as Painting-by-Numbers. 
    (a) An RGB image of the Cityscapes training set and the respective ground-truth in (d). 
    We paint the \textit{numbers}, i.e., ground-truth IDs of (d) randomly, leading to the texture-free representation shown in (c).
    Painting the numbers randomly is essential since these colors are not likely to appear in real imagery.
    The final training image (b) is then generated by alpha-blending (a) and (c).
    A fraction of training data is augmented as in b), which is used as training data that increases the robustness against common corruptions
    }
    \label{fig:paintingbynumbers}
\end{figure}

Our goal is to generically increase the robustness of semantic segmentation models for common image corruptions.
Here, \textit{robustness} refers to training a model on clean data and subsequently validating it on corrupted data.
Simply adding corrupted data to the training set does certainly increase the robustness against common corruptions. 
However, this approach comes along with drawbacks: Firstly, a significantly increased training time. Secondly, the possibility to overfit to specific image corruptions~\cite{geirhos_generalisation_2018,vasiljevic_examining_2016} and reduced performance on clean data~\cite{vasiljevic_examining_2016}.
Thirdly, it further may be hard to actually identify all sources of corruption for new test scenarios.
For our training schema, we build on the finding of~\cite{geirhos_imagenet-trained_2019}. 
We propose an augmentation schema (Painting-by-Numbers) that modifies the training process so that the model develops shape-based cues for the decision of how to segment a pixel, resulting in a generic increase of model robustness.

The basis of our schema is that we treat the segmentation ground-truth as a texture-free representation of the original training data (see Fig.~\ref{fig:paintingbynumbers}).
We then colorize (or \textit{paint}) the ground-truth labels (or \textit{numbers}) randomly (Painting-by-Numbers) to generate a representation as shown in Fig.~\ref{fig:paintingbynumbers} (c).
We uniformly sample the color from the sRGB color gamut with range $[0, 255]$, similar to the images of the Cityscapes dataset.
Painting the numbers randomly is essential since these colors are not likely to appear in real imagery.
Finally, we alpha-blend this this representation with the original training image, according to eq.~\ref{eq:alphablend},

\begin{equation}
I_{\mathit{blended}} = \alpha \times I_{\mathit{painting-by-numbers}} + (1 - \alpha) \times I_{\mathit{original}}
\label{eq:alphablend}
\end{equation}

where $I_{\mathit{blended}}$ is the resulting alpha-blended training image (Fig.~\ref{fig:paintingbynumbers} b), $\alpha$ is the blend parameter (where $\alpha=1$ corresponds to a representation where original training input is entirely blended), $I_{\mathit{painting-by-numbers}}$ is the texture-free representation (Fig.~\ref{fig:paintingbynumbers} c) and $I_{\mathit{original}}$ is the original training image (Fig.~\ref{fig:paintingbynumbers} a).
Training a network on such data forces the network to develop (or increase) its shape-bias since we actively corrupt the textural content of the image.
The texture features of the image are, therefore, less reliable, and a model needs to develop additional cues to segment pixels correctly.
Painting-by-Numbers is computationally efficient. 
For our setup, the training time increases by only~\SI{2.5}{\%} (please see the supplementary material for more details).

\textbf{Motivation blending with $ \bm{0 < \alpha < 1}$.}
We conducted the following analysis.
We trained a model solely on texture-free images, as shown in Fig.~\ref{fig:paintingbynumbers} c, meaning that the blend parameter $\alpha$ is fixed to $1$.
This network achieved a decent performance when tested on a texture-free variant of the Cityscapes validation set.
This is a positive signal because it means that the model is able to learn from entirely texture-free training data.

When we augmented only half of a training batch, instead of every image, ($\alpha$ is still fixed to $1$), the performance on both, the original validation set and texture-free validation set was, again, considerably high; However, the robustness of the new model with respect to common image corruptions was not increased.
We hypothesize that such a model learns to predict well for two different domains, which are the original data and the texture-free data.
This motivates us to choose $\alpha < 1$ for some training images.
As we will see, with a varying degree of alpha-blending, the robustness of the model towards common image corruptions increases significantly and, at the same time, keeps a consistently good performance on clean data.

\textbf{Training protocol.} 
We use the state-of-the-art DeepLabv3$+$~\cite{chen_deeplab:_2017,chen_encoder-decoder_2018,chen_rethinking_2017,chen_semantic_2015} semantic segmentation architecture as baseline model.
We show the effectiveness of Painting-by-Numbers for many network backbones: 
MobileNet-V2~\cite{sandler_mobilenetv2:_2018}, ResNet-50~\cite{he_deep_2016}, ResNet-101, Xception-41, Xception-71~\cite{chollet_xception:_2017}.
We augment exactly half of a batch by our Painting-by-Numbers approach and leave the remaining images unchanged. 
Doing so ensures that the performance on clean data is comparable to a network that is trained regularly on clean data only.
We kindly refer to the next section for reasonable choices of the hyperparameters.
We apply a similar training protocol as in~\cite{chen_encoder-decoder_2018}: 
crop size $513 \times 513$\footnote{Due to hardware limitations we are not able to train on the suggested crop size of 769.}, initial learning rate $0.01$, ``poly''~\cite{liu_parsenet:_2015} learning rate schedule, using the Atrous Spatial Pyramid Pooling (ASPP) module~\cite{chen_semantic_2015,grauman_pyramid_2005,he_spatial_2014,lazebnik_beyond_2006,zhao_pyramid_2017}, fine-tuning batch normalization~\cite{ioffe_sergey_batch_2015} parameters, output stride $16$, random scale data augmentation and random flipping during training.
As suggested by~\cite{chen_encoder-decoder_2018}, we apply no global average pooling~\cite{lin_network_2014}. 
We train every model using TensorFlow~\cite{abadi_tensorflow:_2016}.

\textbf{Evaluation protocol.} 
We use the image transformations provided by the ImageNet-C~\cite{hendrycks_benchmarking_2019} dataset to generate Cityscapes-C, similar to~\cite{kamann_benchmarking_2020}.
The ImageNet-C corruptions give a huge selection of transformations.
They consist of several types of blur (Gaussian, motion, defocus, frosted glass), image noise (Gaussian, impulse, shot, speckle), weather (snow, spatter, fog, frost), and digital transformations (JPEG, brightness, contrast).
Please see the supplementary material for examples.
Each corruption type (e.g., Gaussian noise) is parameterized in five severity levels. 
We evaluate the mean-IoU~\cite{everingham_pascal_2010} of many variants of the Cityscapes validation set, which is corrupted by the ImageNet-C transformations.

\section{Experimental Evaluation and Validation}
\label{sec:experiments}
In this section, we demonstrate the effectiveness of Painting-by-Numbers.
In section~\ref{sec:ablationstudies} we discuss implementation details.
We then show the results w.r.t the Cityscapes dataset in section~\ref{sec:results_cityscapes}.
We conduct a series of experiments to validate the increased shape-bias of a model trained with Painting-by-Numbers in section~\ref{sec:validation}.

\subsection{Implementation Details}
\label{sec:ablationstudies}
We experiment with varying implementations and augmentation schemes, which we discuss next.

\textbf{Parameters for alpha-blending.}
Our experiments show that a fixed value for $\alpha$ does not yield the best results.
Instead, we use two parameters for alpha-blending, $\alpha_{min}$ and $\alpha_{max}$.
These values define an interval from which $\alpha$ is drawn. 
They are the essential hyperparameters needed to achieve the best results towards common image corruptions.
If $\alpha_{min}$ is too low, i.e., the amount of texture in the image is high, the robustness increase for common corruptions is minor.
If $\alpha_{min}$ is too high, i.e., the amount of texture in the image is further diminished, the robustness decreases with respect to common corruptions (as discussed previously).
We observe that the models only connect learned features from the two domains (original data domain and alpha-blended data domain) if the latter's texture is present, i.e., $0 < \alpha < 1$. 

\textbf{Batch augmentation schemes.}
We always augment exactly the half of a batch by Painting-By-Numbers for each iteration of the forward path.
To summarize, the only parameters to be optimized are $\alpha_{min}$ and $\alpha_{max}$.
We do not observe better results when for every image in the mini-batch is individually decided if it shall be augmented by Painting-by-Numbers.

\begin{figure}
\centering
\subfloat[Instances are painted randomly]{%
  \includegraphics[width=0.3\linewidth]{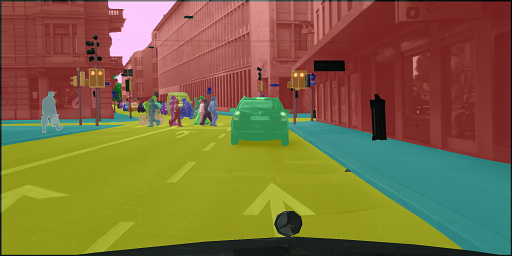}%
}\hfil
\subfloat[Instances are painted by RGB-mean]{%
  \includegraphics[width=0.3\linewidth]{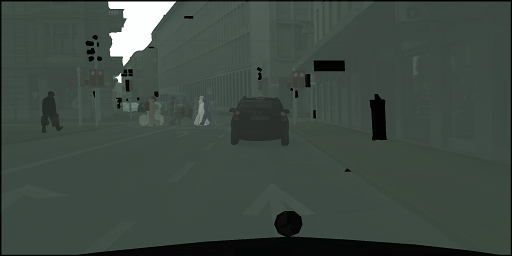}%
}\hfil
\subfloat[Instances are painted by RGB-median]{%
\includegraphics[width=0.3\linewidth]{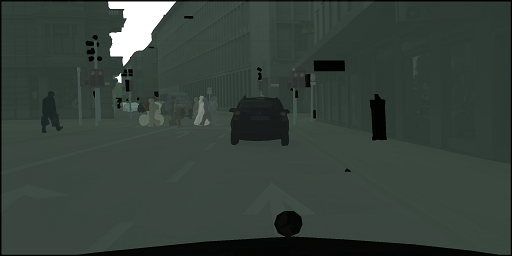}%
}\hfil
\caption{
Examples of several coloring schemes used for Painting-by-Numbers}
\label{fig:instancelabels}
\end{figure}

\textbf{Incorporating instance labels.}
Beside semantic segmentation ground-truth, the Cityscapes dataset also contains instance labels for several classes.
We additionally utilize them in our augmentation scheme to paint each instance with a randomly chosen color (instead of painting each instance of a class with the same color), as illustrated in Fig.~\ref{fig:instancelabels} (a).
This produces promising results with respect to further increasing network robustness.
Since Painting-by-Numbers is targeted for semantic segmentation task, we base our schema on the more general semantic labels, which are available for all reference datasets.

\textbf{Paint with mean and median RGB.}
We further paint the images with a more consistent color, such as the mean and median RGB value of the class or instance (instead of painting the semantic classes randomly), as illustrated in Fig.~\ref{fig:instancelabels} (b) and (c).
This approach does, as expected, not increase the model robustness.
Instead of forcing a model to not rely on texture and color appearance, by corrupting these very properties, the network learns to assign a mean or median value to classes and instances, contrary to the effect of random painting. 
Hence, there is no need to increase the shape-bias for predicting the segmentation map when the colors are likely to appear in real imagery.

\textbf{Best Setup.}
We train MobileNet-V2, ResNet-50, ResNet-101, Xception-41, and Xception-71 with Painting-by-Numbers.
We evaluate the models on Gaussian noise to select the final values for $\alpha$.
For ResNet-50, and Xception-41, we observe the best results when we draw $\alpha$ uniformly from the interval $\alpha_{min}=0.70$ and $\alpha_{max}=0.99$.
For the remaining networks, we observe the best results for $\alpha_{min}=0.50$ and $\alpha_{max}=0.99$.

\subsection{Results on Cityscapes}
\label{sec:results_cityscapes}
\begin{figure}[ht!]
\centering
\resizebox{\textwidth}{!}{
\begin{tabular}{c c c c }
\centering
{ (a) Speckle Noise } & { (b) Shot Noise} & {(c) Contrast }& {(d) Spatter}\\ [-2ex]
\subfloat{
\rotatebox[origin=b]{90}{RGB}
  \includegraphics[valign=m,width=0.225\linewidth]{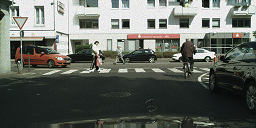}
}
&
\subfloat{
  \includegraphics[valign=m,width=0.225\linewidth]{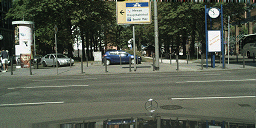}
} 
&
\subfloat{
 \includegraphics[valign=m,width=0.225\linewidth]{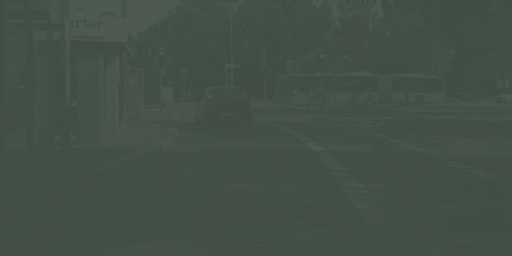}
} 
&
\subfloat{
  \includegraphics[valign=m,width=0.225\linewidth]{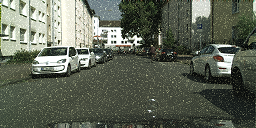} 
} 
\\ [-2ex]
\subfloat{
  \rotatebox[origin=b]{90}{GT}
  \includegraphics[valign=m,width=0.225\linewidth]{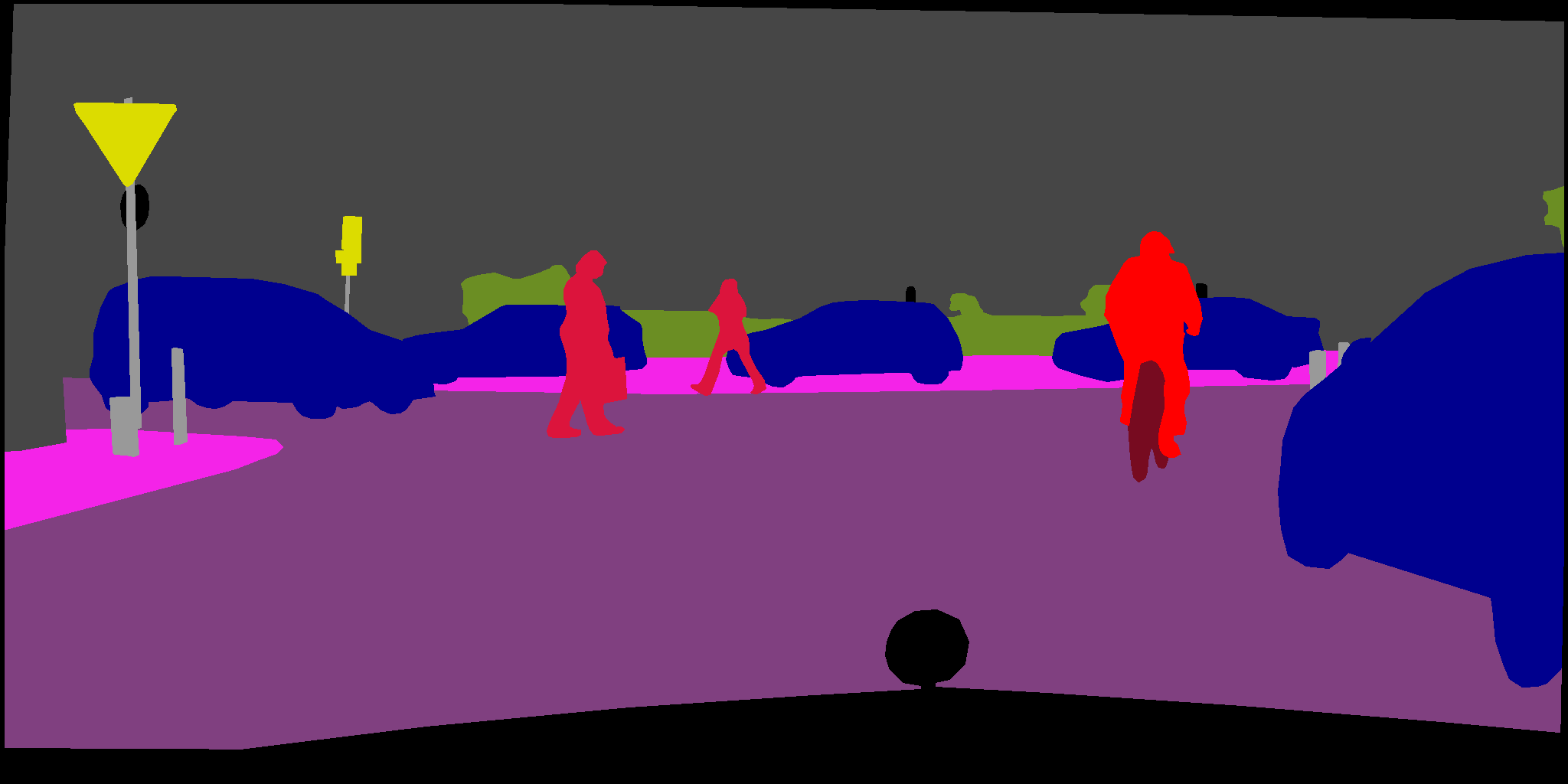}
} 
&
\subfloat{
  \includegraphics[valign=m,width=0.225\linewidth]{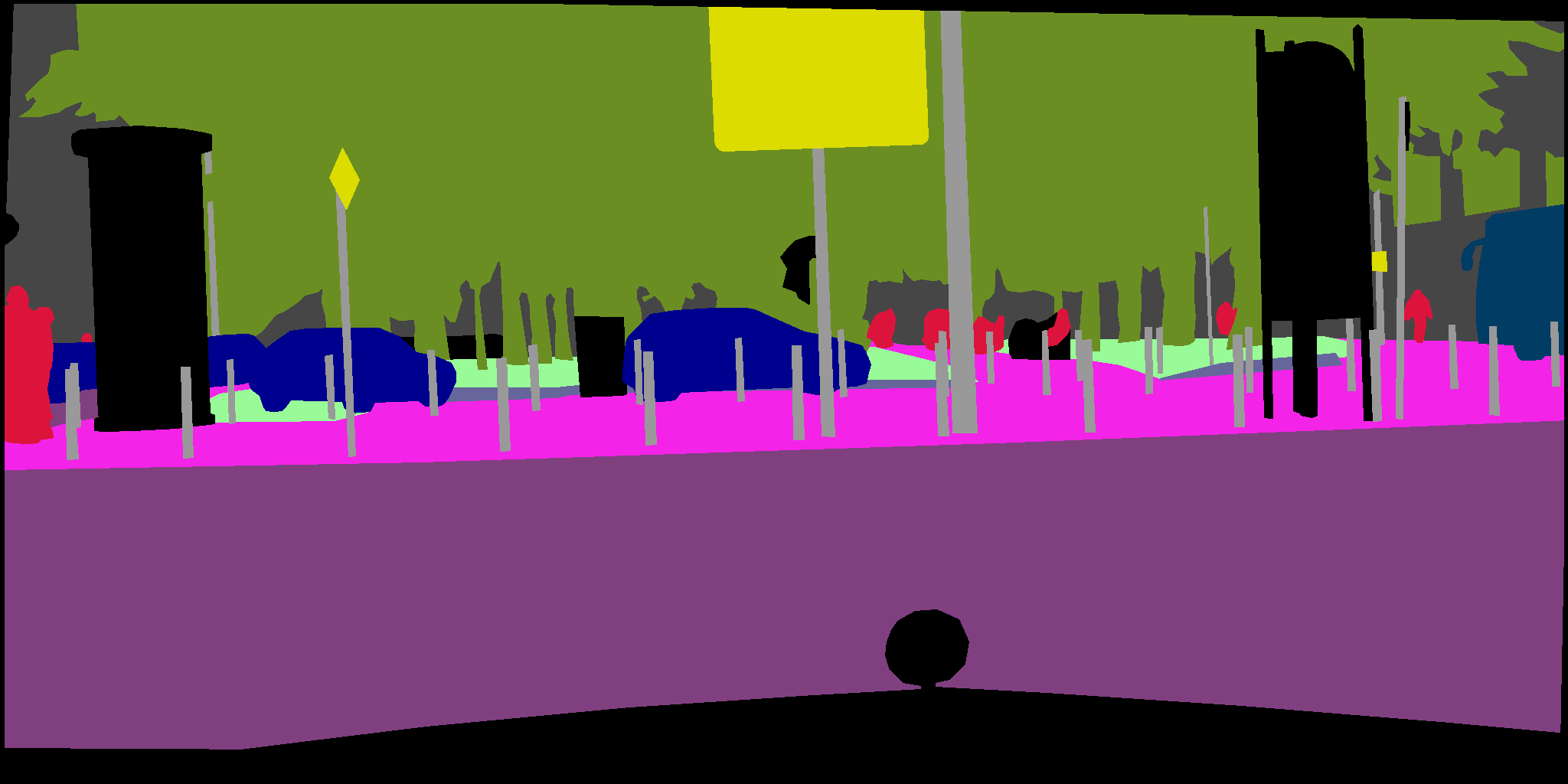}
} 
&
\subfloat{
  \includegraphics[valign=m,width=0.225\linewidth]{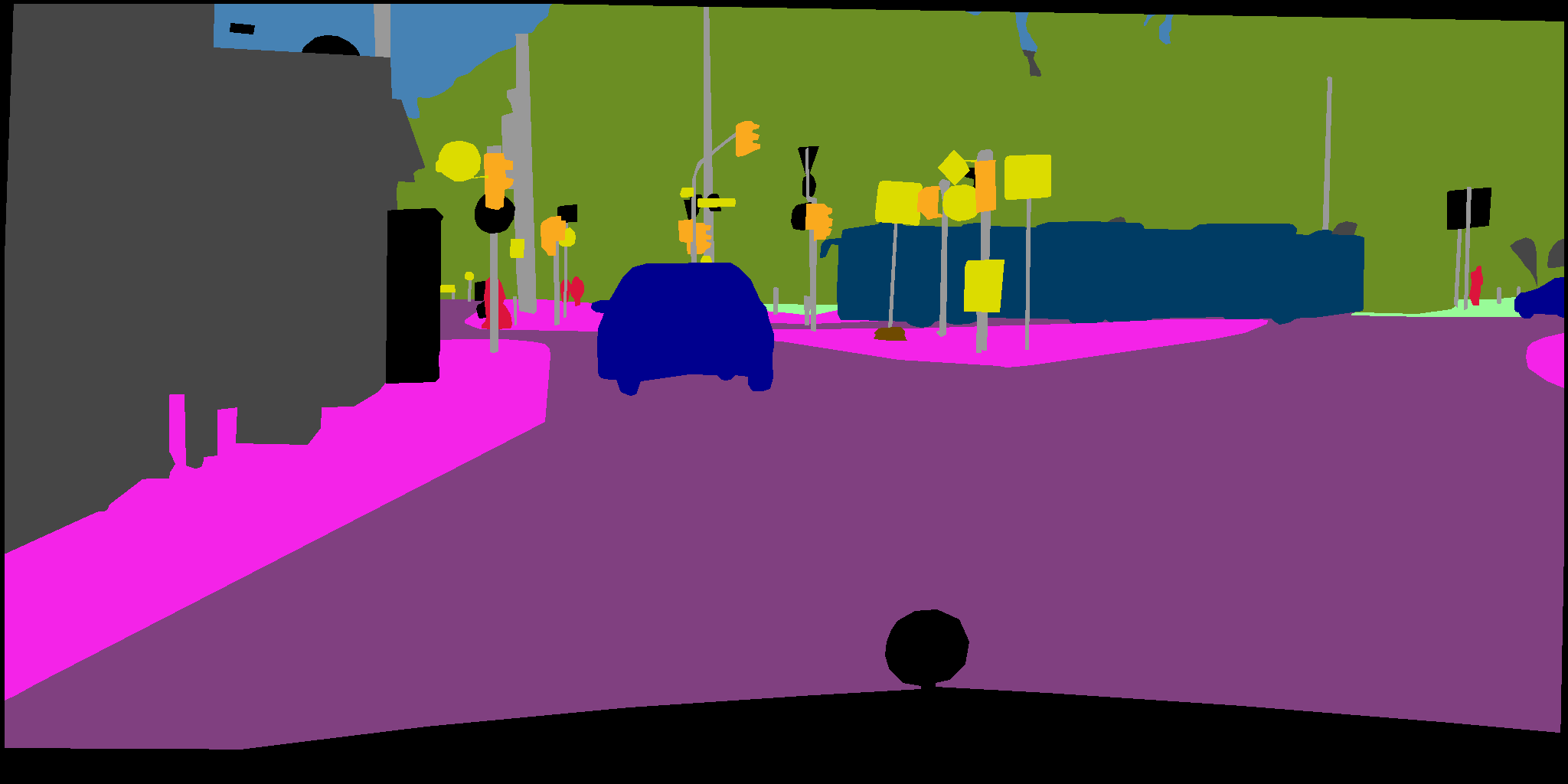}
} 
&
\subfloat{
  \includegraphics[valign=m,width=0.225\linewidth]{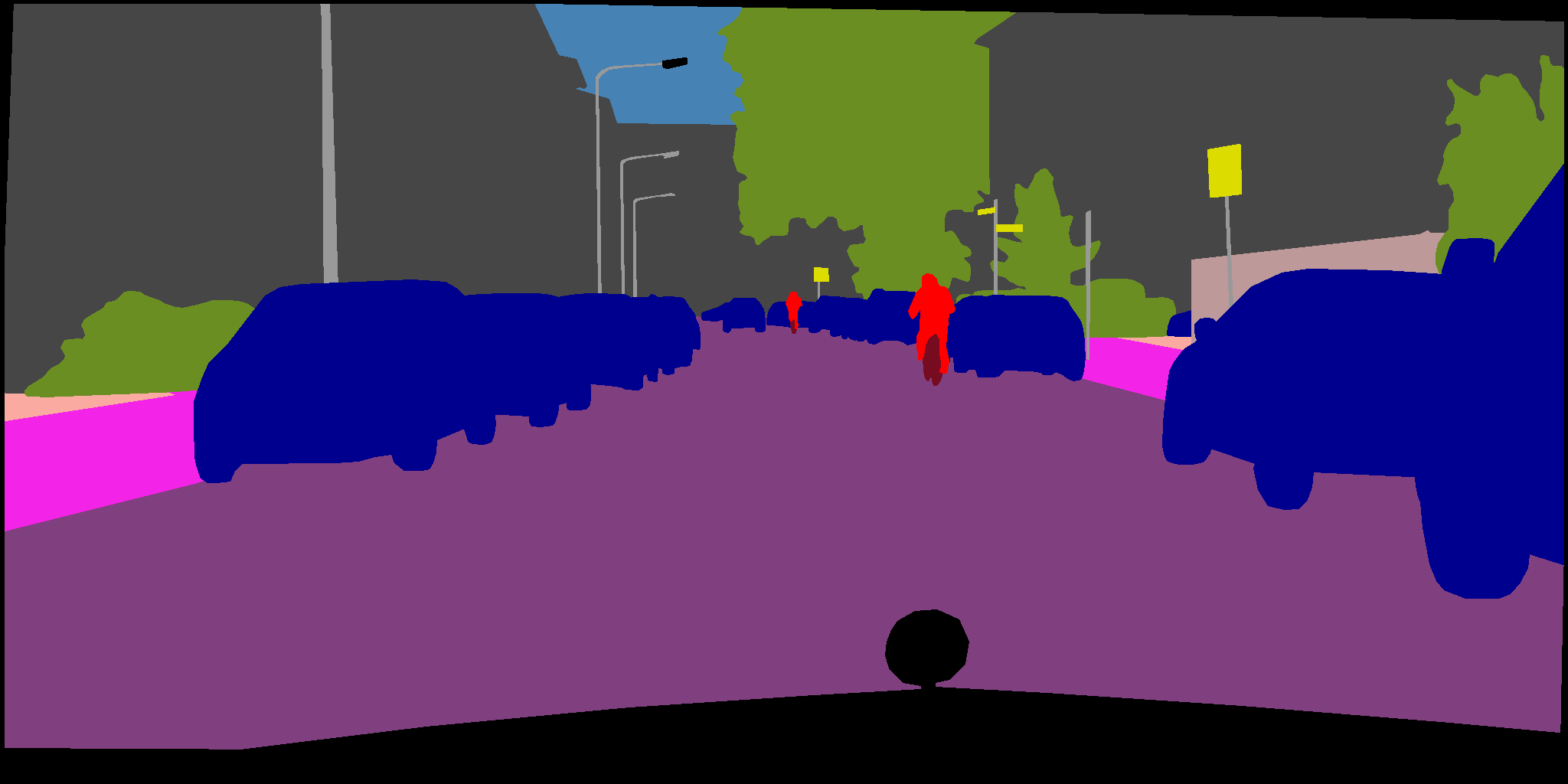}
} 
\\ [-2ex]
\subfloat{
   \rotatebox[origin=b]{90}{Reference}
  \includegraphics[valign=m,width=0.225\linewidth]{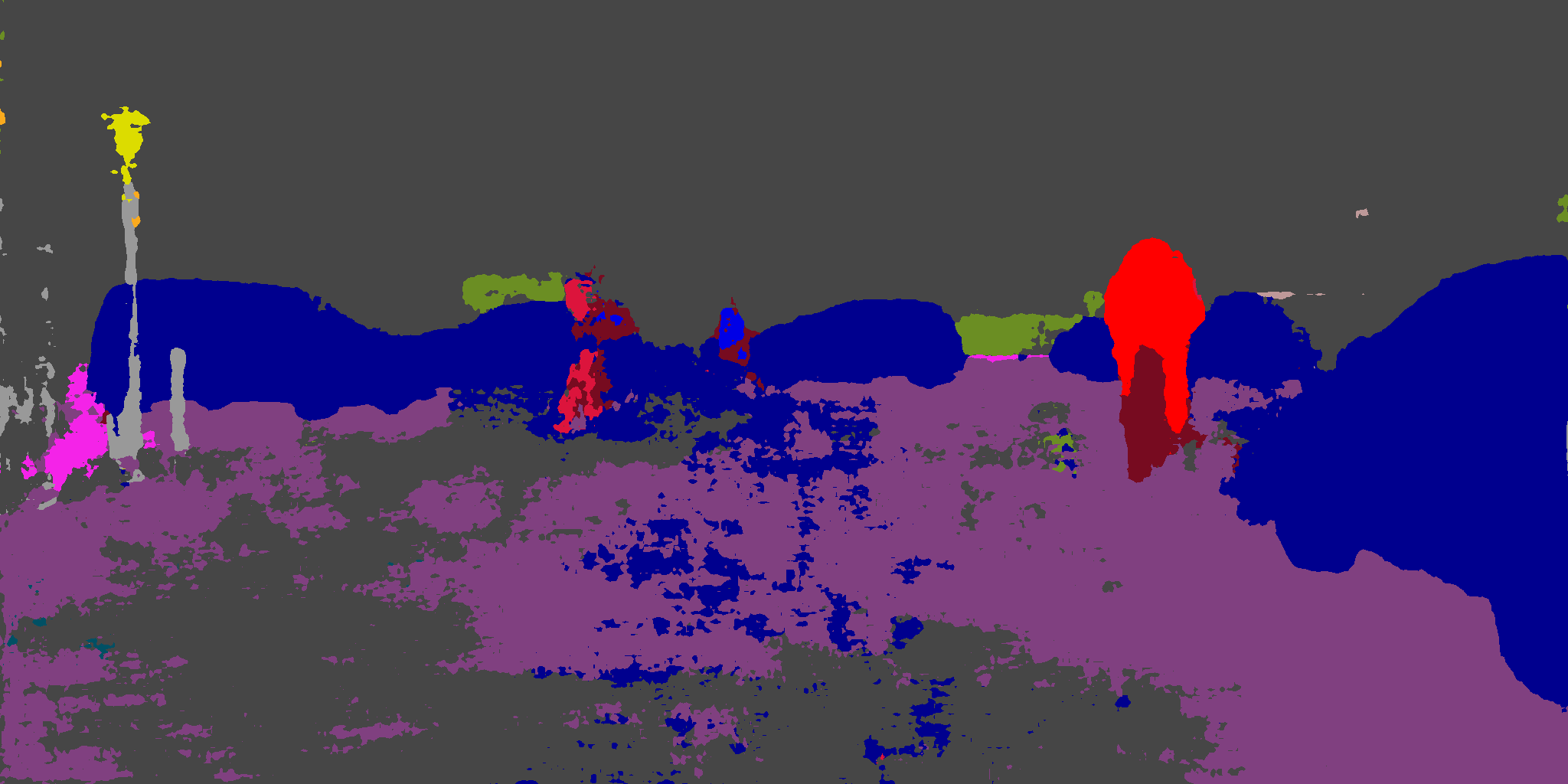}
} 
&
\subfloat{
  \includegraphics[valign=m,width=0.225\linewidth]{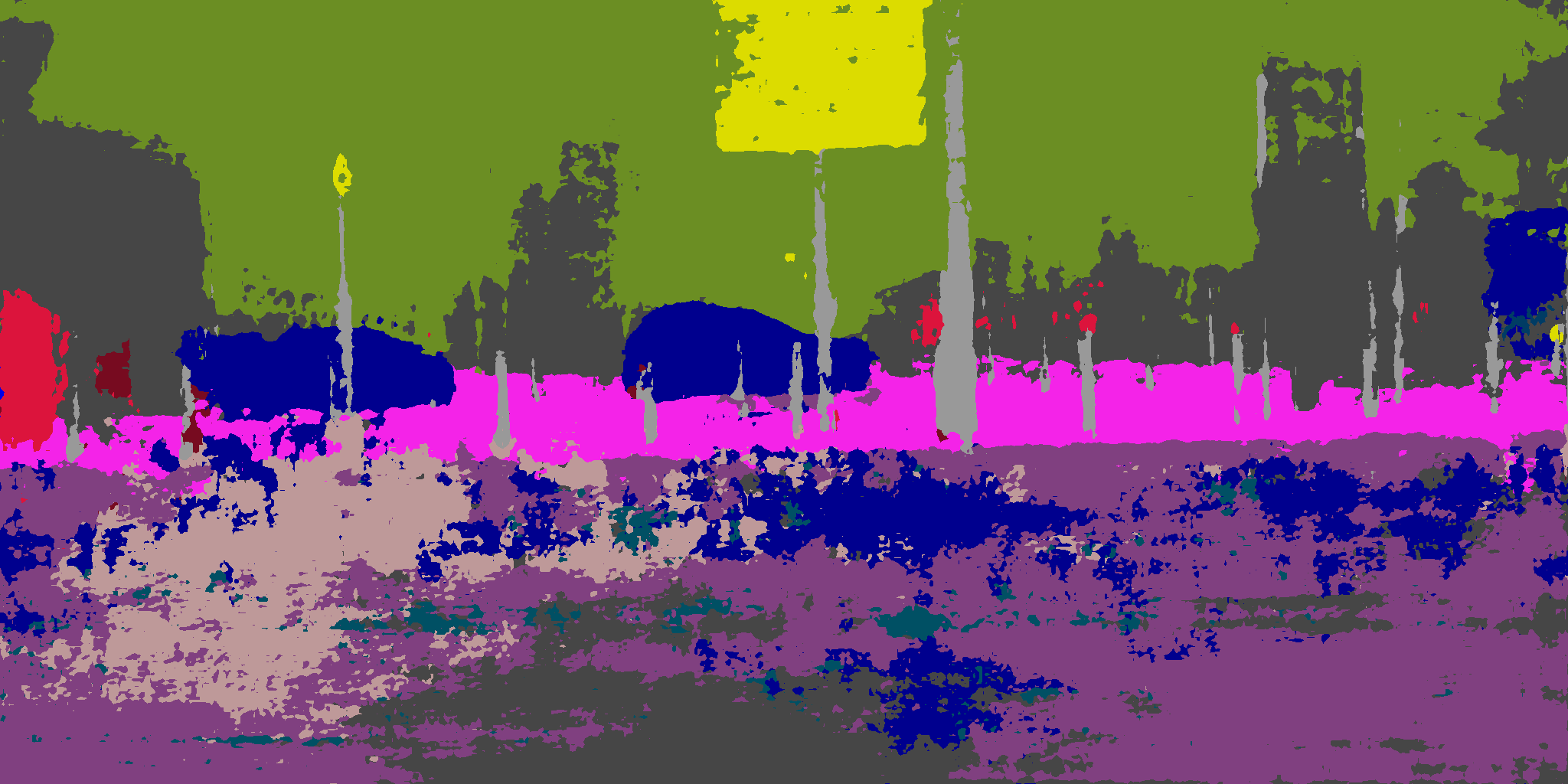}
} 
&
\subfloat{
  \includegraphics[valign=m,width=0.225\linewidth]{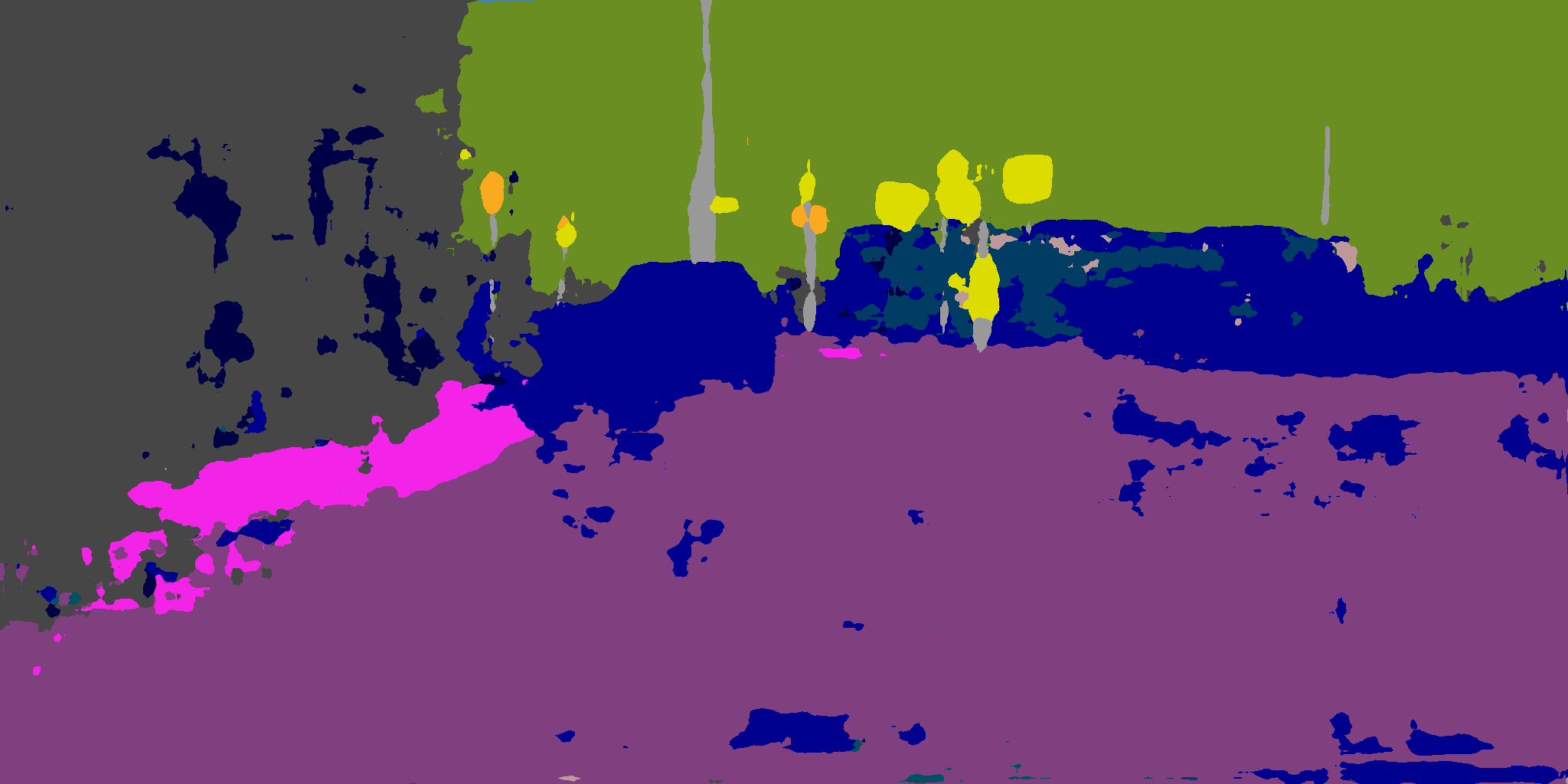}
} 
&
\subfloat{
  \includegraphics[valign=m,width=0.225\linewidth]{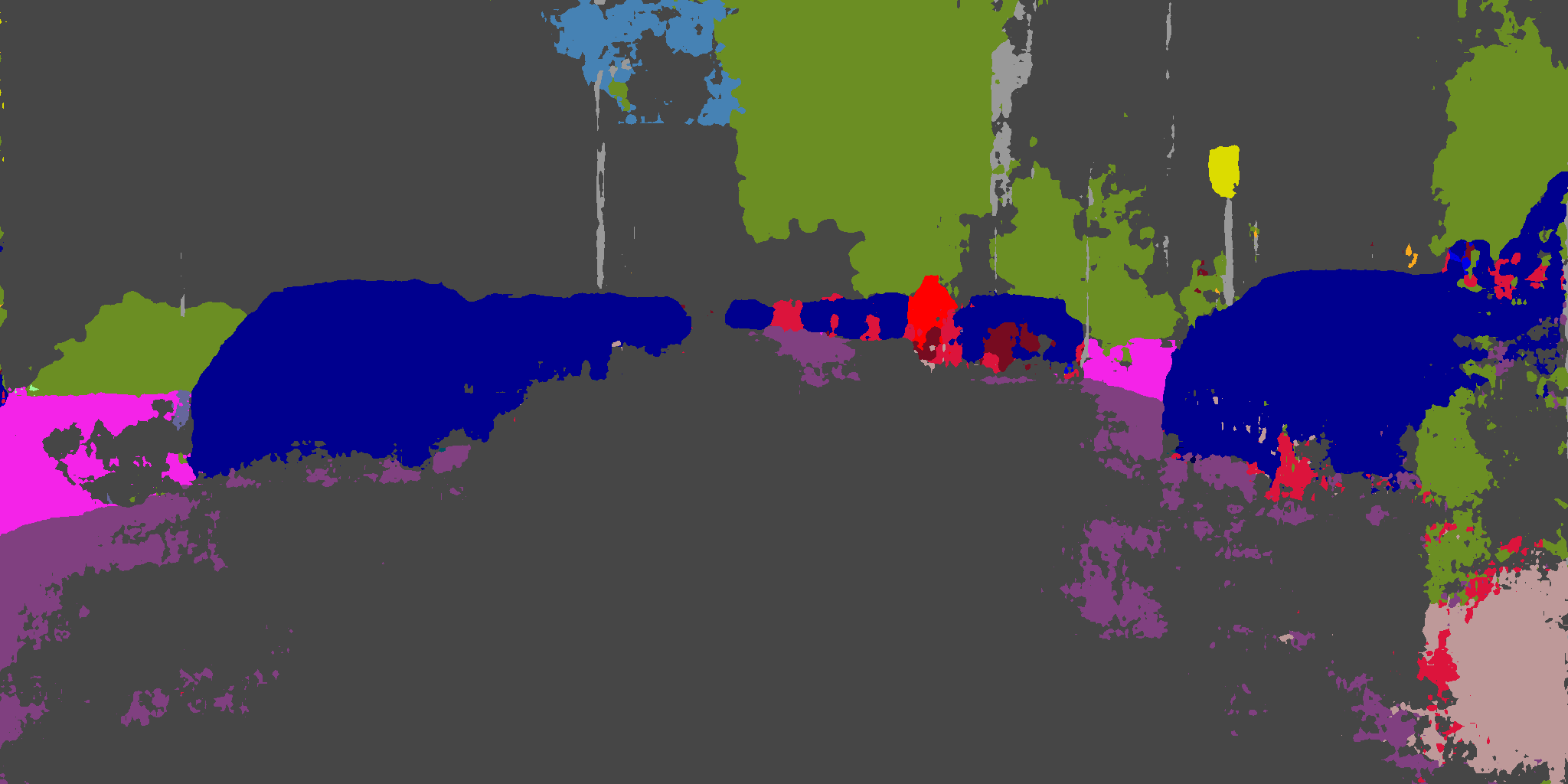}
} 
\\ [-2ex]
\subfloat{
   \rotatebox[origin=b]{90}{Ours}
  \includegraphics[valign=m,width=0.225\linewidth]{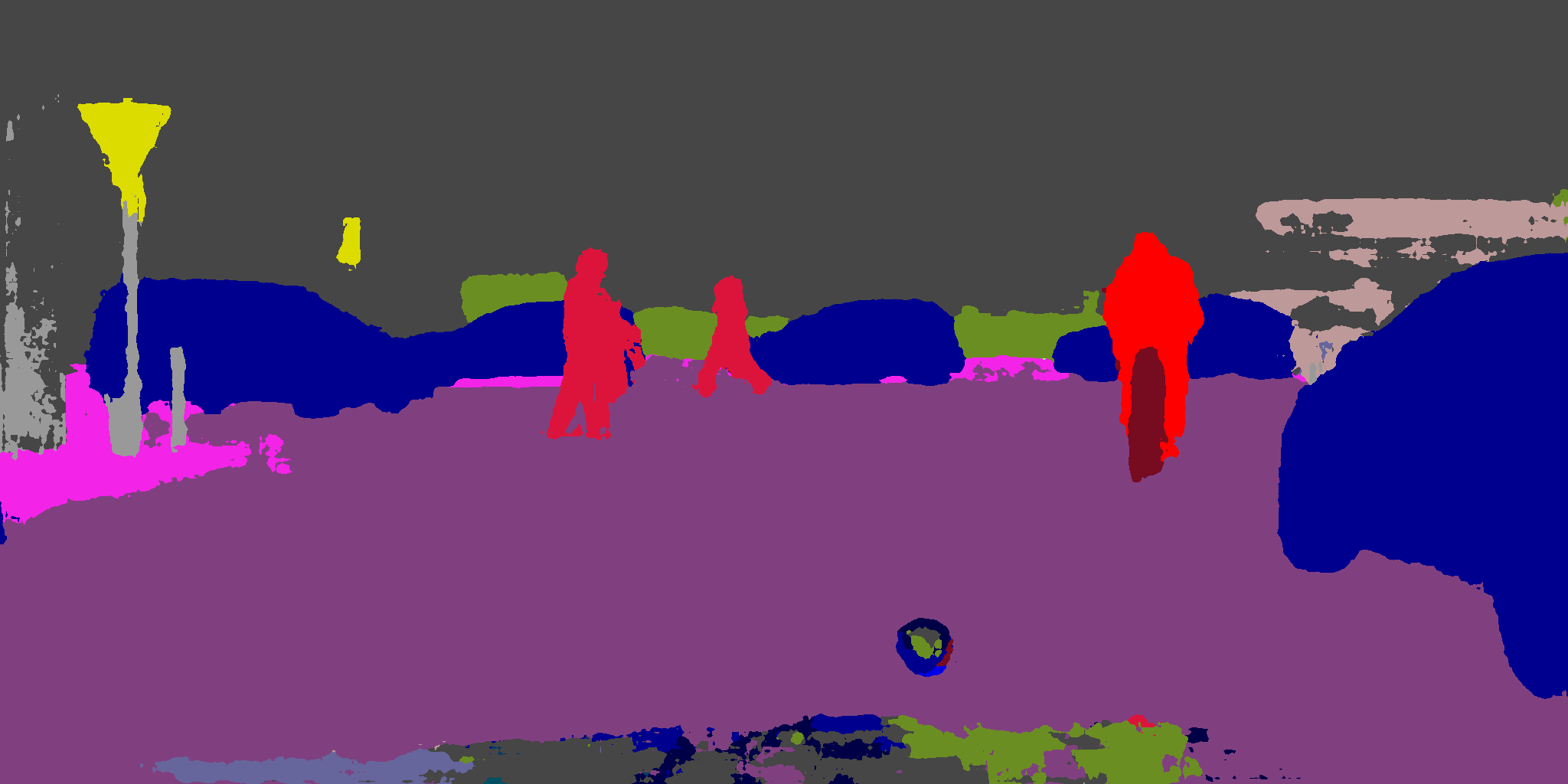}
}
&
\subfloat{
  \includegraphics[valign=m,width=0.225\linewidth]{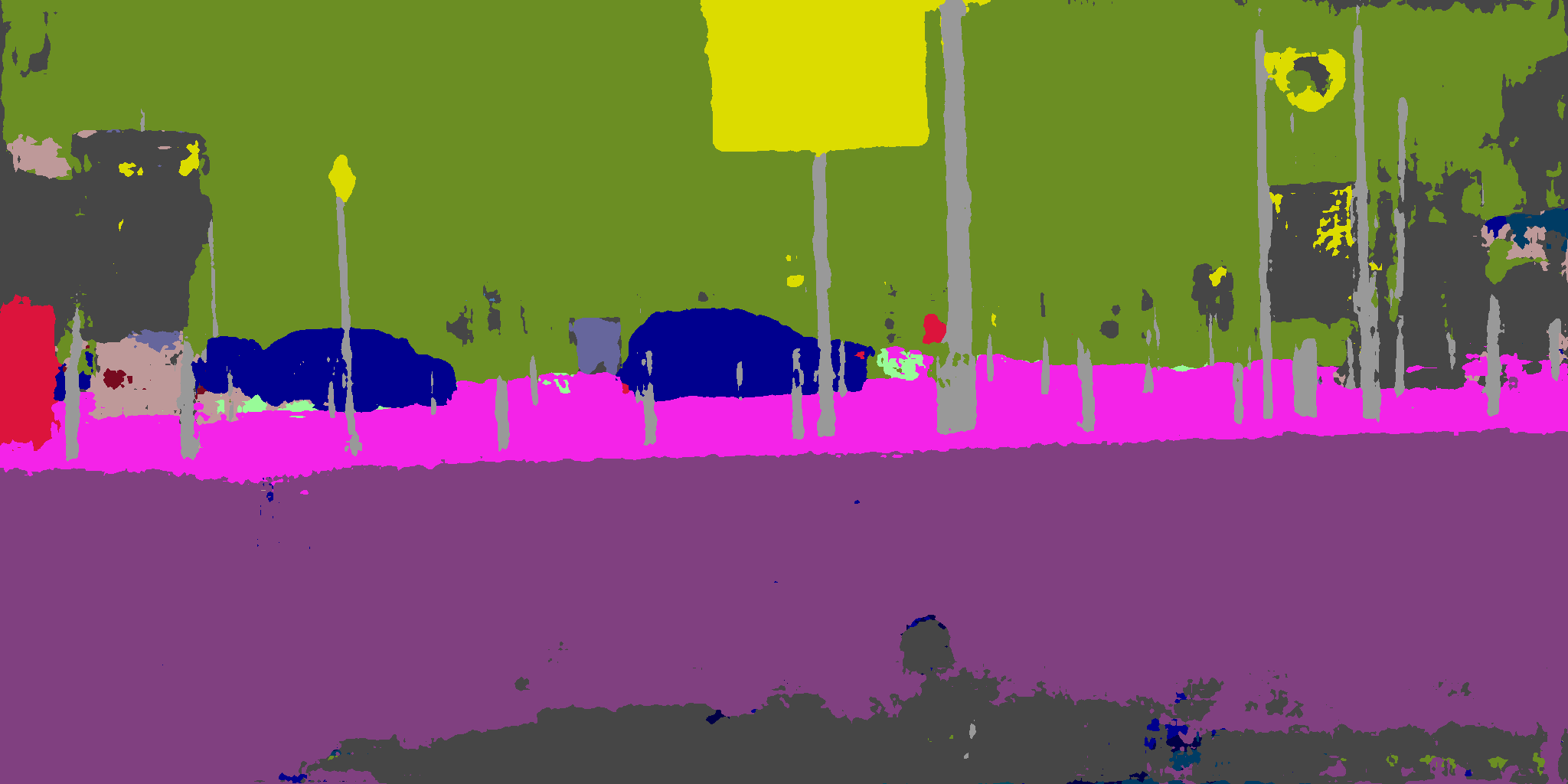}
} 
&
\subfloat{
  \includegraphics[valign=m,width=0.225\linewidth]{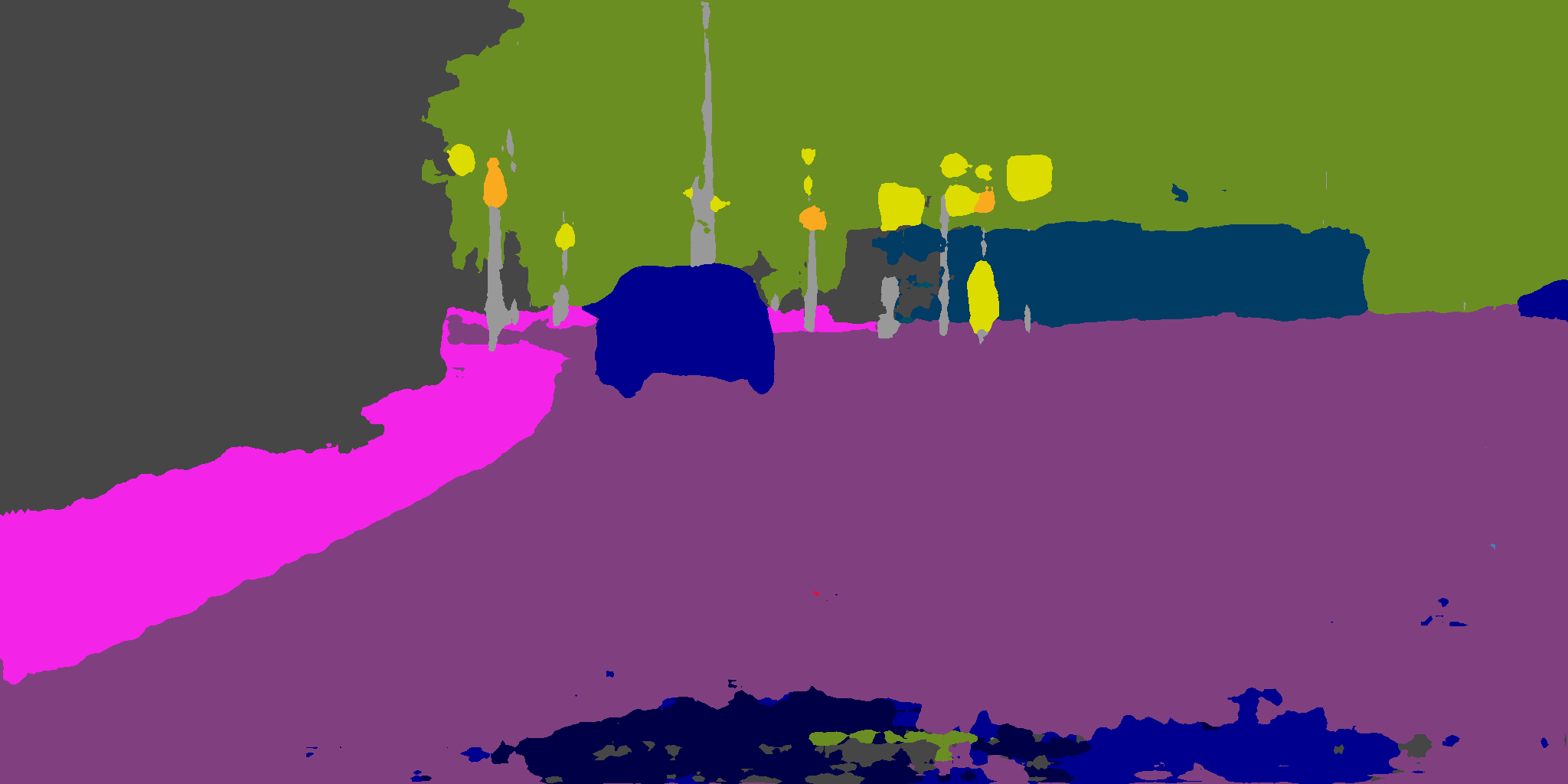}
} 
&
\subfloat{
  \includegraphics[valign=m,width=0.225\linewidth]{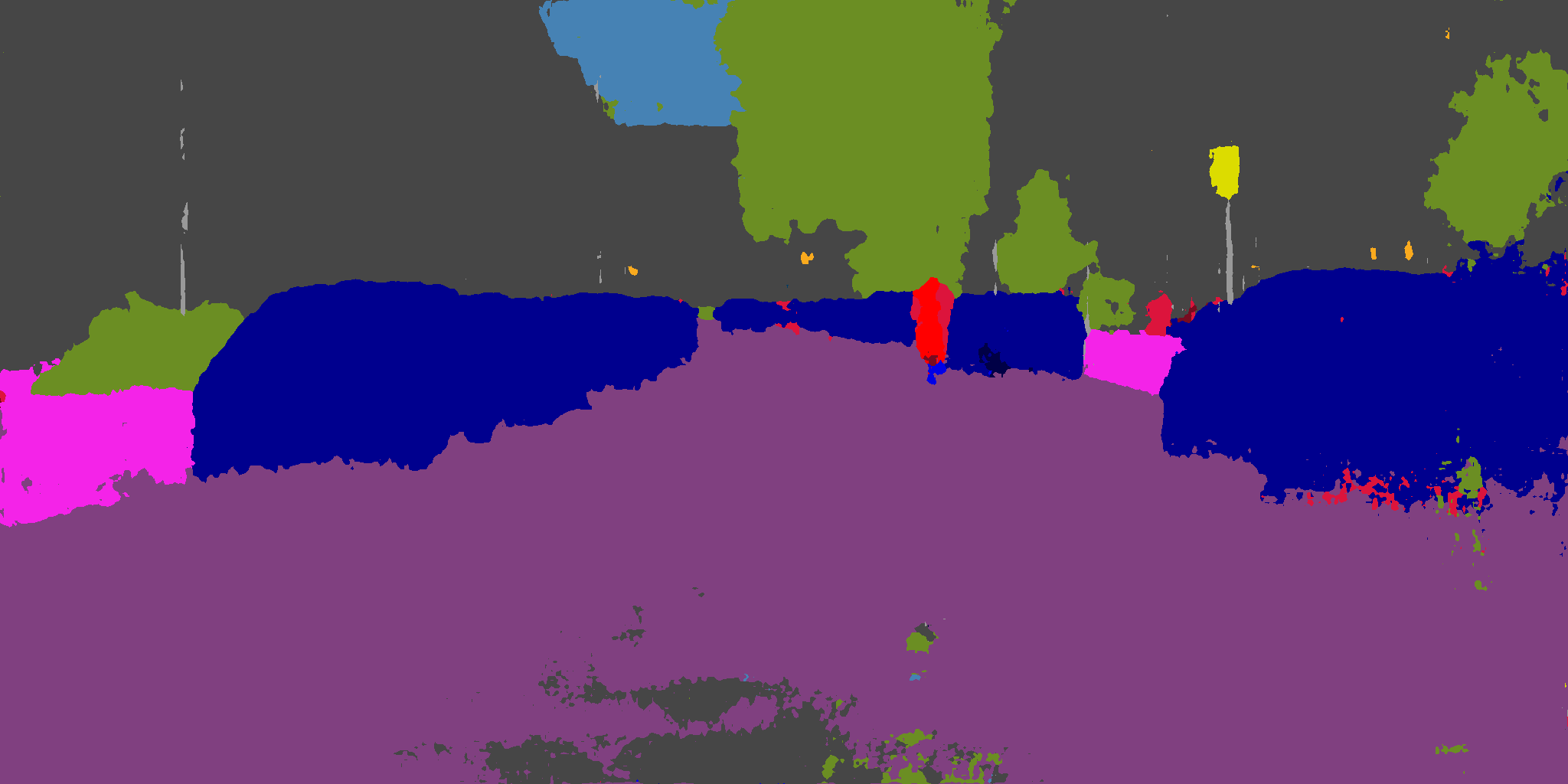}
} 
\\ [+5ex]
\multicolumn{4}{l}{
\subfloat{
  \includegraphics[width=1\linewidth]{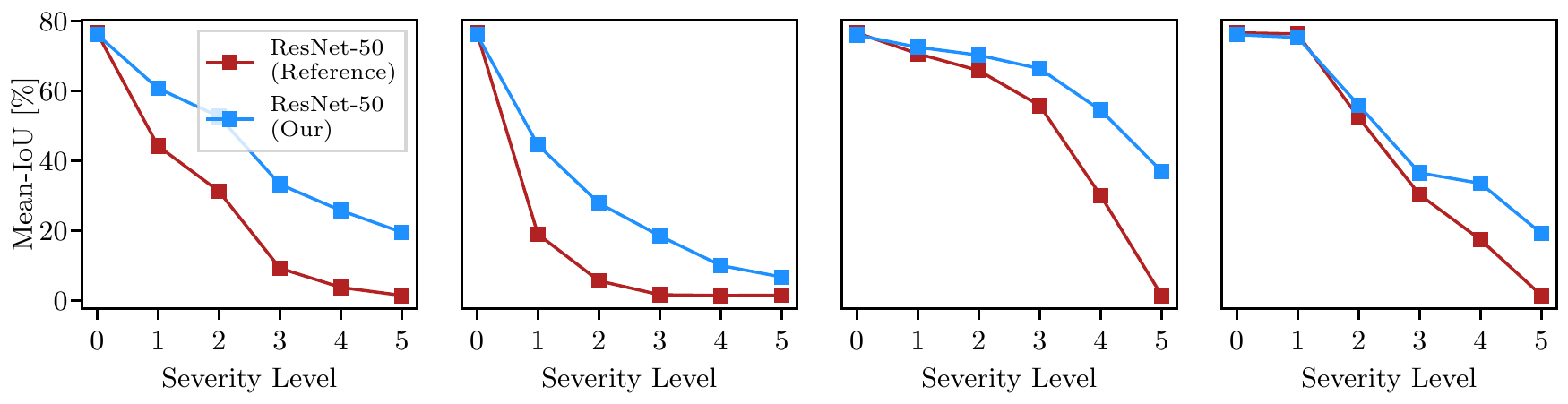}
}  
}
\end{tabular}~
}
\caption{
\textbf{(top)} Qualitative results by the ResNet-50 backbone on four corrupted images of the Cityscapes validation dataset for both the reference model and our model (i.e., trained with Painting-by-Numbers).
\textbf{(bottom)} Quantitative results on the corrupted variants of the Cityscapes dataset.
Each image corruption is parameterized with five severity levels, where severity level $0$ corresponds to clean (i.e., original) data.
While for clean data, both models' performance is more or less the same, we see that our model is clearly superior for all types of noise added.
For consistent performance on clean data, the performance on corrupted data increases when the model is trained with Painting-by-Numbers.
For the first severity level of shot noise, the mIoU of our model is higher by \SI{25}{\%}
    }     
    \label{fig:corruptions}
\end{figure}

In the following, we refer to a network that is trained with standard training schema as the reference model (i.e., trained on clean data only), and to a model that is trained with Painting-by-Numbers as our model.
Fig.~\ref{fig:corruptions} shows qualitative and quantitative results on corrupted variants of the Cityscapes dataset, when a network (ResNet-50) is trained with both training schemes.
Every image corruption is parameterized with five severity levels. 
Severity level $0$ corresponds to the clean data. 

The reference model (third row) struggles to predict well in the presence of image corruptions (Fig.~\ref{fig:corruptions} top).
It segments large parts of \textit{road} wrongly as \textit{building} for \textit{spatter} and \textit{image noise}.
When the same model is trained with Painting-by-Numbers, the predictions are clearly superior (fourth row).
With respect to quantitative results (Fig.~\ref{fig:corruptions} bottom), our model performs significantly better for image corruptions of category \textit{speckle noise}, \textit{shot noise}, and \textit{contrast}.
Corruption \textit{contrast} decreases the contrast of the full image, corrupting hence the textural image content strongly.
A network that is able to rely also on shape-based cues for the image segmentation is hence a well-performing model for \textit{contrast reduction}.
The mean IoU on \textit{spatter} is for both models comparable for the first severity level, but it is for our model higher by almost \SI{15}{\%} for the fourth severity level.
%\begin{minipage}[c]{0.03\textwidth}
%    \rotatebox[origin=r]{90}{GT}
%\end{minipage}

The results for the remaining image corruptions for the Cityscapes dataset are listed in Table~\ref{tab:results_cs}.
We show the effectiveness of Painting-by-Numbers besides ResNet-50 also for MobileNet-V2, ResNet-101, Xception-41, and Xception-71.
In the first column, we report the performance on clean data, i.e., the original Cityscapes validation set.
The mIoU evaluated on several types of image corruptions is listed accordingly.
Each value is the average for up to five severity levels.
We report for both clean and corrupted data, the result of the reference model and our model.
In the following, we discuss the main results of Table~\ref{tab:results_cs}.

\begin{table}[hbtp]
    \centering
    \caption{        
        Results on the Cityscapes dataset. 
        Each entry is the mean IoU of several corrupted variants of the Cityscapes dataset.
        Every image corruption is parameterized with five severity levels, and the resulting mean IoU are averaged.
        For image noise-based corruption, we exclude every severity level whose signal-to-noise ratio less than 10.
        The higher mIoU of either the reference model or the respective model trained with Painting-by-Numbers is bold.
        Overall, we see many (\SI{74}{\%}) more bold numbers for our Painting-by-Numbers model
        }
    \label{tab:results_cs}
    \begin{adjustbox}{width=\textwidth}

\begin{tabular}{@{}cccccccccccccccccc@{}}
\toprule
 &  & \multicolumn{4}{c}{\textbf{Blur}} & \multicolumn{4}{c}{\textbf{Noise}} & \multicolumn{4}{c}{\textbf{Digital}} & \multicolumn{4}{c}{\textbf{Weather}} \\ \midrule
\multicolumn{1}{c|}{Network} & \multicolumn{1}{c|}{Clean} & Motion & Defocus & \begin{tabular}[c]{@{}c@{}}Frosted \\ Glass\end{tabular} & \multicolumn{1}{c|}{Gaussian} & Gaussian & Impulse & Shot & \multicolumn{1}{c|}{Speckle} & Brightness & Contrast & Saturate & \multicolumn{1}{c|}{JPEG} & Snow & Spatter & Fog & Frost \\ \midrule
\multicolumn{1}{c|}{\textbf{Reference}} & \multicolumn{1}{c|}{} &  &  &  & \multicolumn{1}{c|}{} &  &  &  & \multicolumn{1}{c|}{} &  &  &  & \multicolumn{1}{c|}{} &  &  &  &  \\
\multicolumn{1}{c|}{MobileNet-V2} & \multicolumn{1}{c|}{\textbf{73.0}} & \textbf{52.4} & \textbf{47.0} & \textbf{44.7} & \multicolumn{1}{c|}{\textbf{48.1}} & 9.6 & 14.2 & 9.8 & \multicolumn{1}{c|}{25.6} & 50.4 & 43.8 & 32.5 & \multicolumn{1}{c|}{\textbf{20.3}} & 10.8 & 43.3 & 47.7 & 16.1 \\
\multicolumn{1}{c|}{ResNet-50} & \multicolumn{1}{c|}{\textbf{76.6}} & 57.1 & \textbf{55.2} & 45.3 & \multicolumn{1}{c|}{\textbf{56.5}} & 10.7 & 13.4 & 12.1 & \multicolumn{1}{c|}{37.7} & 59.8 & 52.7 & 41.7 & \multicolumn{1}{c|}{\textbf{23.4}} & \textbf{12.9} & 39.8 & 56.2 & 19.0 \\
\multicolumn{1}{c|}{ResNet-101} & \multicolumn{1}{c|}{76.0} & \textbf{58.9} & \textbf{55.3} & 47.8 & \multicolumn{1}{c|}{\textbf{56.3}} & 22.9 & 22.9 & 23.1 & \multicolumn{1}{c|}{45.5} & 57.7 & 56.8 & 41.6 & \multicolumn{1}{c|}{\textbf{32.5}} & \textbf{11.9} & 45.5 & 55.8 & 23.2 \\
\multicolumn{1}{c|}{Xception-41} & \multicolumn{1}{c|}{77.8} & 61.6 & \textbf{54.9} & 51.0 & \multicolumn{1}{c|}{\textbf{54.7}} & 27.9 & 28.4 & 27.2 & \multicolumn{1}{c|}{53.5} & 63.6 & 56.9 & 51.7 & \multicolumn{1}{c|}{\textbf{38.5}} & 18.2 & 46.6 & 57.6 & 20.6 \\
\multicolumn{1}{c|}{Xception-71} & \multicolumn{1}{c|}{77.9} & 62.5 & \textbf{58.5} & \textbf{52.6} & \multicolumn{1}{c|}{\textbf{57.7}} & 22.0 & 11.5 & 21.6 & \multicolumn{1}{c|}{48.7} & 67.0 & 57.2 & 45.7 & \multicolumn{1}{c|}{\textbf{36.1}} & 16.0 & 48.0 & 63.9 & 20.5 \\ \midrule
\multicolumn{1}{c|}{\textbf{Painting-by-Numbers}} & \multicolumn{1}{c|}{} &  &  &  & \multicolumn{1}{c|}{} &  &  &  & \multicolumn{1}{c|}{} &  &  &  & \multicolumn{1}{c|}{} &  &  &  &  \\
\multicolumn{1}{c|}{MobileNet-V2} & \multicolumn{1}{c|}{72.2} & 49.5 & 41.4 & 40.7 & \multicolumn{1}{c|}{43.0} & \textbf{17.4} & \textbf{18.4} & \textbf{16.8} & \multicolumn{1}{c|}{\textbf{35.7}} & \textbf{62.5} & \textbf{50.8} & \textbf{51.0} & \multicolumn{1}{c|}{17.6} & \textbf{12.1} & \textbf{46.9} & \textbf{56.5} & \textbf{22.4} \\
\multicolumn{1}{c|}{ResNet-50} & \multicolumn{1}{c|}{76.1} & \textbf{58.1} & 53.5 & \textbf{50.3} & \multicolumn{1}{c|}{55.1} & \textbf{35.7} & \textbf{34.3} & \textbf{36.1} & \multicolumn{1}{c|}{\textbf{56.7}} & \textbf{68.8} & \textbf{64.2} & \textbf{60.5} & \multicolumn{1}{c|}{21.3} & 10.6 & \textbf{46.1} & \textbf{61.0} & \textbf{22.9} \\
\multicolumn{1}{c|}{ResNet-101} & \multicolumn{1}{c|}{\textbf{76.3}} & 58.1 & 54.2 & \textbf{48.7} & \multicolumn{1}{c|}{54.7} & \textbf{41.6} & \textbf{44.3} & \textbf{40.6} & \multicolumn{1}{c|}{\textbf{57.4}} & \textbf{70.5} & \textbf{64.4} & \textbf{65.0} & \multicolumn{1}{c|}{25.6} & 10.8 & \textbf{50.1} & \textbf{56.9} & \textbf{28.0} \\
\multicolumn{1}{c|}{Xception-41} & \multicolumn{1}{c|}{\textbf{78.5}} & \textbf{65.5} & 54.2 & \textbf{51.1} & \multicolumn{1}{c|}{51.8} & \textbf{46.9} & \textbf{44.9} & \textbf{46.9} & \multicolumn{1}{c|}{\textbf{64.3}} & \textbf{73.4} & \textbf{60.2} & \textbf{68.8} & \multicolumn{1}{c|}{15.7} & \textbf{19.3} & \textbf{55.8} & \textbf{65.7} & \textbf{28.2} \\
\multicolumn{1}{c|}{Xception-71} & \multicolumn{1}{c|}{\textbf{78.6}} & \textbf{63.0} & 53.6 & 48.6 & \multicolumn{1}{c|}{52.2} & \textbf{35.5} & \textbf{38.4} & \textbf{34.2} & \multicolumn{1}{c|}{\textbf{57.6}} & \textbf{74.9} & \textbf{63.9} & \textbf{69.1} & \multicolumn{1}{c|}{22.2} & \textbf{18.2} & \textbf{57.4} & \textbf{65.4} & \textbf{25.5} \\ \bottomrule

\end{tabular}

    \end{adjustbox}
\end{table}

\textbf{Performance w.r.t clean data.} 
Even though we paint the exact half of the training data and train both models for the same amount of iterations, the performance on clean data is oftentimes barely affected.

\textbf{Performance w.r.t image blur.} 
The robustness of our model with respect to image blur does not notably increase.
We assume that Painting-by-Numbers does not increase the performance for this category of image corruptions because blur corrupts the object shapes by smearing the object boundaries.
Hence, our learned shape-bias does not work well. 

\textbf{Performance w.r.t image noise.}
Painting-by-Numbers increases the robustness with respect to image noise the most (see figures above). 
For example, the absolute mIoU of Xception-41 for Gaussian noise, impulse noise, shot noise, and speckle noise increases by \SI{19.0}{\%}, \SI{16.5}{\%}, \SI{19.7}{\%}, and \SI{11.0}{\%}, respectively.

\textbf{Performance w.r.t digital corruptions.} 
A network trained with Painting-by-Numbers increases significantly the robustness against the corruptions \textit{brightness}, \textit{contrast}, and \textit{saturation}--but not JPEG artifacts.
The reason is that \textit{JPEG compression} corrupts the boundary of objects and incorporates new boundaries through posterization artifacts.
Our network cannot hence profit from its increased shape-bias. 
We refer to the supplement for an illustration.

\textbf{Performance w.r.t weather corruptions.} 
Xception-71 and Xception-41 increases the performance with respect to \textit{spatter} by \SI{9.4}{\%} and \SI{9.2}{\%}, respectively.
Xception-41 further increases the mIoU against \textit{frost} by \SI{7.6}{\%}.
Every model increases the performance against \textit{fog}.
We cannot observe a significant performance increase for \textit{snow}.

Though the performance increase on image corruptions of category \textit{weather} is less than, e.g., for image noise, the predictions of a network trained with Painting-by-Numbers are improved for key-classes such as \textit{cars}, \textit{persons}, and \textit{traffic signs} than for a regularly trained network.
Please see the supplementary material for more results.

\subsection{Understanding Painting-by-Numbers}
\label{sec:validation}
\begin{figure}[hbtp]
\centering
\subfloat[Replaced \textit{car} by RGB-mean]{
  \includegraphics[width=0.3\linewidth]{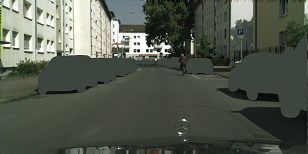}
}\hfil
\subfloat[Corrupted \textit{building} by severe noise]{
 \includegraphics[width=0.3\linewidth]{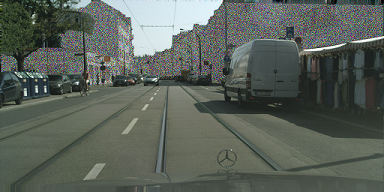}
}\hfil
\subfloat[Corrupted \textit{car} by severe blur]{
  \includegraphics[width=0.3\linewidth]{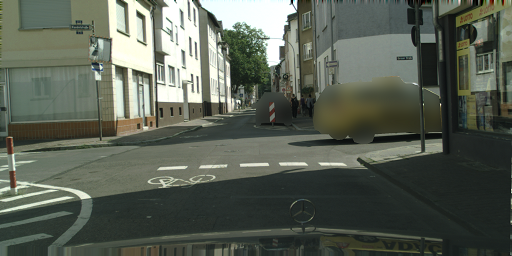}
}

\caption{
     Examples of image data to validate an increasing shape-bias when models are trained with Painting-by-Numbers.
     We remove, or strongly corrupt, the texture of each class in the Cityscapes dataset and evaluate the segmentation performance when a network cannot rely on the class texture.
    (a) Texture is fully replaced by the dataset-wide RGB-mean value of the respective class.
    (b) Class is corrupted by severe noise.
    (c) Class is corrupted by severe blur
}
\label{fig:validation}
\end{figure}

We explain the increased robustness towards common image corruptions, i.e., when a network is trained with Painting-by-Numbers, by an increased shape-bias.
To validate this assumption, we conduct a series of experiments that are based on the following consideration:
Classes that either have a) no texture at all or b) texture that is strongly corrupted should be more reliably segmented by a network trained with Painting-by-Numbers.
In more detail, we generate numerous, on class-level corrupted, variants of the Cityscapes validation set, as illustrated in Fig.~\ref{fig:validation}. 
In (a), we remove the texture of \textit{cars} and replace it by the dataset-wide RGB-mean of the training set of the respective class.
The respective class does, in this way, not contain any texture but homogeneous color information. 
In (b) and (c) we corrupt \textit{building} and \textit{car} by a high degree of additive Gaussian noise and Gaussian blur, respectively.
Please note that Fig.~\ref{fig:validation} shows only a small set of examples.
We apply these corruptions for every class.

We test the models on such images to evaluate if they are capable of segmenting the respective class when they cannot rely on the class texture.
To achieve this, a network needs to utilize other cues, such as shape-based cues.

Instead of IoU, we use the sensitivity $s$ ($s = TP/(TP+FN)$, where $TP$ are true-positives, and $FN$ are false-negatives) as evaluation metric. 
The sensitivity is for these experiments more appropriate than IoU ($IoU = TP/(TP+FN+FP)$) since we are solely interested in the segmentation performance on the class-level.
Because all classes but one is clean (i.e., not corrupted), false-positively (FP) segmented pixels are of less interest.
Utilizing IoU could, especially for classes covering fewer image regions, result in misleading scores.
The results of these experiments are listed in Table~\ref{tab:validation_results}.
\begin{table}[ht]
    \centering
    \caption{
        Sensitivity score per class for several corrupted variants on the class-level of the Cityscapes datasets.
        \textbf{Clean:} The performance on clean (i.e. original, non-corrupted) data.
        \textbf{RGB-mean:} The texture of a class is replaced by the dataset-wide RGB mean of that class.
        \textbf{Noise:} The texture of a class is corrupted by severe additive Gaussian noise.    
        \textbf{Blur:} The texture of a class is corrupted by severe Gaussian blur.
        The higher sensitivity score of a network backbone of either the reference (top) or our model (bottom) is bold.
        Overall, we see many more bold numbers for our Painting-by-Numbers model
}
    \label{tab:validation_results}
\begin{adjustbox}{width=\textwidth}
    \begin{tabular}{@{}cccccccccccccccccccc@{}}    & \rotatebox[origin=l]{90}{road} & \rotatebox[origin=l]{90}{sidewalk} & \rotatebox[origin=l]{90}{building} & \rotatebox[origin=l]{90}{wall} & \rotatebox[origin=l]{90}{fence} & \rotatebox[origin=l]{90}{pole} & \rotatebox[origin=l]{90}{traffic light} & \rotatebox[origin=l]{90}{traffic sign} & \rotatebox[origin=l]{90}{vegetation} & \rotatebox[origin=l]{90}{terrain} & \rotatebox[origin=l]{90}{sky} & \rotatebox[origin=l]{90}{person} & \rotatebox[origin=l]{90}{rider} & \rotatebox[origin=l]{90}{car} & \rotatebox[origin=l]{90}{truck} & \rotatebox[origin=l]{90}{bus} & \rotatebox[origin=l]{90}{train} & \rotatebox[origin=l]{90}{motorcycle} & \rotatebox[origin=l]{90}{bicycle}  \\ \midrule
        \textbf{Reference} &  &  &  &  &  &  &  &  &  &  &  &  &  &  &  &  &  &  &  \\
Clean & 98.8 & \textbf{93.1} & \textbf{96.6} & 53.3 & \textbf{69.6} & \textbf{74.5} & \textbf{81.7} & \textbf{85.7} & \textbf{96.7} & \textbf{74.1} & \textbf{97.8} & \textbf{91.5} & \textbf{76.9} & \textbf{97.6} & 85.1 & \textbf{92.8} & \textbf{70.7} & \textbf{79.2} & \textbf{88.7} \\
RGB-mean & 92.7 & 21.1 & \textbf{88.9} & \textbf{40.7} & 5.4 & 68.9 & 12.8 & 31.5 & 1.4 & \textbf{3.3} & \textbf{97.7} & 73.4 & 62.5 & 24.5 & 13.6 & \textbf{16.3} & \textbf{9.0} & 2.9 & 0.9 \\
Noise ($\mathit{scale}=0.5$) & 5.8 & 0.8 & \textbf{95.0} & 0.2 & 1.7 & 4.7 & 6.4 & \textbf{39.2} & 0.1 & 1.4 & 2.8 & 8.1 & 2.9 & 4.5 & 0.0 & 0.0 & \textbf{7.1} & 0.2 & 3.4 \\
Noise ($\mathit{scale}=1.0$) & 2.9 & 0.0 & \textbf{94.0} & 0.1 & 1.2 & 2.0 & 6.4 & \textbf{40.2} & 0.0 & 0.5 & 0.2 & 4.4 & 1.2 & 3.6 & 0.0 & 0.0 & \textbf{3.0} & 0.2 & 2.0 \\
Blur ($\sigma=20$) & \textbf{94.6} & 42.8 & \textbf{89.3} & \textbf{38.0} & 1.8 & 63.6 & 18.4 & 19.1 & 0.6 & \textbf{7.3} & 94.0 & 55.4 & 55.5 & 56.7 & 32.5 & \textbf{24.0} & \textbf{7.7} & 9.0 & 0.9 \\
Blur ($\sigma=30$) & \textbf{94.1} & 42.1 & \textbf{89.4} & \textbf{33.2} & 1.2 & 62.3 & 14.0 & 16.2 & 0.6 & \textbf{2.3} & 93.8 & 54.6 & 51.8 & 44.2 & 29.8 & \textbf{18.6} & \textbf{8.1} & 4.2 & 1.1 \\ \midrule
\textbf{Painting-by-Numbers} &  &  &  &  &  &  &  &  &  &  &  &  &  &  &  &  &  &  &  \\
Clean & \textbf{99.0} & 90.3 & 96.3 & \textbf{56.0} & 67.1 & 68.9 & 76.5 & 81.1 & 96.2 & 66.3 & 97.1 & 89.5 & 74.0 & 96.8 & \textbf{89.7} & 86.0 & 59.6 & 72.2 & 87.8 \\
RGB-mean & \textbf{97.9} & \textbf{53.8} & 51.2 & 34.2 & \textbf{14.9} & \textbf{79.7} & \textbf{38.4} & \textbf{40.5} & \textbf{1.8} & 2.3 & 97.4 & \textbf{78.4} & \textbf{66.3} & \textbf{78.6} & \textbf{37.6} & 3.5 & 0.4 & \textbf{9.1} & \textbf{4.6} \\
Noise ($\mathit{scale}=0.5$) & \textbf{97.4} & \textbf{50.9} & 92.1 & \textbf{8.4} & \textbf{37.4} & \textbf{34.1} & \textbf{8.2} & 11.1 & \textbf{23.3} & \textbf{30.6} & \textbf{32.3} & \textbf{50.1} & \textbf{19.7} & \textbf{49.8} & \textbf{31.5} & \textbf{1.9} & 0.0 & \textbf{0.3} & \textbf{26.7} \\
Noise ($\mathit{scale}=1.0$) & \textbf{95.9} & \textbf{51.7} & 91.3 & \textbf{9.6} & \textbf{29.4} & \textbf{32.3} & \textbf{7.1} & 9.9 & \textbf{12.2} & \textbf{27.2} & \textbf{33.6} & \textbf{52.7} & \textbf{21.3} & \textbf{40.6} & \textbf{25.8} & \textbf{1.1} & 0.0 & \textbf{0.4} & \textbf{23.3} \\
Blur ($\sigma=20$) & 49.3 & \textbf{43.5} & 86.5 & 18.7 & \textbf{4.7} & \textbf{73.6} & \textbf{55.1} & \textbf{29.8} & \textbf{1.0} & 0.8 & \textbf{94.3} & \textbf{75.5} & \textbf{73.2} & \textbf{71.9} & \textbf{56.6} & 7.9 & 0.5 & \textbf{20.2} & \textbf{3.5} \\
Blur ($\sigma=30$) & 46.3 & \textbf{48.0} & 83.2 & 14.1 & \textbf{4.5} & \textbf{73.6} & \textbf{49.7} & \textbf{25.4} & \textbf{1.0} & 0.5 & \textbf{94.7} & \textbf{74.6} & \textbf{71.1} & \textbf{73.7} & \textbf{47.9} & 3.8 & 0.2 & \textbf{18.2} & \textbf{3.8} \\ \bottomrule

    \end{tabular}
\end{adjustbox}
\end{table}

\textbf{Quantitative results.}
The results in Table~\ref{tab:validation_results} are created by DeepLabv3$+$ with ResNet-50 as network backbone.
As previously, we refer to a network that is trained with the standard training schema as the reference model (i.e., only clean data used), and to a model that is trained with Painting-by-Numbers as our model.
The top (bottom) part of the Table contains the sensitivity score for each class of the reference model (our model).
Each line shows the sensitivity for the corrupted data as described previously (the performance on clean data is also listed).
The higher sensitivity of a network backbone of either the reference model (top) or our model (bottom) is bold.
We separately discuss in the following the quantitative results for class categories ``stuff'' and ``things''.

Both networks perform well for classes ``stuff'' since the amount of texture is often poor, such as for \textit{road}, \textit{wall}, \textit{sidewalk}, and \textit{sky}.
The sensitivity of both models differs for \textit{road} by \SI{5.2}{\%}, for \textit{wall} by \SI{6.5}{\%}, and for sky by \SI{0.3}{\%}.
Whereas the absolute sensitivity for both models is above \SI{90.0}{\%} for \textit{road} and \textit{sky}, it is less than \SI{41}{\%} for \textit{wall}.
Our model performs for \textit{sidewalk} %significantly 
better by \SI{32.7}{\%}. 

% building terrain 
Painting-by-Numbers performs worse than the reference for classes ``stuff'' with a large amount of textual information, such as \textit{building}, \textit{vegetation}, and \textit{terrain}.
For example, the sensitivity score of our model for \textit{building} is \SI{37.7}{\%} less.
Classes ``stuff'' have no distinct shape, hence, Painting-by-Numbers does not aid performance.
When, additionally, the amount of texture of a class is large, the sensitivity of our model is less than of the reference model.

% persons reference
The reference model performs well when the texture of the category ``things'' is replaced by RGB-mean.
Its sensitivity for \textit{person} is \SI{73.4}{\%}, which is only \SI{5.0}{\%} less than for our model.
The result for class \textit{rider} is similar.

% car
However, our model performs often significantly better than the reference model for most of the remaining ``things'' such as \textit{car}.
The sensitivity score of our model for this class is $s_\mathit{ours}=\SI{78.6}{\%}$, which is \SI{54.1}{\%} higher than the sensitivity score of the reference model.
We explain this high score with a large shape-bias due to both the distinct shape of \textit{cars} and the comparatively large number of \textit{cars} in the training set~\cite{cordts_cityscapes_2016}. 
Our model performs for other classes of ``things'' also better than the reference model.
For example, the sensitivity score for classes \textit{traffic light}, \textit{traffic sign} and \textit{pole} is higher by \SI{25.6}{\%}, \SI{9.0}{\%}, and \SI{9.8}{\%}, respectively.
Both models perform poorly on ``vehicles'' that are, compared to \textit{cars}, less frequent present in the training set (e.g., \textit{truck, motorcycle, train}).

\begin{figure}[ht]
\centering
\resizebox{\textwidth}{!}{
\begin{tabular}{c c c c}
\centering
(a) Original & (b) Mean-RGB & (c) Blur & (d) Noise \\ 
\subfloat{
\rotatebox[origin=b]{90}{RGB}
  \includegraphics[valign=m,width=0.23\linewidth]{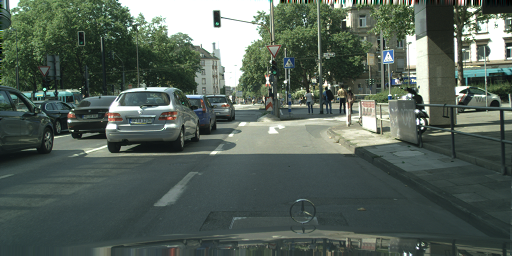}
}
&
\subfloat{
  \includegraphics[valign=m,width=0.23\linewidth]{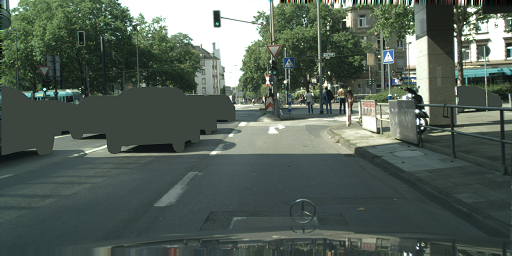}
}
&
\subfloat{
 \includegraphics[valign=m,width=0.23\linewidth]{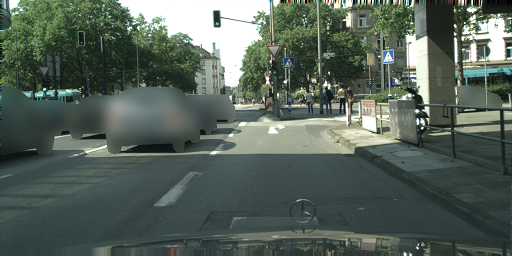}
}
&
\subfloat{
  \includegraphics[valign=m,width=0.23\linewidth]{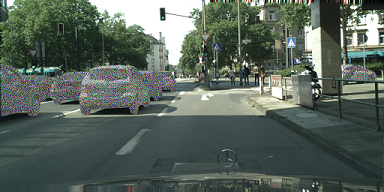}
}
\\ [-2ex]
\subfloat{
  \rotatebox[origin=b]{90}{GT}
  \includegraphics[valign=m,width=0.23\linewidth]{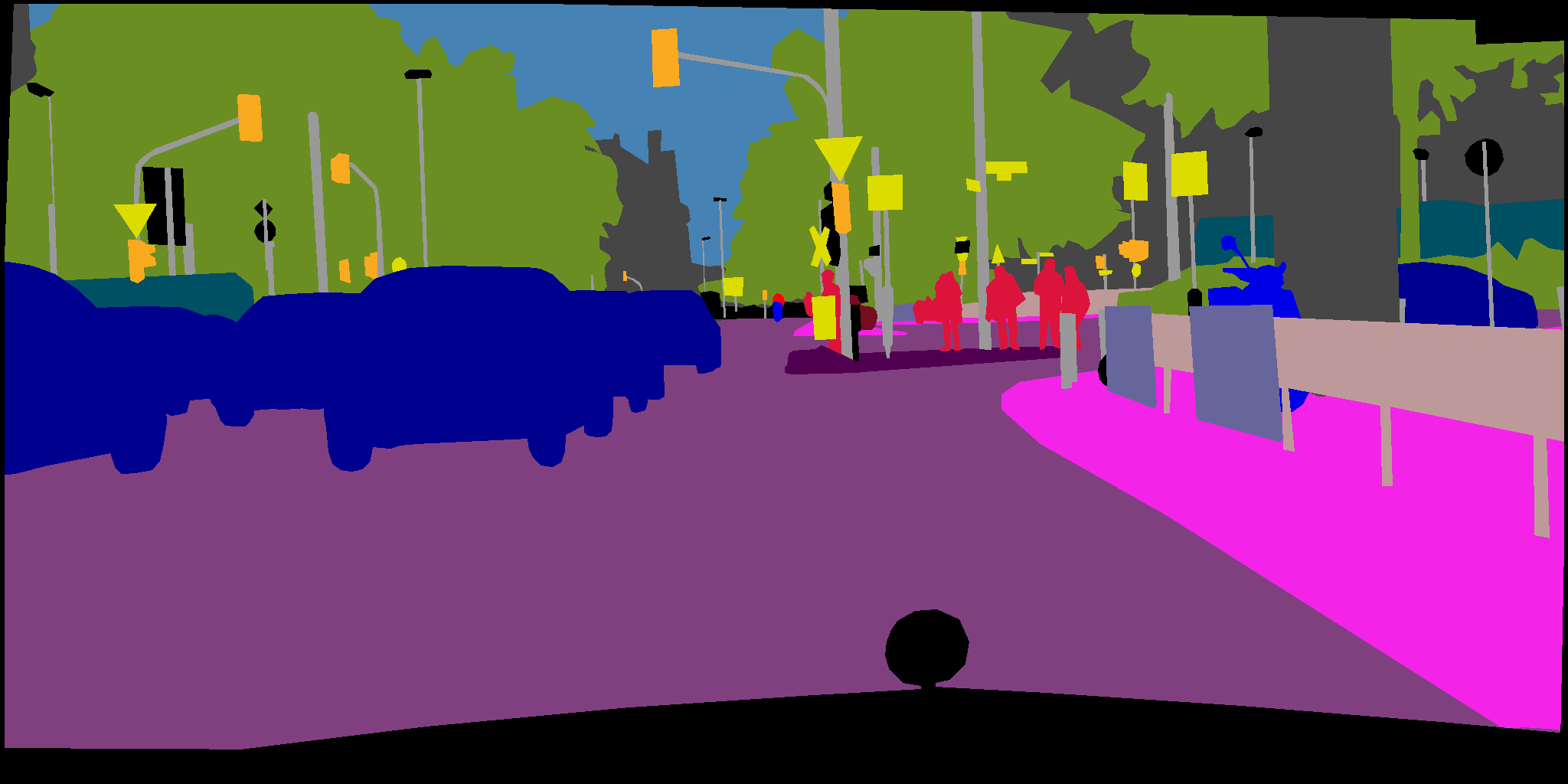}
}
&
\subfloat{
  \includegraphics[valign=m,width=0.23\linewidth]{2gt.png}
}
&
\subfloat{
  \includegraphics[valign=m,width=0.23\linewidth]{2gt.png}
}
&
\subfloat{
  \includegraphics[valign=m,width=0.23\linewidth]{2gt.png}
}
\\ [-2ex]
\subfloat{
  \rotatebox[origin=b]{90}{Reference}
  \includegraphics[valign=m,width=0.23\linewidth]{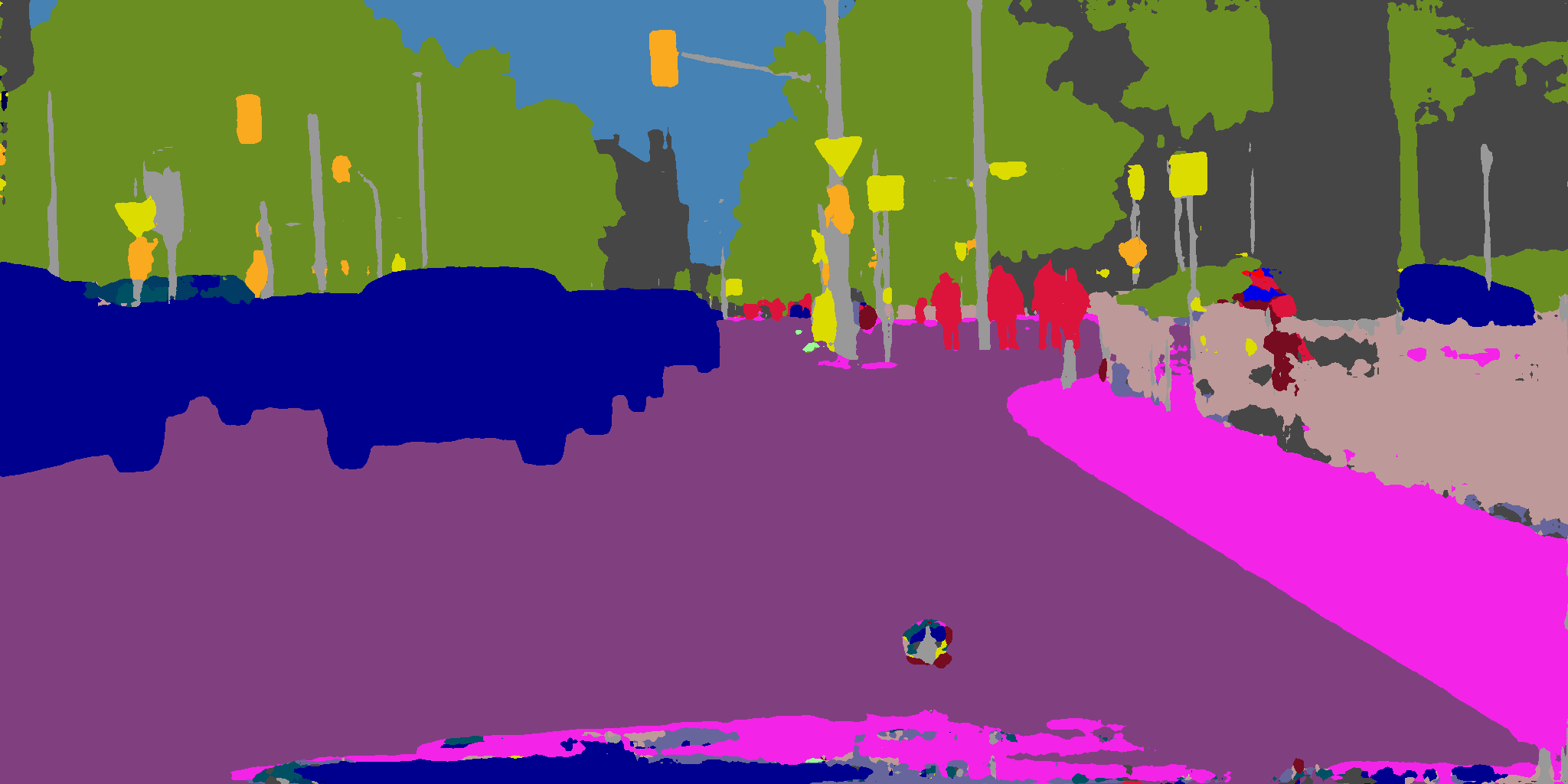}
}
&
\subfloat{
  \includegraphics[valign=m,width=0.23\linewidth]{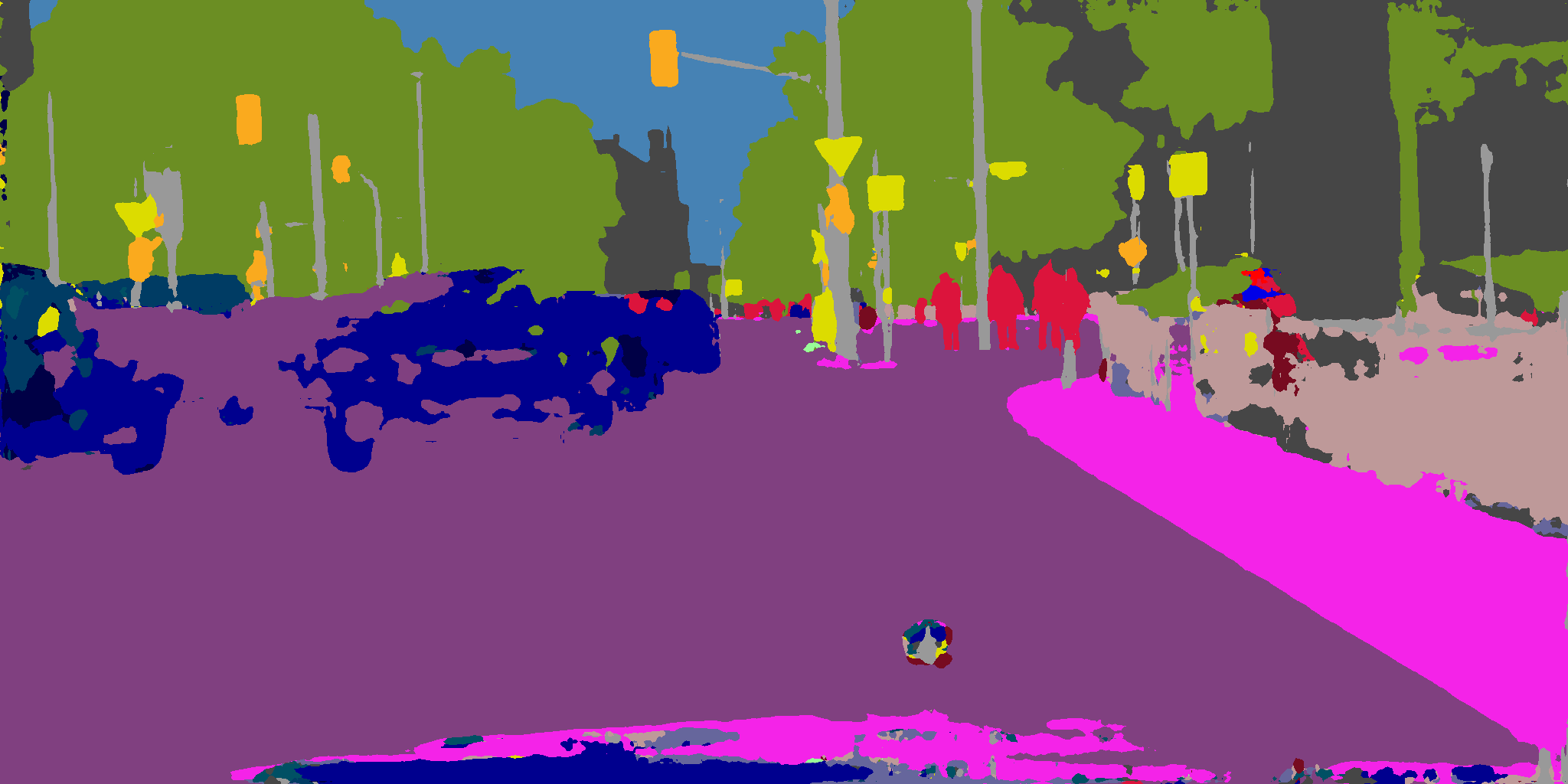}
}
&
\subfloat{
  \includegraphics[valign=m,width=0.23\linewidth]{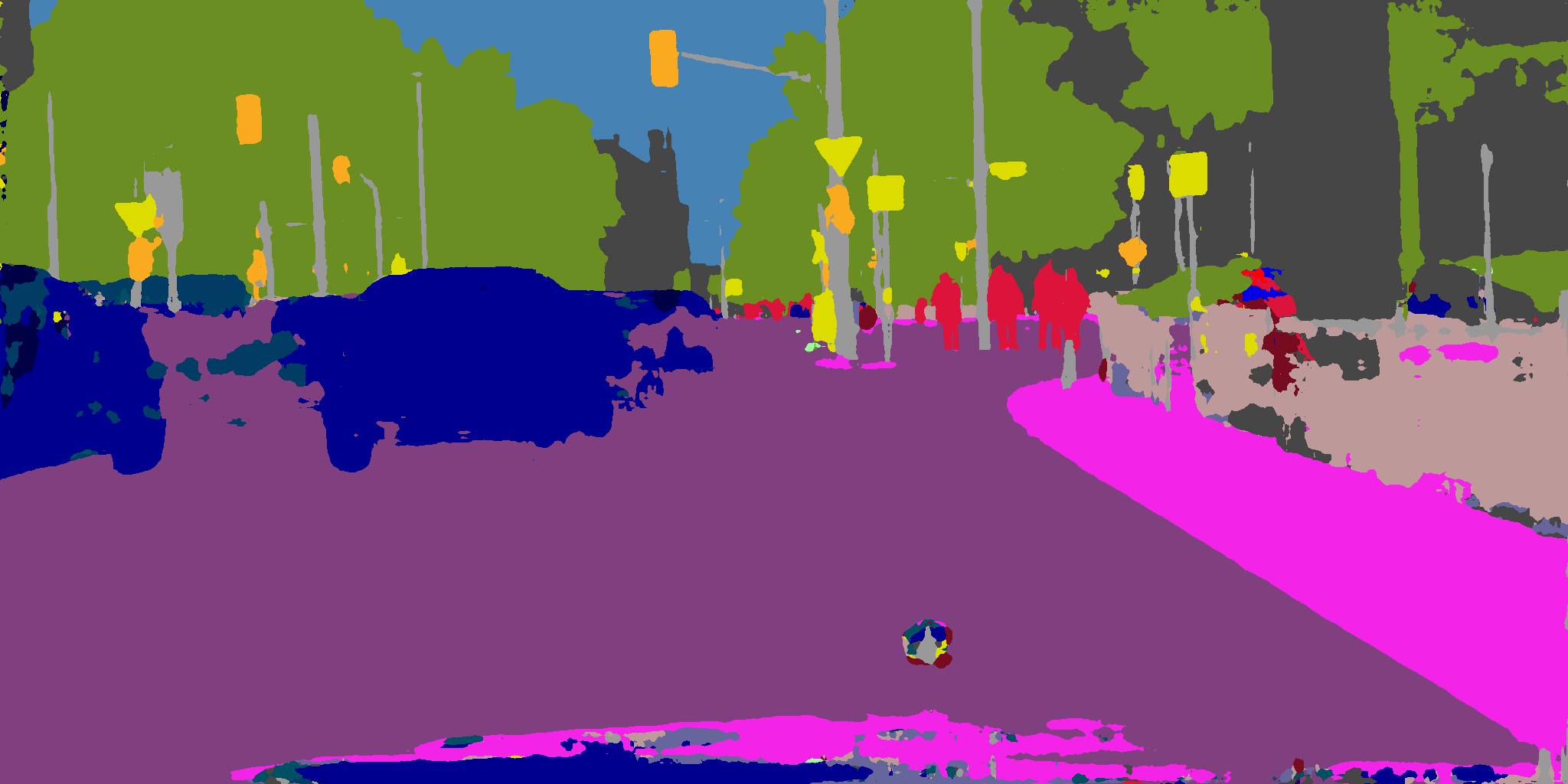}
}
&
\subfloat{
  \includegraphics[valign=m,width=0.23\linewidth]{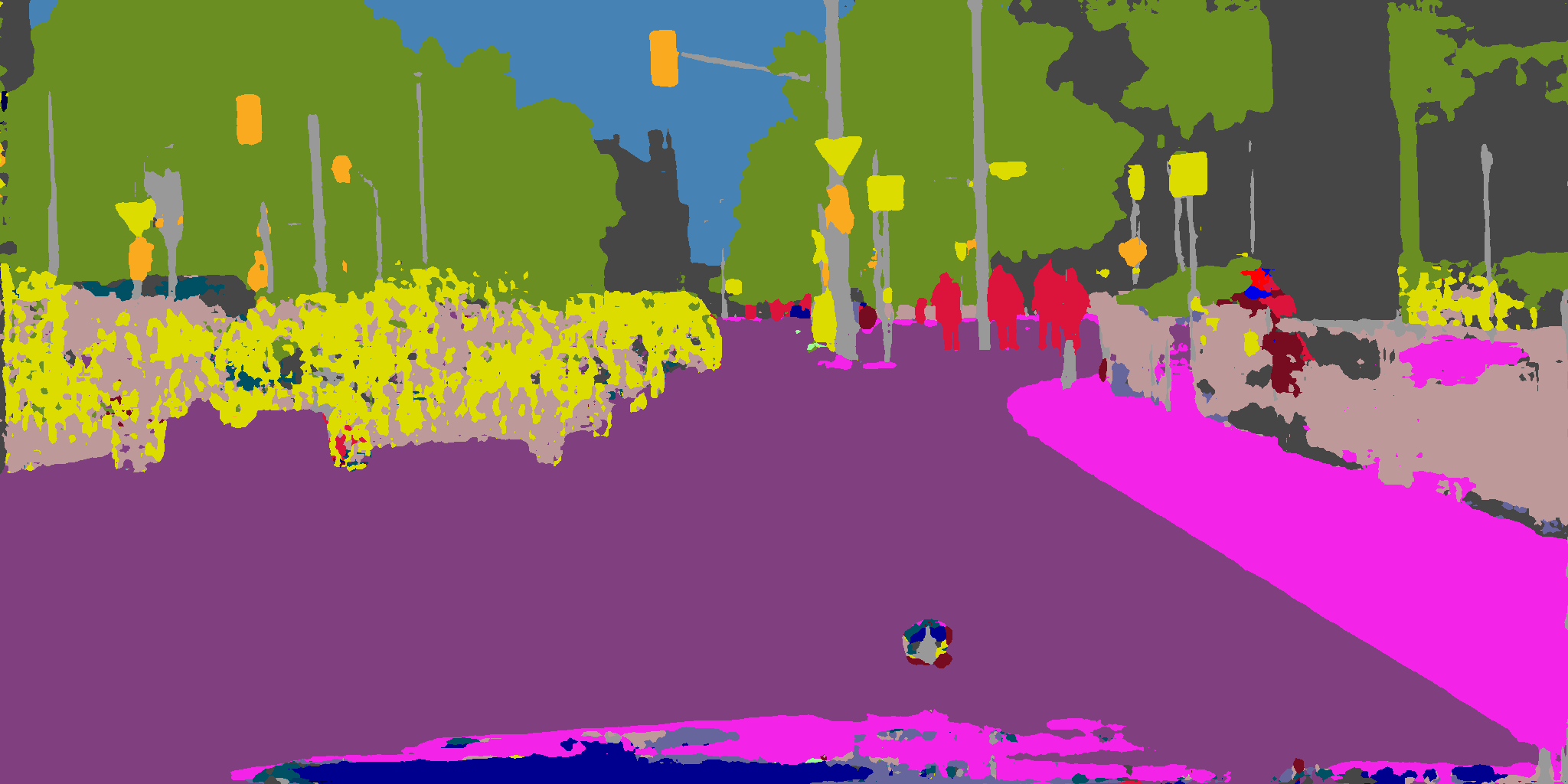}
}
\\ [-2ex]
\subfloat{
  \rotatebox[origin=b]{90}{Ours}
  \includegraphics[valign=m,width=0.23\linewidth]{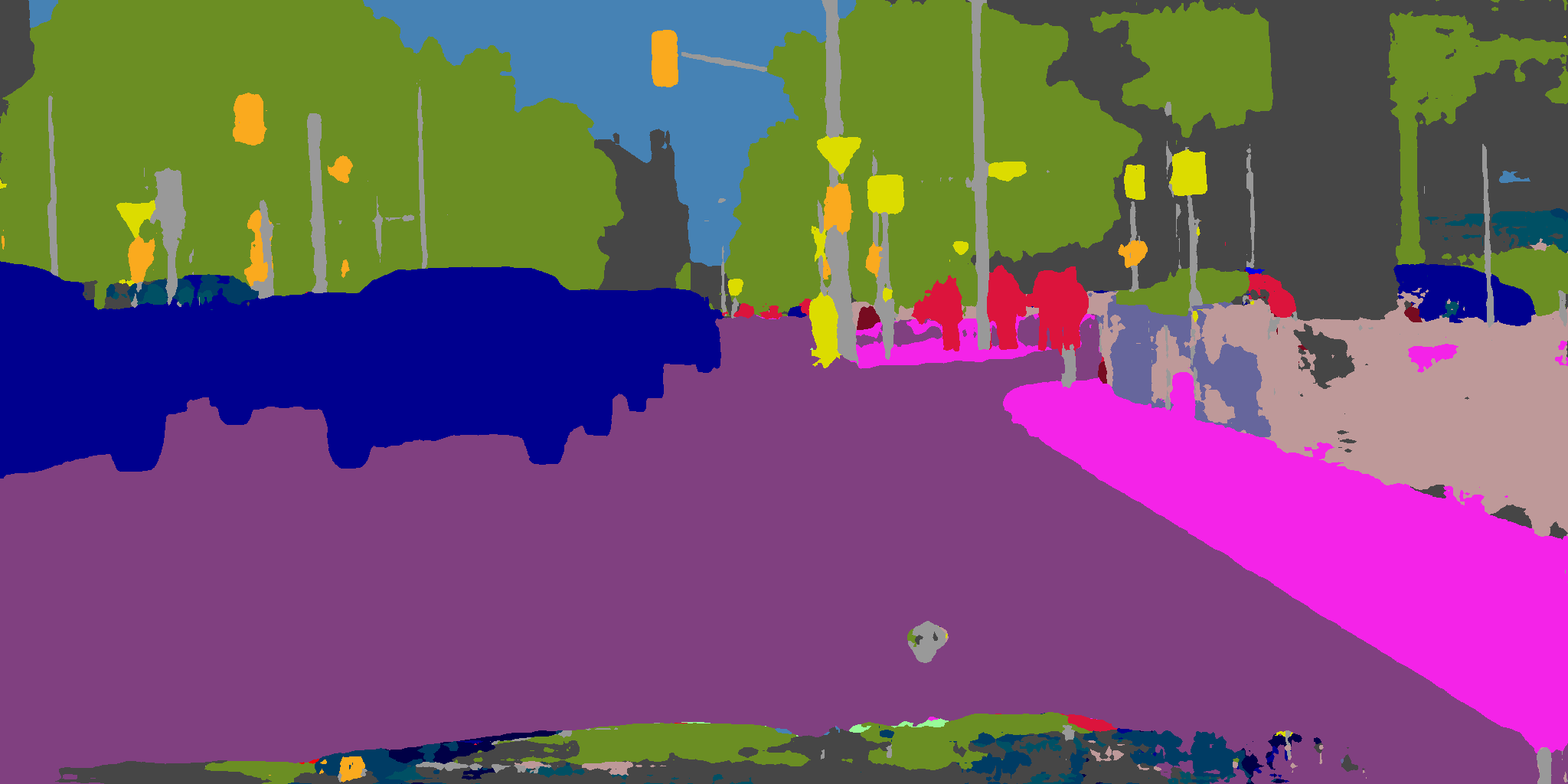}
}
&
\subfloat{
  \includegraphics[valign=m,width=0.23\linewidth]{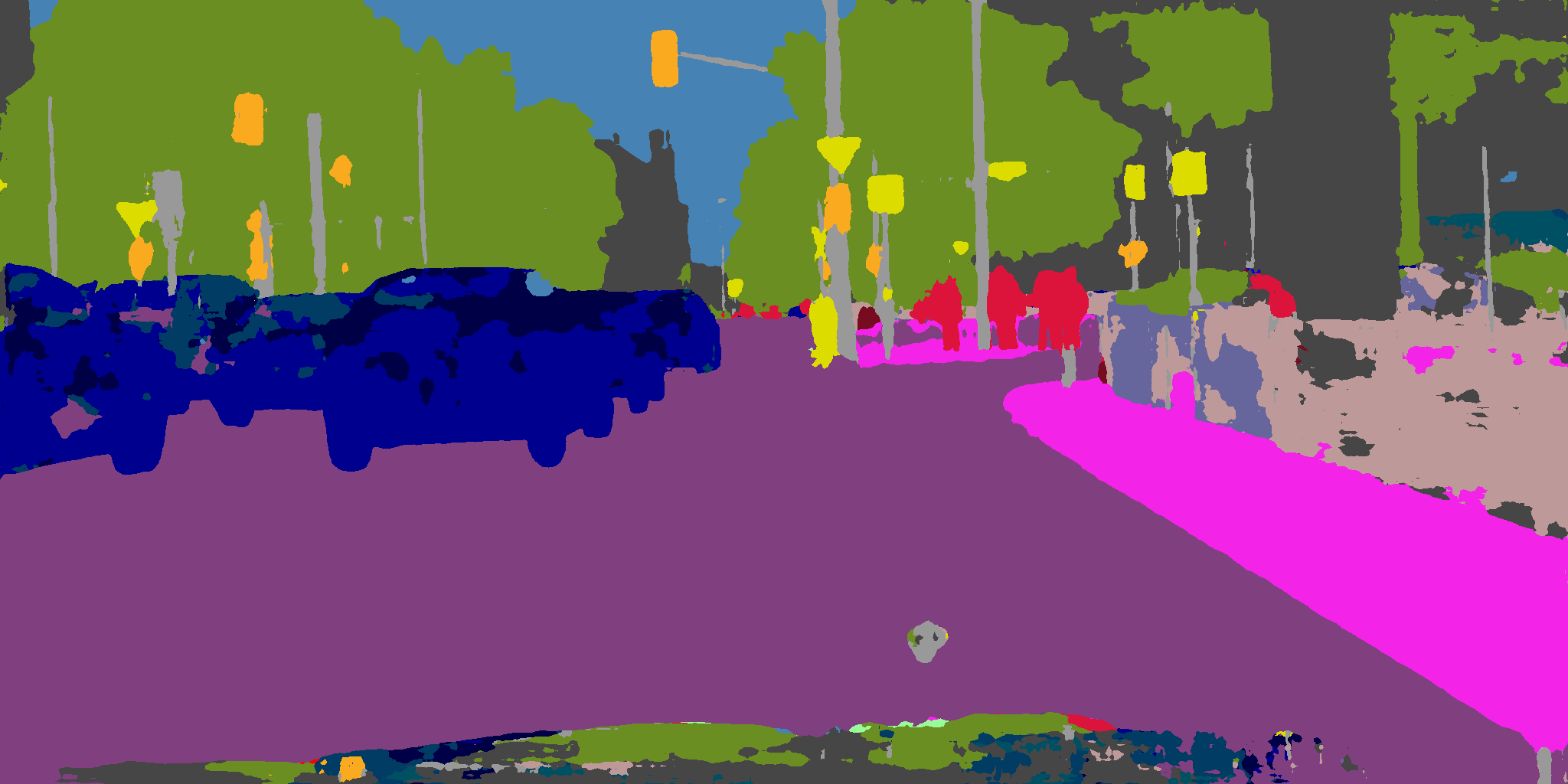}
}
&
\subfloat{
  \includegraphics[valign=m,width=0.23\linewidth]{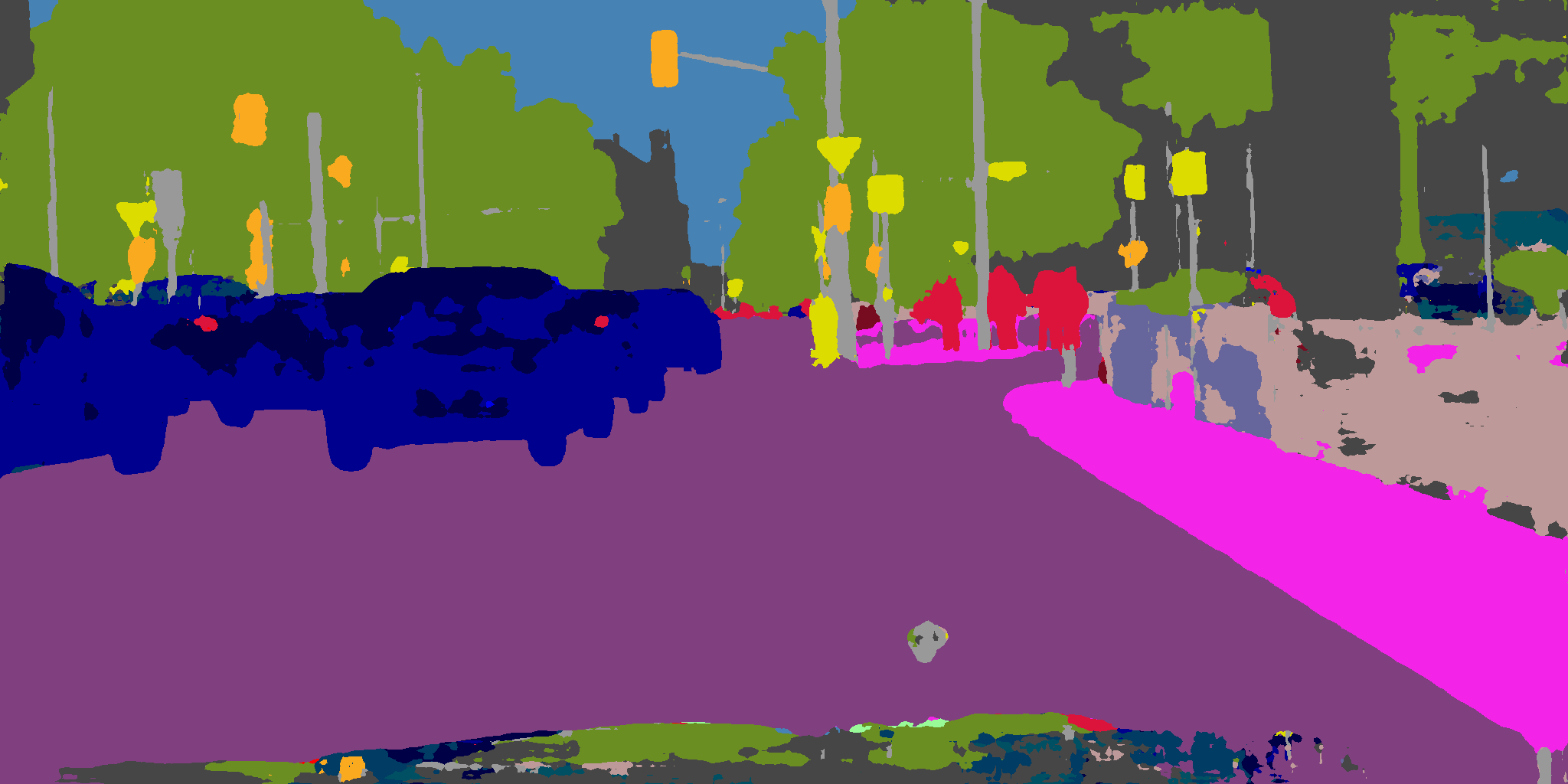}
}
&
\subfloat{
  \includegraphics[valign=m,width=0.23\linewidth]{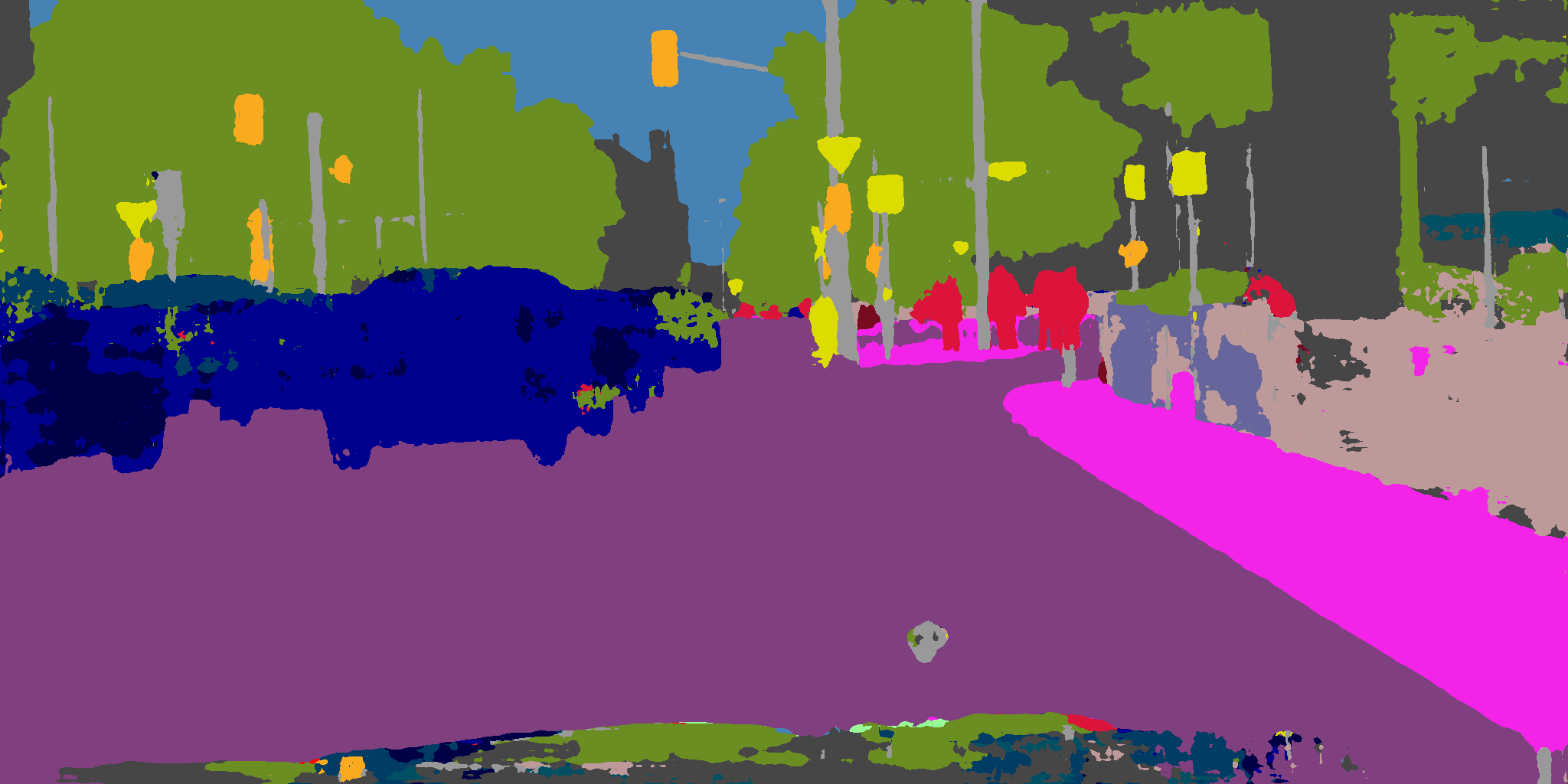}
} 
\\ 
\subfloat{
  \rotatebox[origin=b]{90}{RGB}
  \includegraphics[valign=m,width=0.23\linewidth]{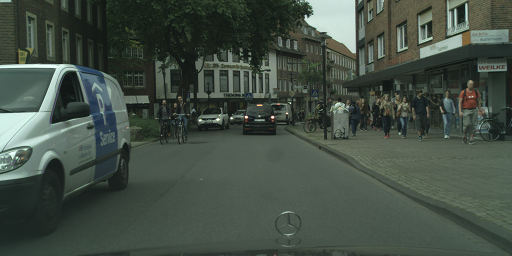}
}
&
\subfloat{
  \includegraphics[valign=m,width=0.23\linewidth]{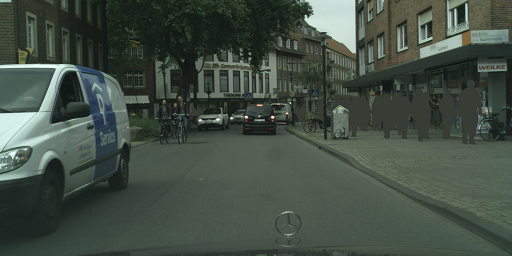}
}
&
\subfloat{
  \includegraphics[valign=m,width=0.23\linewidth]{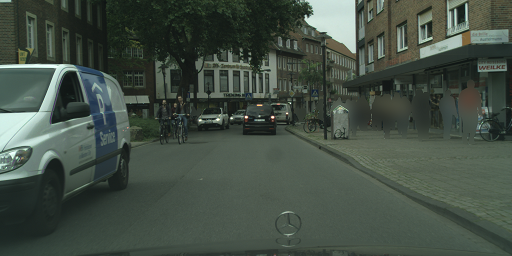}
}
&
\subfloat{
  \includegraphics[valign=m,width=0.23\linewidth]{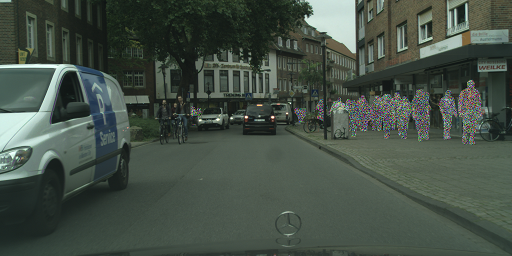}
}
\\ [-2ex]
\subfloat{
  \rotatebox[origin=b]{90}{GT}
  \includegraphics[valign=m,width=0.23\linewidth]{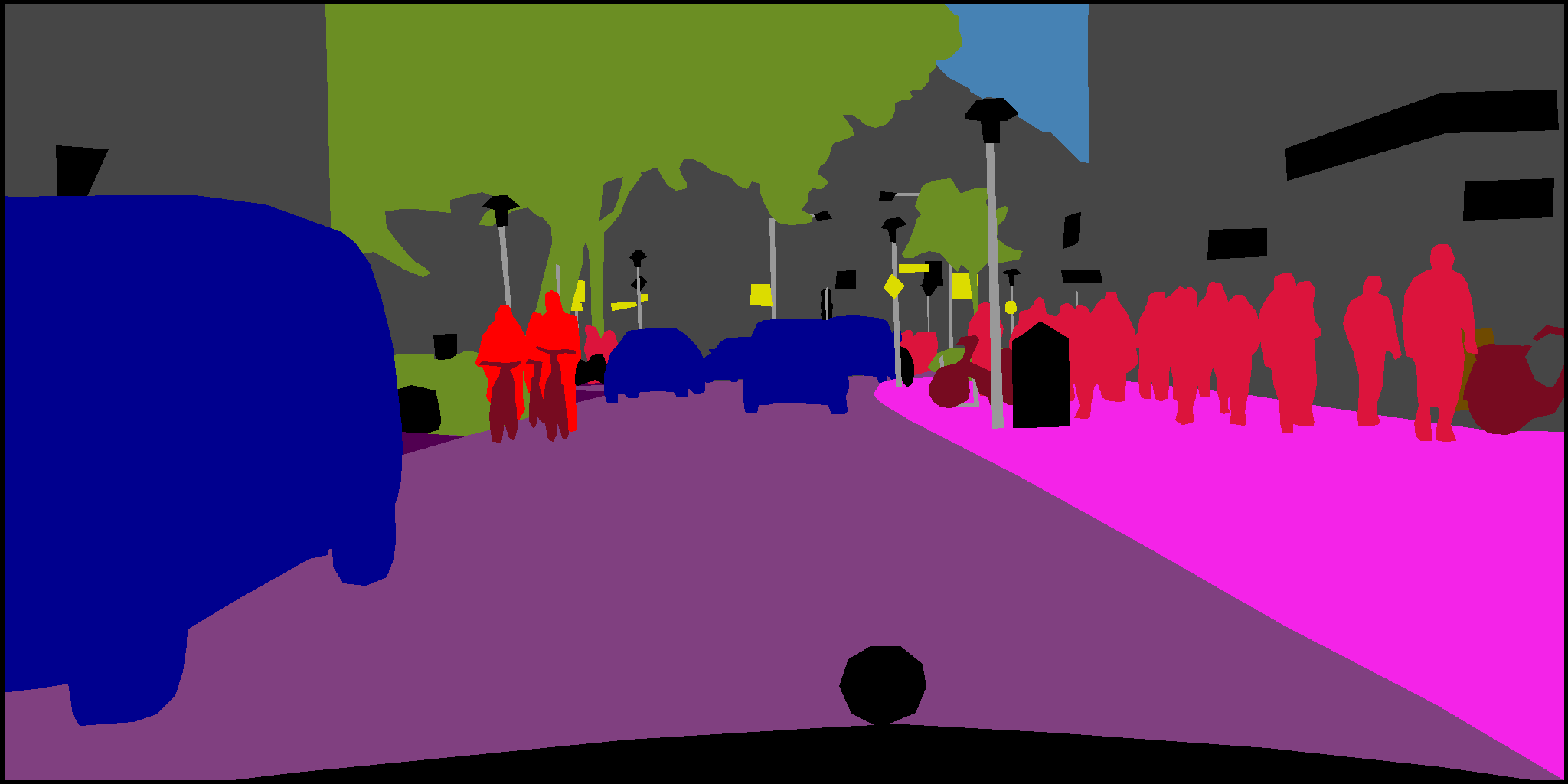}
}
&
\subfloat{
  \includegraphics[valign=m,width=0.23\linewidth]{2gt_pedestrians.png}
}
&
\subfloat{
  \includegraphics[valign=m,width=0.23\linewidth]{2gt_pedestrians.png}
}
&
\subfloat{
  \includegraphics[valign=m,width=0.23\linewidth]{2gt_pedestrians.png}
}
\\ [-2ex]
\subfloat{
  \rotatebox[origin=b]{90}{Reference}
  \includegraphics[valign=m,width=0.23\linewidth]{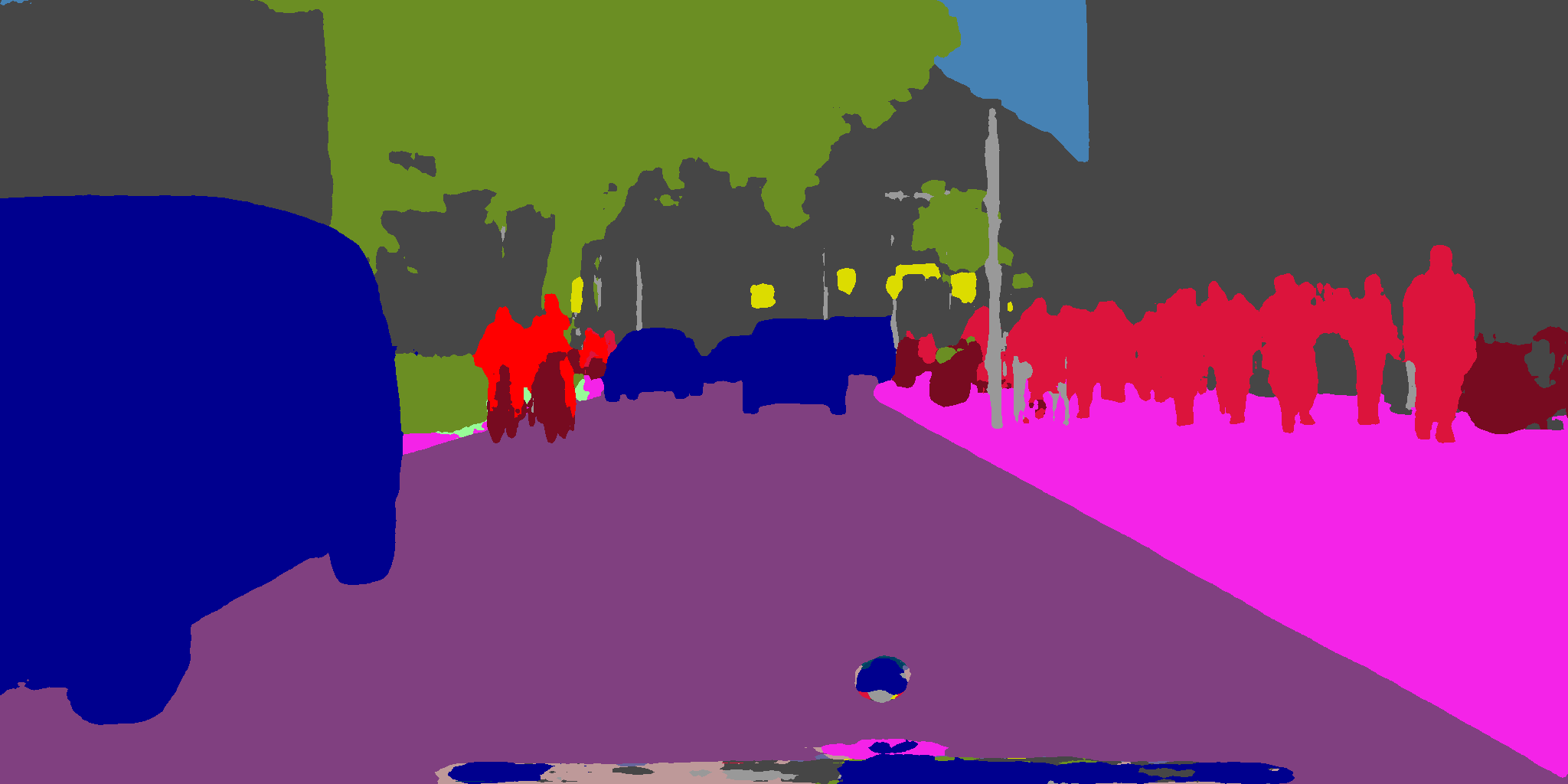}
}
&
\subfloat{
  \includegraphics[valign=m,width=0.23\linewidth]{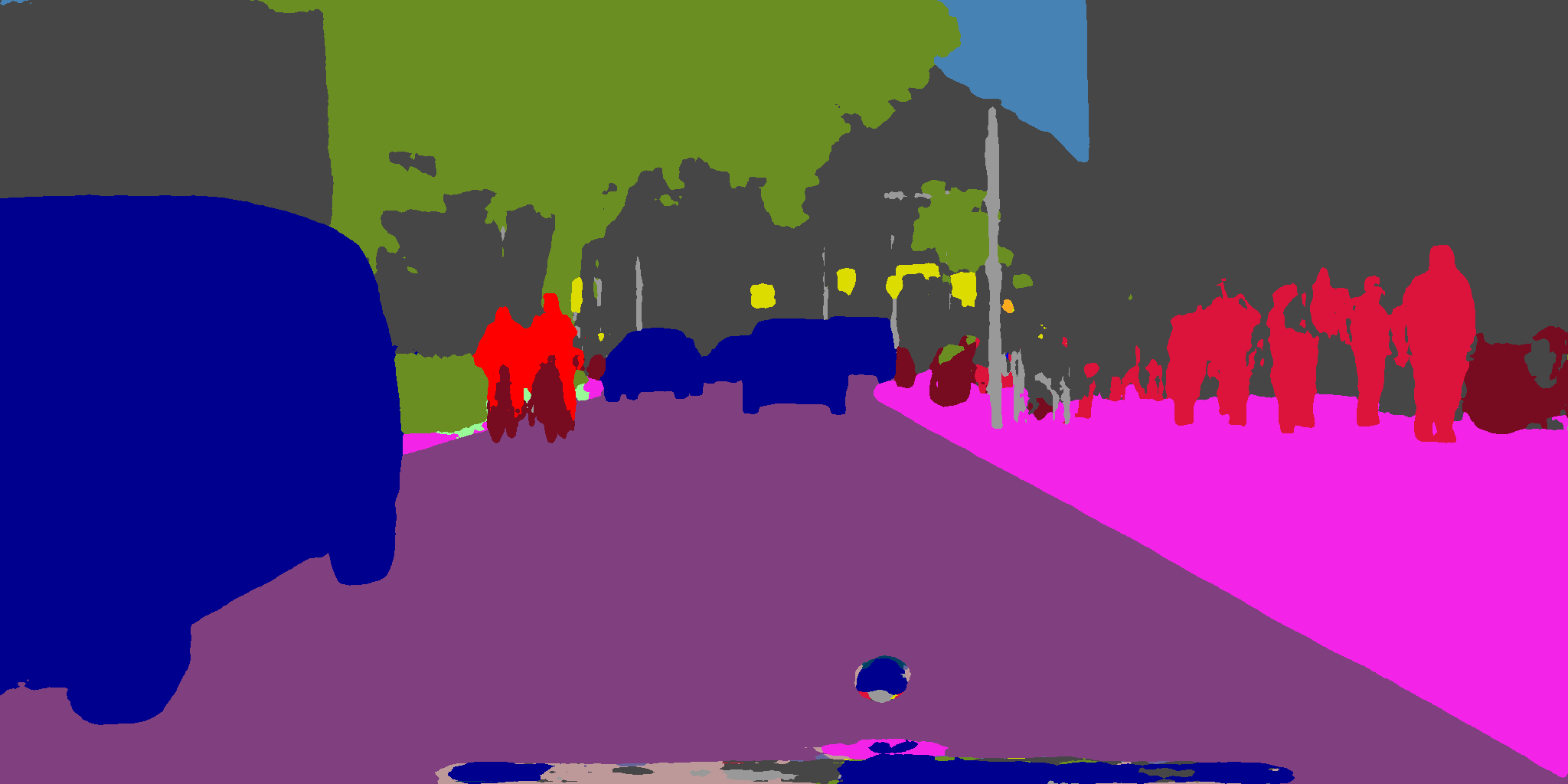}
}
&
\subfloat{
  \includegraphics[valign=m,width=0.23\linewidth]{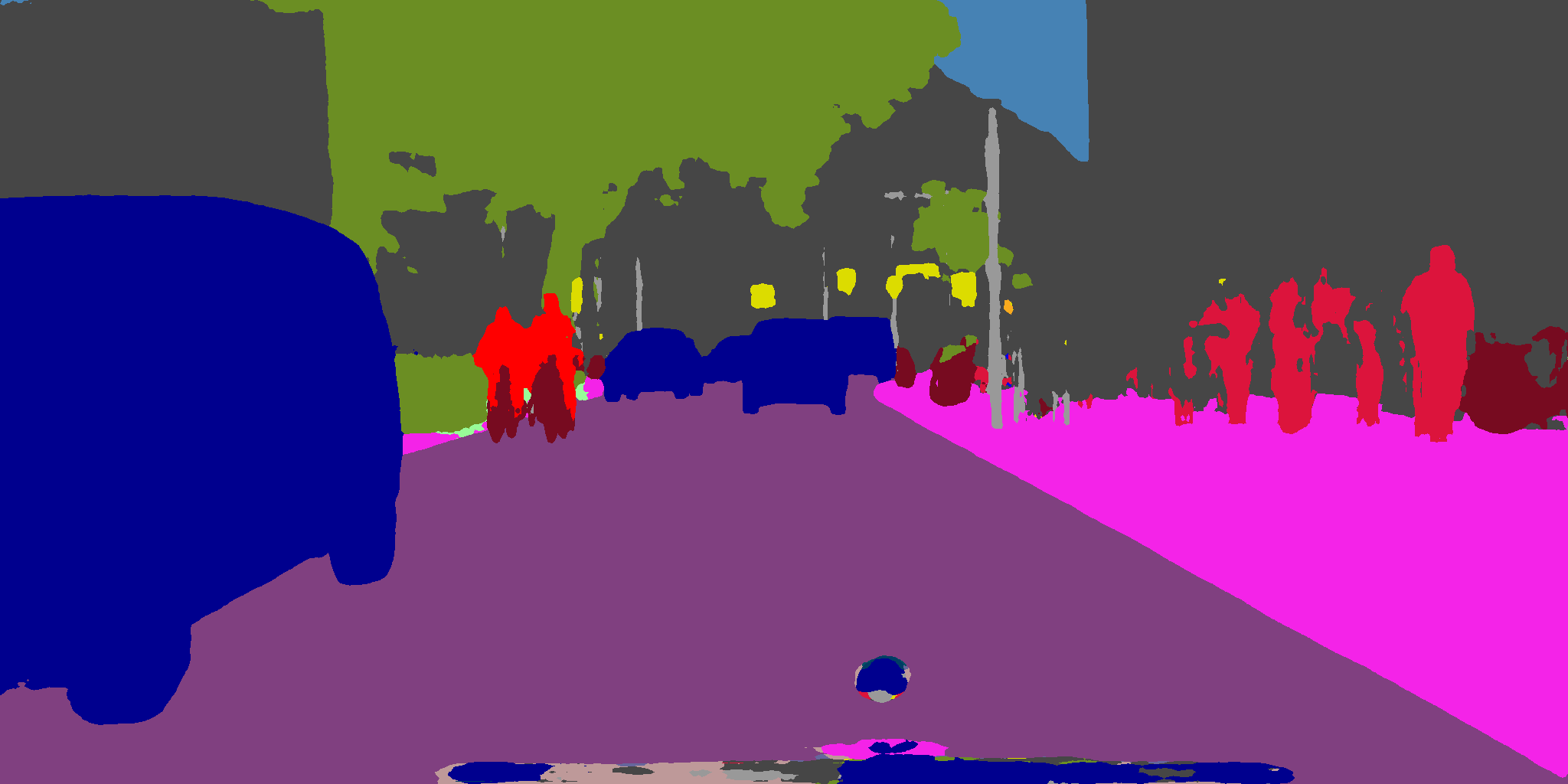}
}
&
\subfloat{
  \includegraphics[valign=m,width=0.23\linewidth]{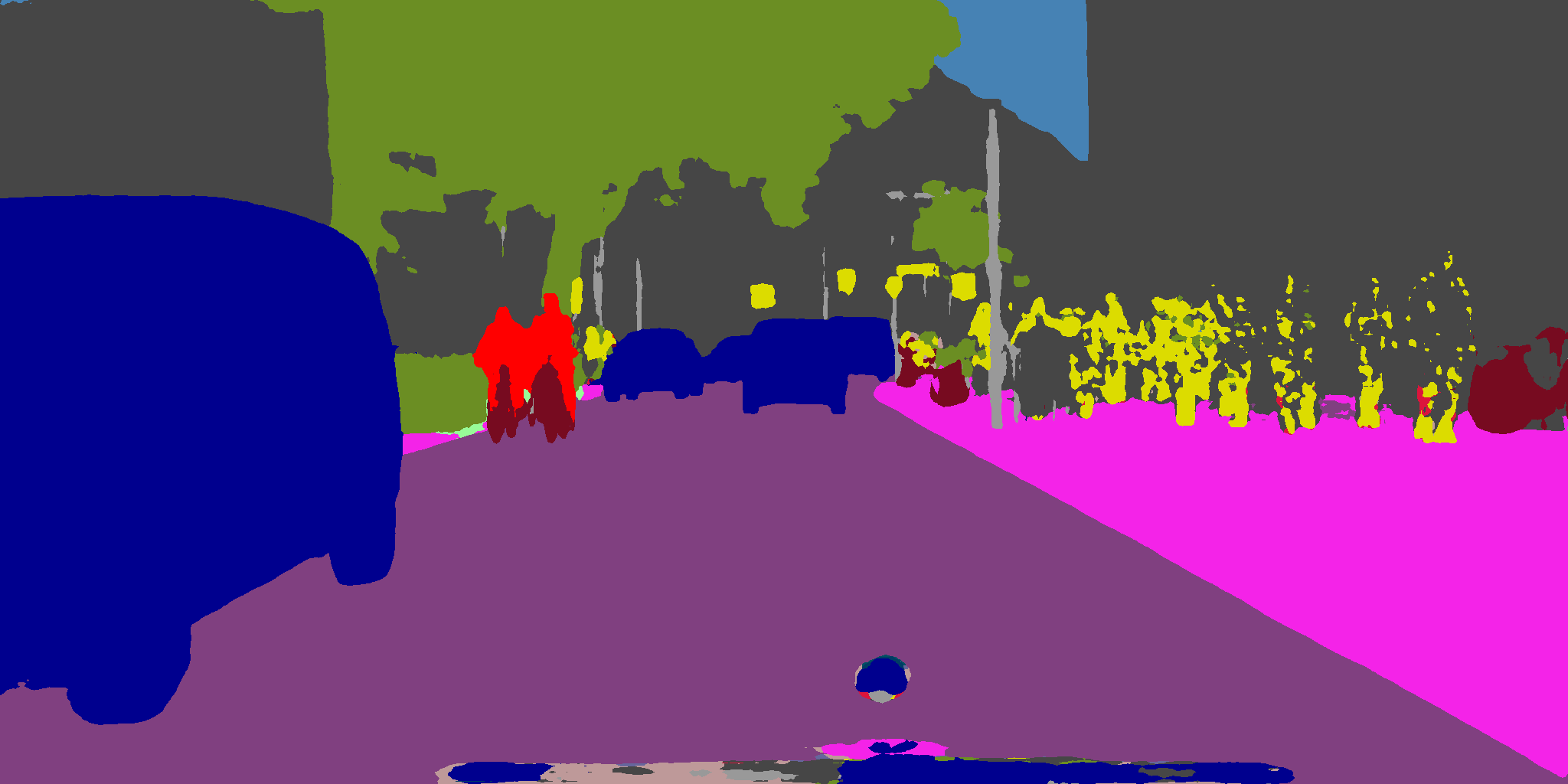}
}
\\ [-2ex]
\subfloat{
  \rotatebox[origin=b]{90}{Ours}
  \includegraphics[valign=m,width=0.23\linewidth]{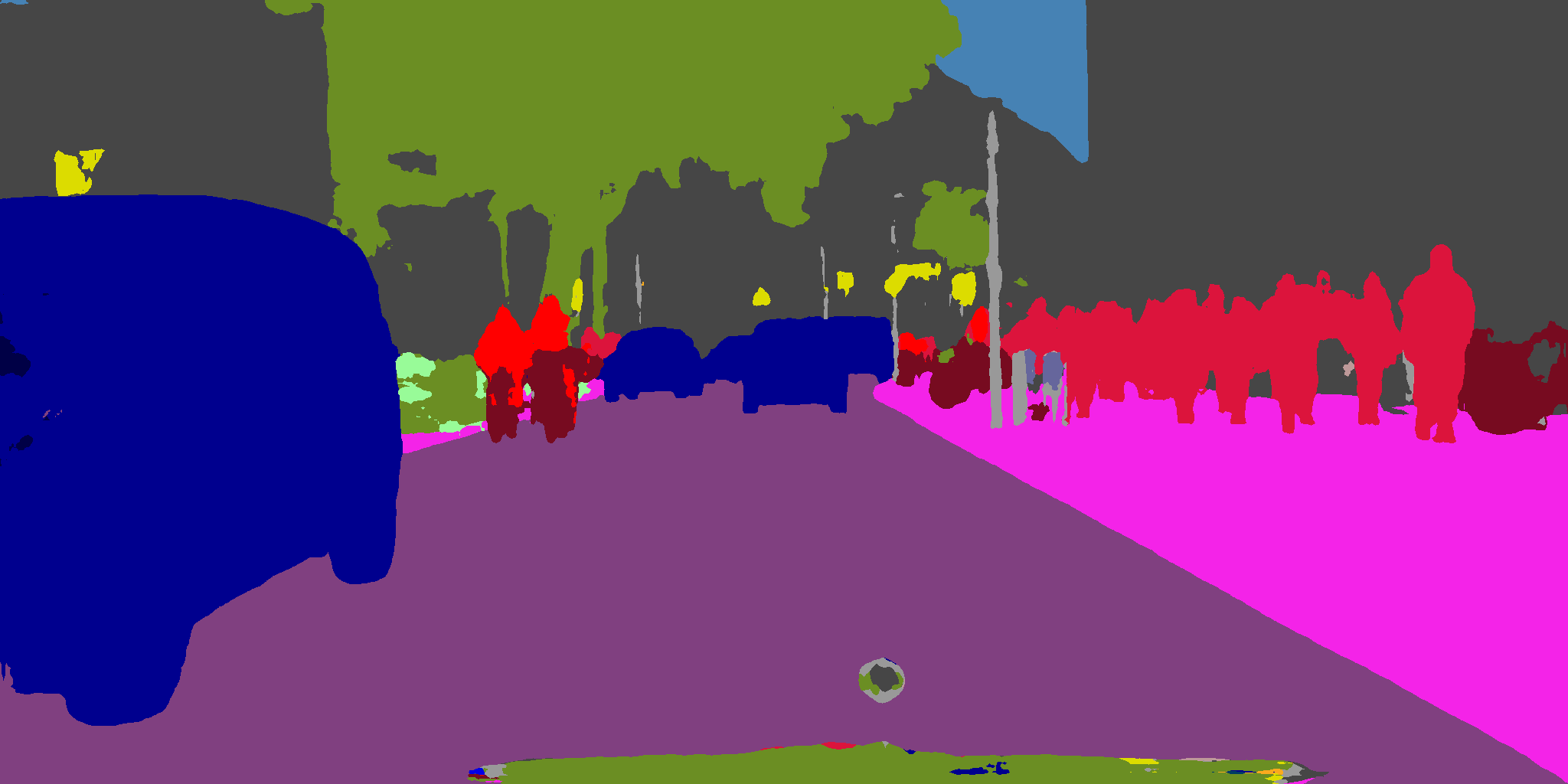}
}
&
\subfloat{
  \includegraphics[valign=m,width=0.23\linewidth]{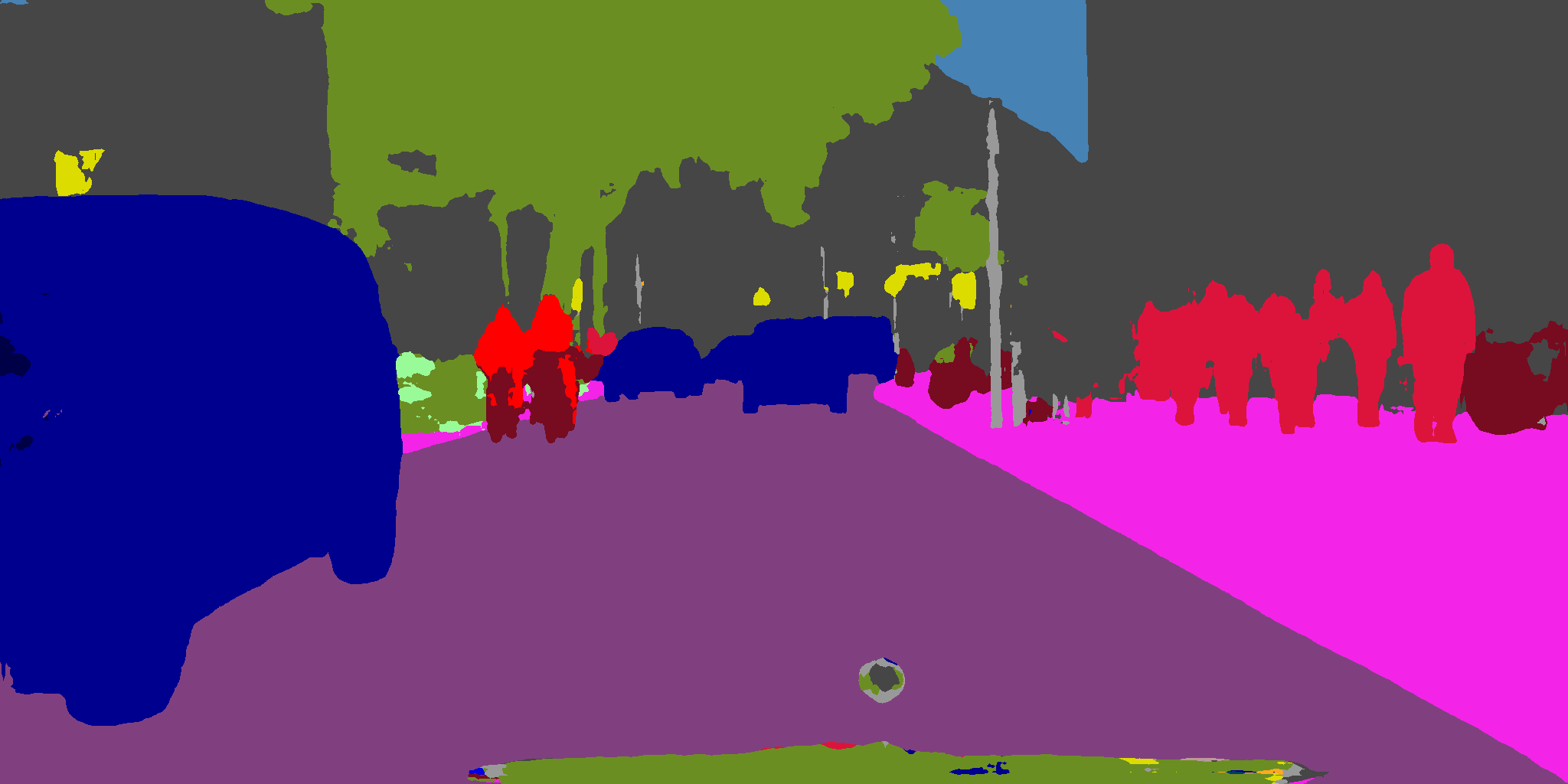}
}
&
\subfloat{
  \includegraphics[valign=m,width=0.23\linewidth]{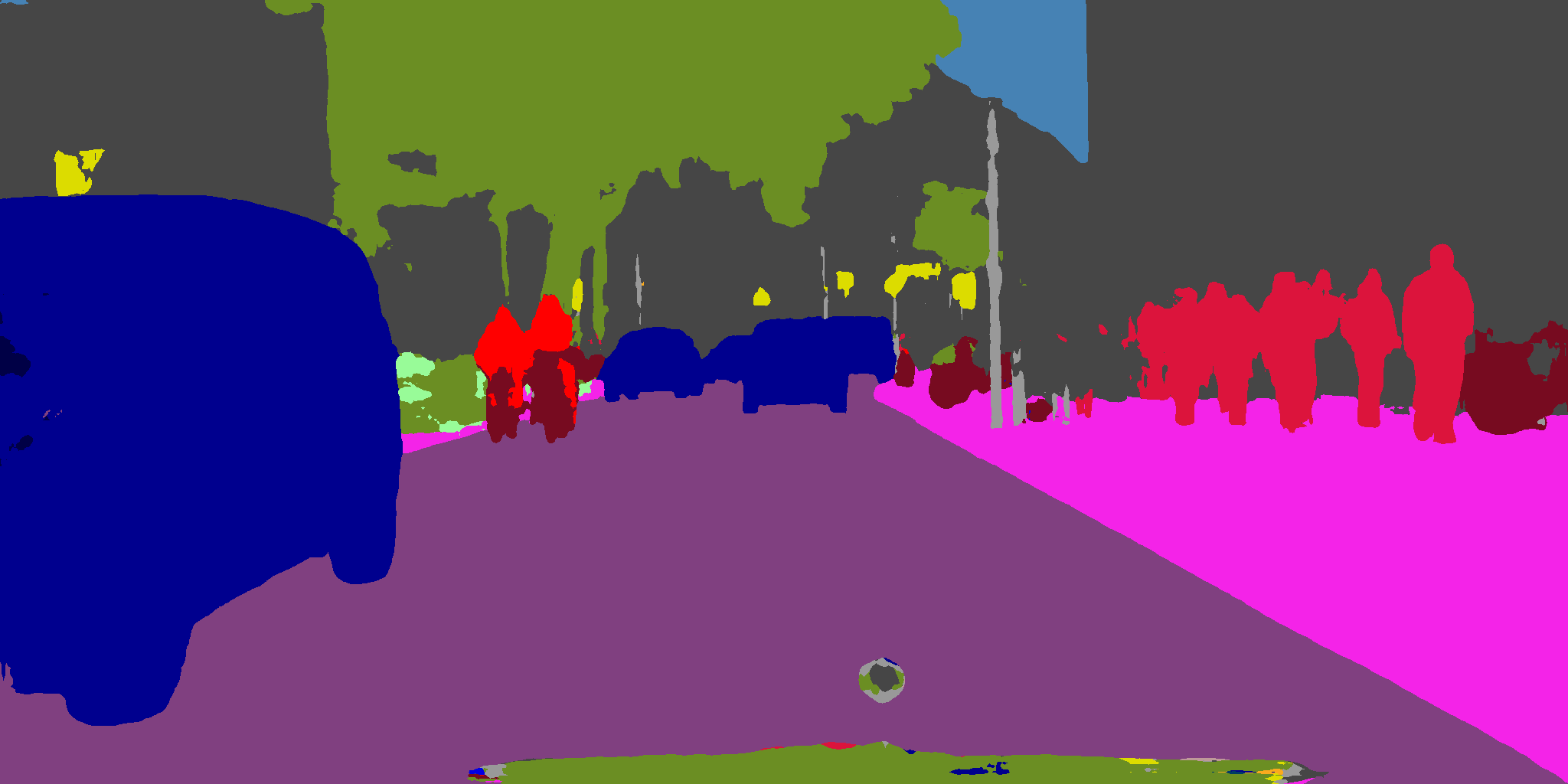}
}
&
\subfloat{
  \includegraphics[valign=m,width=0.23\linewidth]{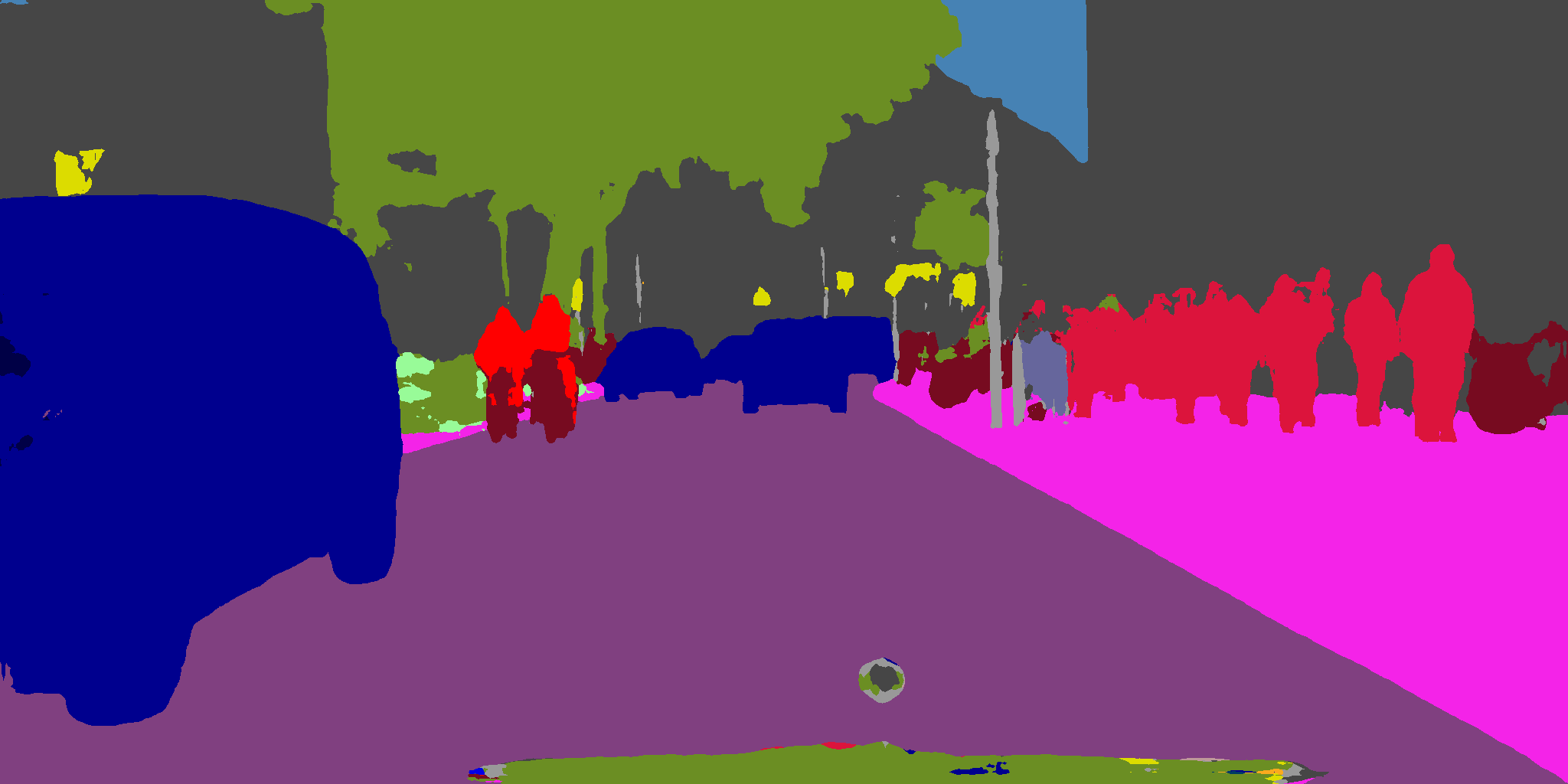}
}

\end{tabular}~
}

\caption{        
Qualitative results of our experiments to understand the effect of Painting-by-Numbers. 
We train the ResNet-50 network backbone on Cityscapes with a standard training schema (i.e., with clean data only, reference model) and with Painting-by-Numbers (our model).
\textbf{(top)} The first row shows the original validation image and the corrupted variants for class \textit{car} and the respective ground truth in the second row.
We replace either the class texture by the dataset-wide RGB-mean, strongly low-pass filtered the class, or added severe Gaussian image noise.
The third row shows the predictions of the reference model.
The fourth row shows the predictions of our model.
The predictions in the fourth row (our model) are superior to the third row (reference model).
Our model is able to withstand the image noise based corruption (last column) for which the reference model confuses \textit{cars} with \textit{traffic signs} mostly.
\textbf{(bottom)} For \textit{persons}, the reference model predicts well, when the RGB-mean replaces the texture of the class.
Both models are relatively robust when the classes are low-pass filtered by severe Gaussian blur.
Similar to the results with respect to class \textit{car}, the reference model struggles to predict well for severe image noise and confuses \textit{persons} also with \textit{traffic signs} mostly
}
\label{fig:validation_qualitive_results}
\end{figure}

% from now on noise and blur
In the presence of severe Gaussian noise, the reference model is struggling to segment classes.
The sensitivity is poor for every class, except for \textit{traffic signs} and \textit{building}.
In the presence of image noise, the reference model tends to segment pixels oftentimes as these very classes, as illustrated in Fig.~\ref{fig:corruptions} and Fig~\ref{fig:validation_qualitive_results}.
The sensitivity scores of our model are often significantly higher.
Similar to the previously discussed results, the sensitivity with respect to ``stuff'' with less texture is often high (e.g., $s_\mathit{ours}=\SI{95.9}{\%}$ for \textit{road}). 
The sensitivity scores are also high for ``things'' such as \textit{persons} and \textit{cars} ($s_\mathit{ours}=\SI{52.7}{\%}$, and $s_\mathit{ours}=\SI{40.6}{\%}$, respectively).
Our model segments many classes well that are corrupted by severe image noise, even though our model has not seen image noise during the training.

The reference model generally performs well when classes are low-pass filtered by severe Gaussian blur.
This result is in accordance with~\cite{kamann_benchmarking_2020}, where the authors found semantic segmentation models to be relatively robust towards image blur.
Again, for class category ``things'', our model outperforms the reference model in most cases.
For example, the sensitivity score of our model for \textit{person}, \textit{rider}, and \textit{car} is by approx. \SI{20.0}{\%} higher.

\textbf{Qualitative results.}
See Fig.~\ref{fig:validation_qualitive_results} for qualitative results of the previously discussed experiments.
Please see the caption of Fig.~\ref{fig:validation_qualitive_results} for discussion.

\section{Conclusions}
\label{sec:conclusion}
We proposed a simple, yet effective, data augmentation schema (Painting-by-Numbers) for semantic image segmentation in this work. 
This training schema increases the robustness for a wealth of common image corruptions in a generic way.
Painting-by-Numbers corrupts training data so that the texture of image classes becomes less reliable, forcing the neural network to develop and increase its shape-bias to segment the image correctly.
Painting-by-Numbers' benefits are that it does not require any additional data, is easy to implement in any supervised segmentation model, and is computationally efficient.
It would be interesting to enforce other network biases, such as context bias or layout bias, and even to combine these with a shape bias, to further increase the robustness of semantic segmentation models with respect to common image corruptions.

\clearpage
\bibliographystyle{splncs04}
\bibliography{literatur}
\clearpage
\setcounter{section}{0}
{\large \textbf{Supplementary Material}}

%\begin{comment}
%\titlerunning{ECCV-20 submission ID \ECCVSubNumber} 
%\authorrunning{ECCV-20 submission ID \ECCVSubNumber} 
%\author{Anonymous ECCV submission}
%\institute{Paper ID \ECCVSubNumber}
%\end{comment}
%******************

%\begin{comment}
%\titlerunning{Increasing Robustness with Painting-by-Numbers}
%\author{Christoph Kamann\inst{1,2}%\orcidID{0000-1111-2222-3333} 
%\and
%Carsten Rother\inst{2}%\orcidID{1111-2222-3333-4444}
%}
%
%\authorrunning{C. Kamann and C. Rother}
%\institute{Corporate Research, Robert Bosch GmbH, Renningen, Germany \\
%\email{christoph.kamann@bosch.com}
%\and
%Visual Learning Lab, Heidelberg University (HCI/IWR), Heidelberg, Germany \\
%\email{carsten.rother@iwr.uni-heidelberg.de} \\
%\url{http://vislearn.de}
%}
%\end{comment}
%******************

\graphicspath{{imgs/}}

\vspace{+1.1cm}
We provide further results and illustrations as mentioned in the main paper.
In more detail, this supplementary material is structured as follows.
We first discuss the computational efficiency of Painting-by-Numbers in section~\ref{sec:secA+} and illustrate the utilized ImageNet-C corruptions, to generate Cityscapes-C, in section~\ref{sec:secA}.
We then discuss in section~\ref{sec:secB} in more detail the effect of several image corruptions of category \textit{digital} and \textit{weather}.
In section~\ref{sec:secC}, we list the individual mIoU scores of Table 1 of the main paper, i.e., the mIoU for each severity level per image corruption.
In section~\ref{sec:secD}, we extend an experiment from the main paper for understanding Painting-by-Numbers.
\section{Computational Efficiency}
\label{sec:secA+}
Painting-by-Numbers is indeed efficient.
We implemented Painting-by-Numbers on CPU using TensorFlow and augmented half of the images of each mini-batch on-the-fly.
For our setup (a machine with 4x 1080 Ti (\SI{11}{GB}), and Intel Xeon CPU E5-2699 with 44 CPU cores with \SI{2.2}{GHz}), the training time with Painting-by-Numbers increases by approx. \SI{2.5}{\%}.
Therefore, it was not necessary to optimize the code in any way.
When implemented on GPU, a network trained with Painting-by-Numbers is approx. \SI{2.0}{\%} slower than training without Painting-by-Numbers.

\section{Cityscapes-C}
\label{sec:secA}
\begin{figure}[ht!]
\centering
\begin{tabular}{c c c }
\centering
\subfloat{
\includegraphics[width=0.3\linewidth]{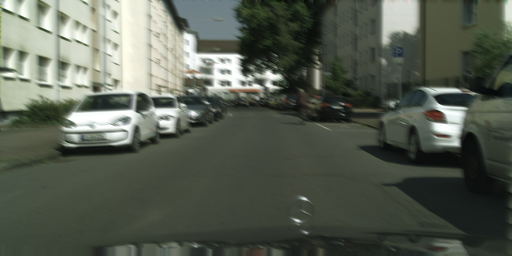}
}
&
\subfloat{
\includegraphics[width=0.3\linewidth]{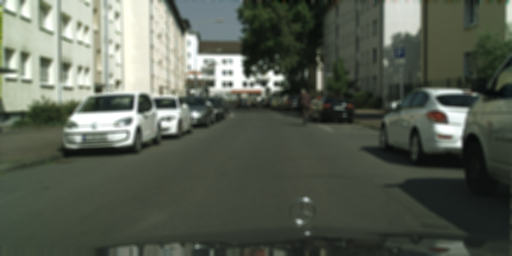}
} 
&
\subfloat{
\includegraphics[width=0.3\linewidth]{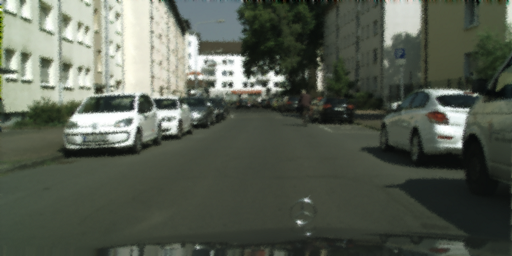}
} \\
\subfloat{
\includegraphics[width=0.3\linewidth]{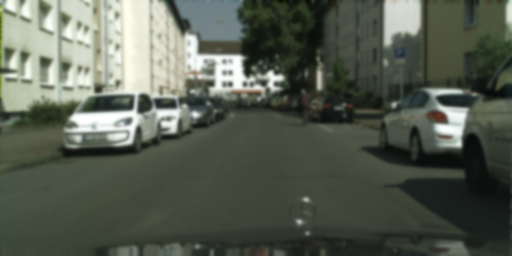}
}
&
\subfloat{
		\includegraphics[width=0.3\linewidth]{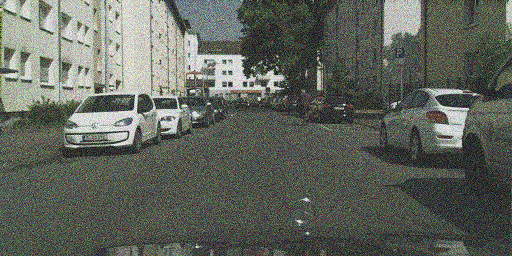}
} 
&
\subfloat{
		\includegraphics[width=0.3\linewidth]{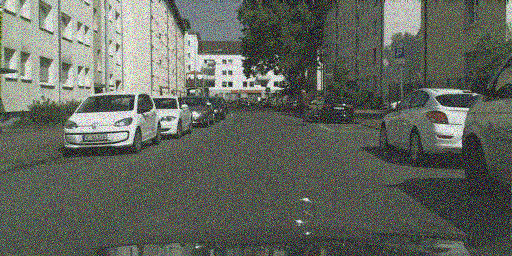}
} \\
\subfloat{
\includegraphics[width=0.3\linewidth]{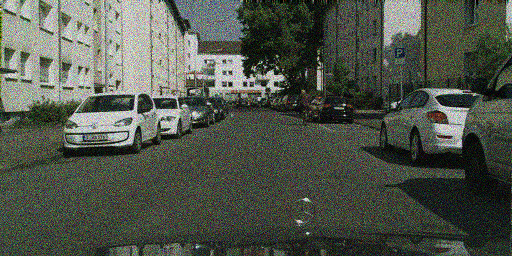}
}
&
\subfloat{
		\includegraphics[width=0.3\linewidth]{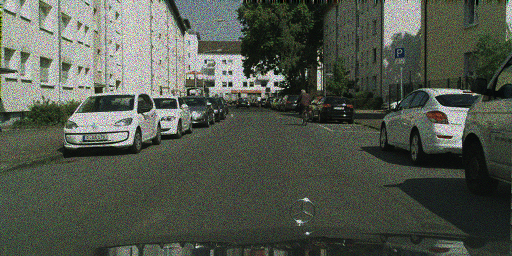}
} 
&
\subfloat{
		\includegraphics[width=0.3\linewidth]{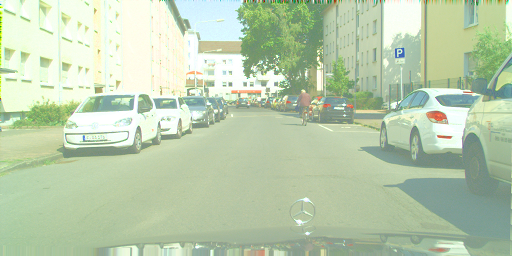}}\\
\subfloat{
		\includegraphics[width=0.3\linewidth]{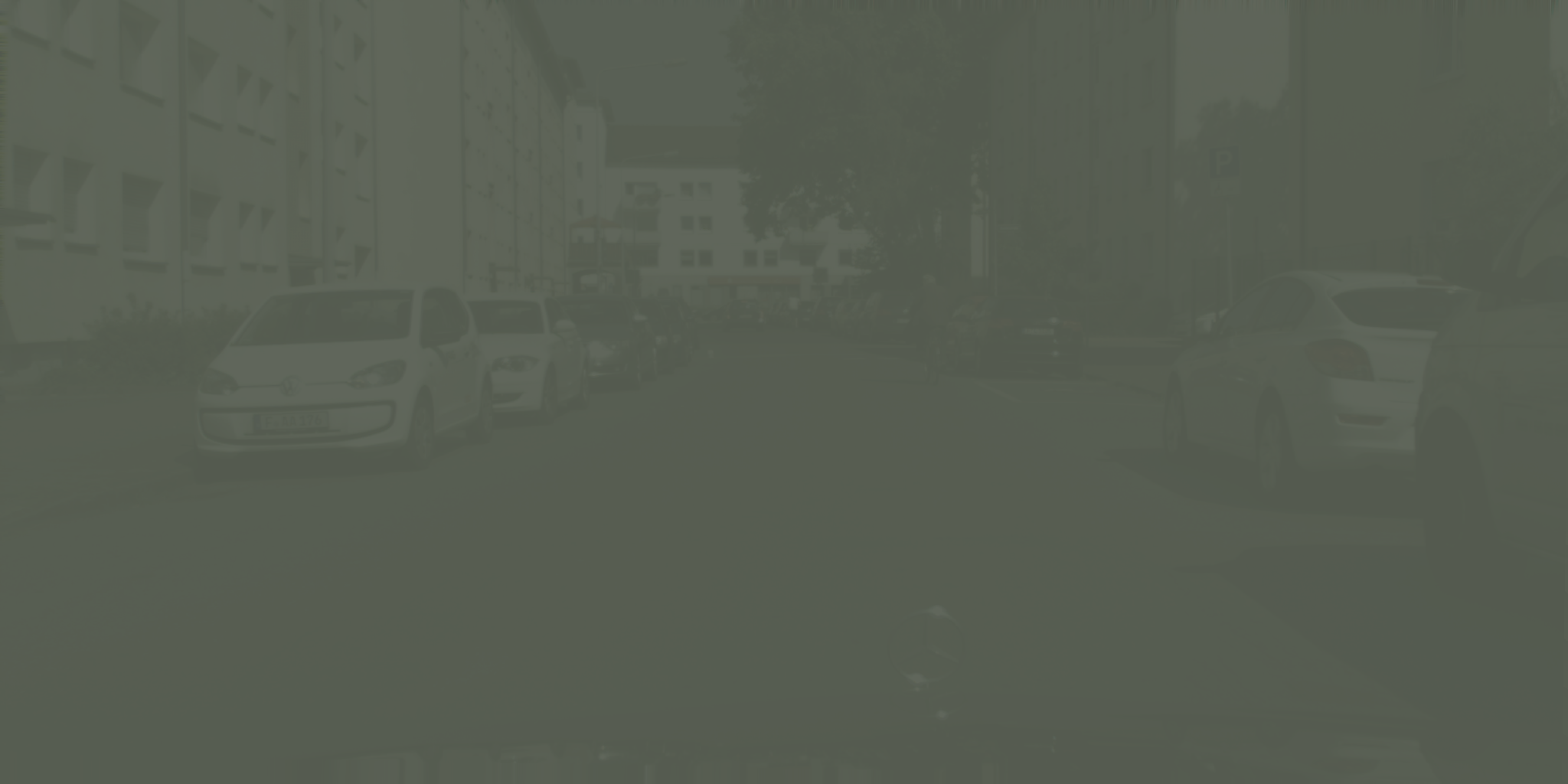}
}
&
\subfloat{
		\includegraphics[width=0.3\linewidth]{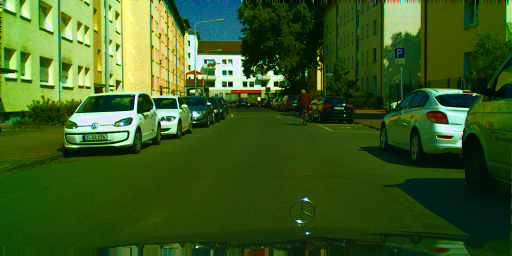}
} 
&
\includegraphics[width=0.3\linewidth]{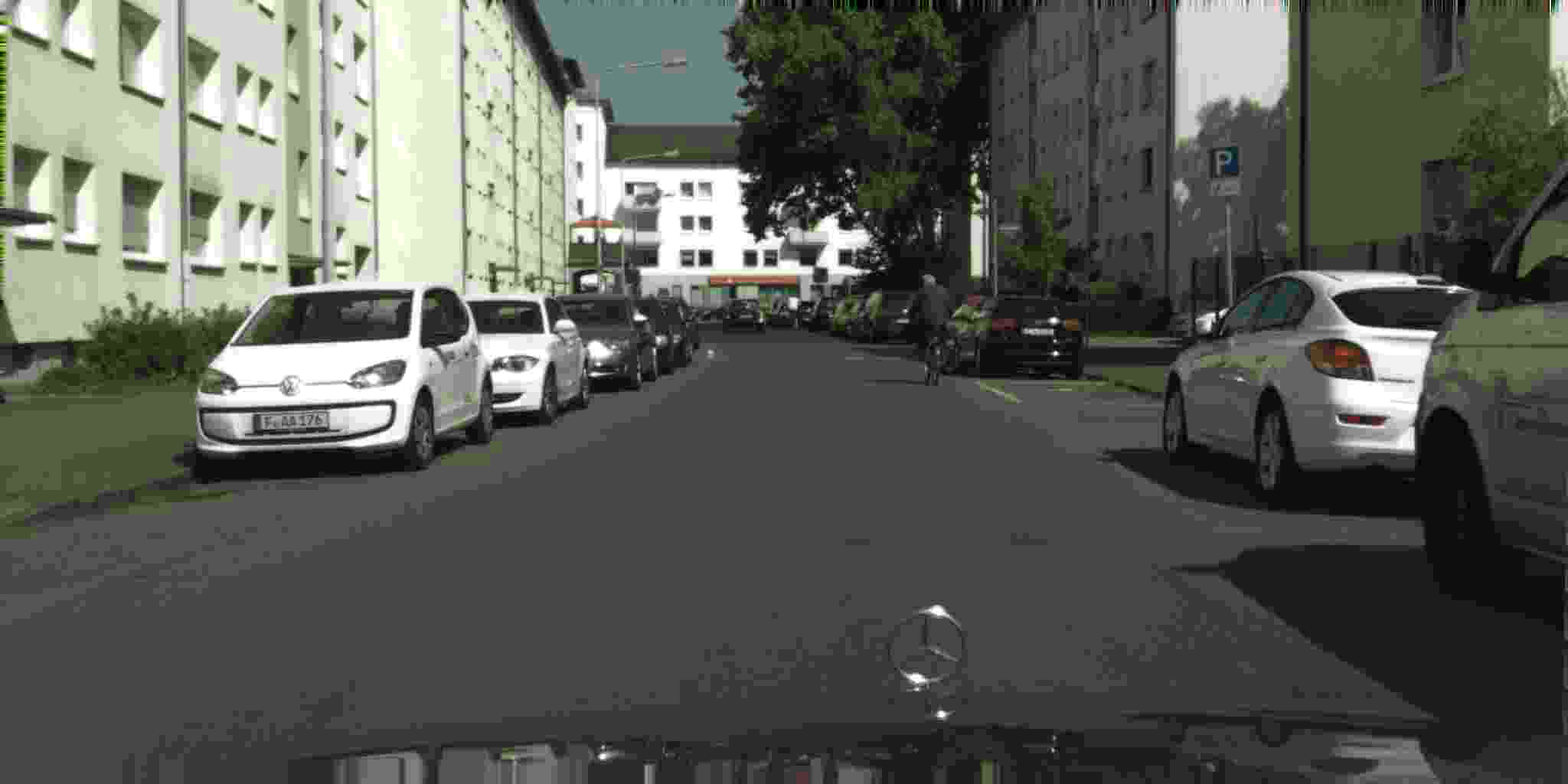}\\

\subfloat{
		\includegraphics[width=0.3\linewidth]{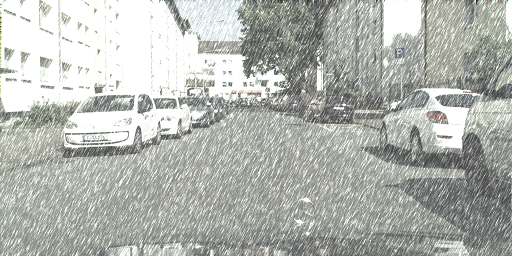}
}
&
\subfloat{
		\includegraphics[width=0.3\linewidth]{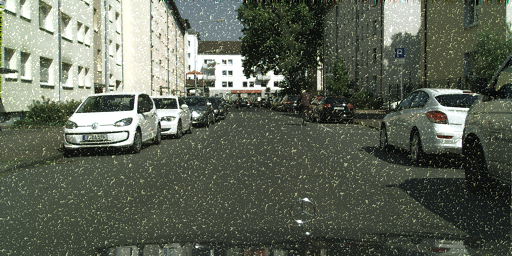}
} 
&\subfloat{
		\includegraphics[width=0.3\linewidth]{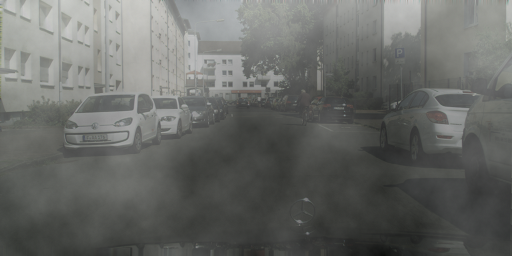}}\\

\multicolumn{3}{l}{
\subfloat{
	\includegraphics[width=0.3\linewidth]{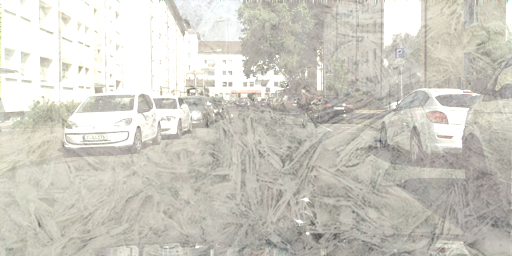}}}
\end{tabular}

\caption{
	Illustration of Cityscapes-C using many transformations of ImageNet-C. 
	First row: Motion blur, defocus blur, frosted glass blur. 
	Second row: Gaussian blur, Gaussian noise, impulse noise. 
	Third row: Shot noise, speckle noise, brightness. 
	Fourth row: Contrast, saturate, JPEG. 
	Fifth row: Snow, spatter, fog. 
	Sixth row: frost}
\label{fig:fig1-supp}    
\end{figure}

\begin{figure}[ht!]
\centering
\begin{tabular}{c c c }
\centering
{ (a) Severity Level 1} & { (b) Severity Level 2} & { (c) Severity Level 3 }\\ 
\subfloat{
\includegraphics[width=0.3\linewidth]{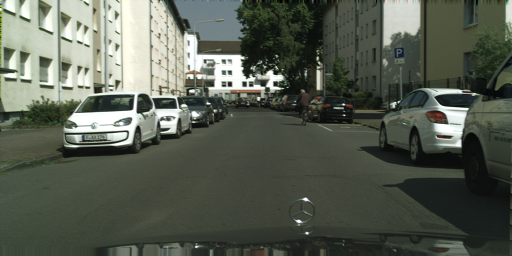}
}
&
\subfloat{
\includegraphics[width=0.3\linewidth]{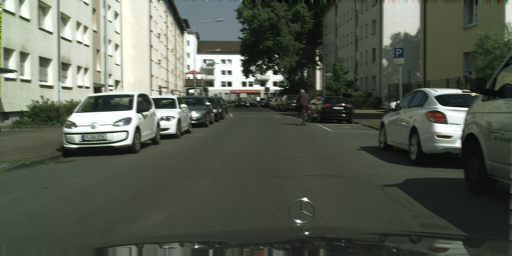}
} 
&
\subfloat{
\includegraphics[width=0.3\linewidth]{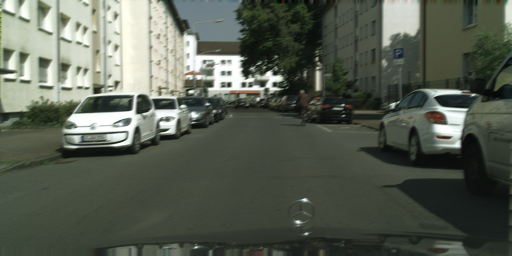}
} \\
\subfloat{
		\includegraphics[width=0.3\linewidth]{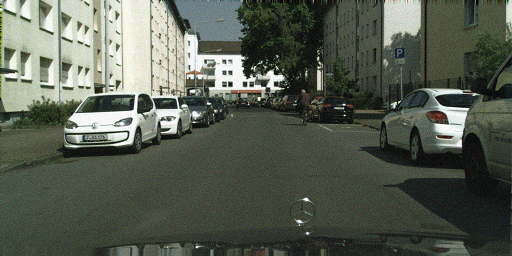}
}
&
\subfloat{
		\includegraphics[width=0.3\linewidth]{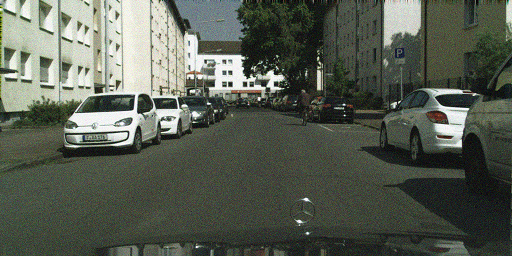}
} 
&
\subfloat{
		\includegraphics[width=0.3\linewidth]{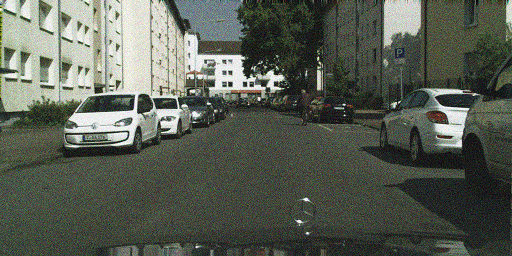}
} \\
\subfloat{
		\includegraphics[width=0.3\linewidth]{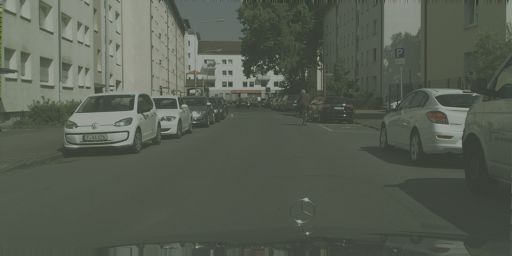}
}
&
\subfloat{
		\includegraphics[width=0.3\linewidth]{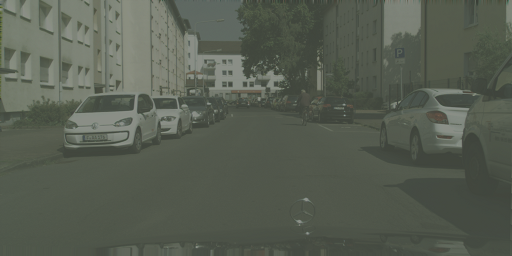}
} 
&
\subfloat{
		\includegraphics[width=0.3\linewidth]{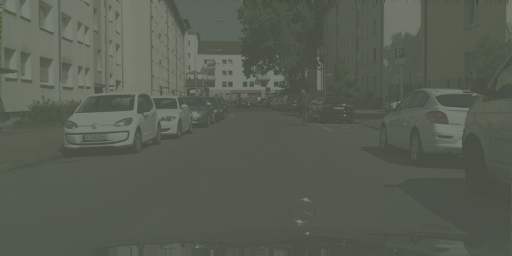}}
		\\
\subfloat{
\includegraphics[width=0.3\linewidth]{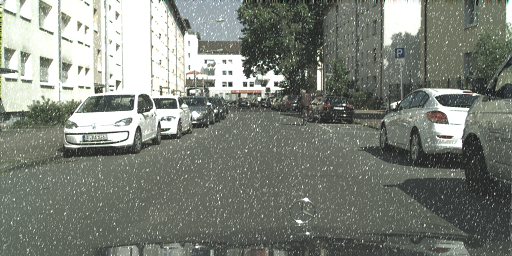}
}
&
\subfloat{
\includegraphics[width=0.3\linewidth]{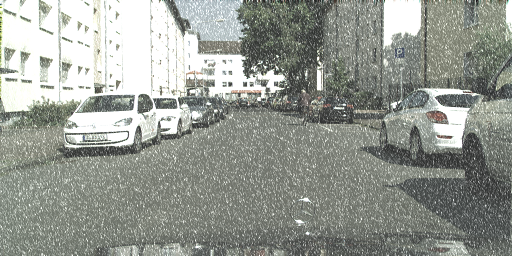}
} 
&
\subfloat{
\includegraphics[width=0.3\linewidth]{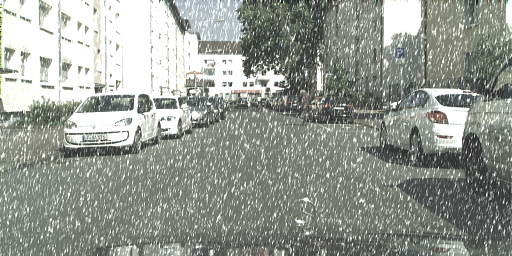}}
\end{tabular}

	\caption{
		Illustration of the first three severity levels of Cityscapes-C for a candidate of the categories \textit{blur}, \textit{noise}, \textit{digital}, and \textit{weather}.
		First row: Motion blur. 
		Second row: Gaussian noise. 
		Third row: Contrast. 
		Fourth row: Snow}
\label{fig:fig2}    
\end{figure}

Fig.~\ref{fig:fig1-supp} illustrates the utilized ImageNet-C\footnote{See \url{https://github.com/hendrycks/robustness/tree/master/ImageNet-C} for the implementation of ImageNet-C.} corruptions to generate Cityscapes-C.
The severity level, i.e., the degree of the respective corruption, for each image is in this Figure mostly four or five.
To illustrate the varying severity levels, we show in Fig.~\ref{fig:fig2} the first three severity levels of a candidate of category \textit{blur}, \textit{noise}, \textit{digital}, and \textit{weather}.

\section{Corruptions of Category \textit{Digital} and \textit{Weather}}
\label{sec:secB}

\begin{figure}[ht]
	\vspace{0cm}
	\centering
	\subfloat[JPEG compressed validation image]{%
      \includegraphics[width=0.48\linewidth]{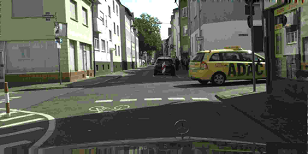}%
    }\hfil
	\subfloat[Zoom of (a)]{%
      \includegraphics[width=0.48\linewidth]{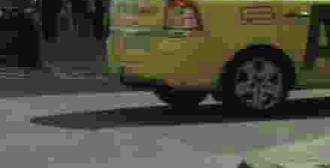}%
    }\hfil
    
	\caption{
		A validation image of Cityscapes corrupted by \textit{JPEG compression} and a respective zoom in (b).
		The crop in (b) visualizes the posterization effect of \textit{JPEG compression}.
		Whereas Painting-by-Numbers alpha-blends the image with a homogeneous, texture-free representation of a class, the JPEG compression causes new, non-distinct shapes within a class
	}
	\label{fig:fig3}    
\end{figure}

The results in the main paper indicate that Painting-by-Numbers does not (compared to a regularly trained network) increase the performance with respect to the corruption \textit{JPEG compression}.
We explain this behavior due to the posterization effect of the JPEG compression algorithm (see Fig.~\ref{fig:fig3}).
The Figure illustrates how the JPEG compression algorithm posterizes larger areas of the \textit{car} and \textit{road}.
This is somewhat contrary to Painting-by-Numbers:
Whereas our training schema (i.e., Painting-by-Numbers) alpha-blends the image with a homogeneous, texture-free representation of a class, the JPEG compression causes new, non-distinct shapes within a class instance, see Fig.~\ref{fig:fig3} b.

Though our model's higher performance with respect to image corruptions of category weather is less than, for example, for image noise, the predictions of a network trained with Painting-by-Numbers are improved for key-classes of category ``things'' such as \textit{cars}, and \textit{persons}.
Table~\ref{tab:iou_weather} lists the individual IoU score for both the reference model (i.e., the network was trained on clean data only) and our model (i.e., the network was trained with Painting-by-Numbers).
We used for both models the ResNet-50 network backbone.

For example, with respect to \textit{spatter}, the IoU score for classes \textit{car}, and \textit{person} is significantly higher by \SI{46.6}{\%}, and \SI{31.3}{\%}.
Whereas our model struggles with respect to several classes for corruption \textit{snow}, its IoU for ``things'' as \textit{person}, and \textit{rider} is higher by more than \SI{7.0}{\%}.
For image corruption \textit{fog}, our model performs better for almost every class.
With respect to frost, the IoU score for \textit{cars} of our model is higher by \SI{8.3}{\%}.

\begin{table}[hbtp]
	\caption{
		IoU for each class of several candidates of category \textit{weather} evaluated on the Cityscapes dataset.
		The IoU score of \textit{spatter} of our model for classes \textit{car} and \textit{person} is significantly higher (\SI{46.6}{\%} and \SI{31.3}{\%}, respectively) than the IoU of the reference model.
		Overall, we see many bold numbers for ``things'' of our Painting-by-Numbers model
	}
	\label{tab:iou_weather}
	\begin{adjustbox}{width=\textwidth}
		\begin{tabular}{@{}ccccccccccccccccccccc@{}}    & \rotatebox[origin=l]{90}{road} & \rotatebox[origin=l]{90}{sidewalk} & \rotatebox[origin=l]{90}{building} & \rotatebox[origin=l]{90}{wall} & \rotatebox[origin=l]{90}{fence} & \rotatebox[origin=l]{90}{pole} & \rotatebox[origin=l]{90}{traffic light} & \rotatebox[origin=l]{90}{traffic sign} & \rotatebox[origin=l]{90}{vegetation} & \rotatebox[origin=l]{90}{terrain} & \rotatebox[origin=l]{90}{sky} & \rotatebox[origin=l]{90}{person} & \rotatebox[origin=l]{90}{rider} & \rotatebox[origin=l]{90}{car} & \rotatebox[origin=l]{90}{truck} & \rotatebox[origin=l]{90}{bus} & \rotatebox[origin=l]{90}{train} & \rotatebox[origin=l]{90}{motorcycle} & \rotatebox[origin=l]{90}{bicycle} & \rotatebox[origin=l]{90}{mean} \\ \midrule
			\textbf{Reference (ResNet-50)} &  &  &  &  &  &  &  &  &  &  &  &  &  &  &  &  &  &  &  & \\
			Snow & \textbf{81.2} & \textbf{16.6} & \textbf{60.0} & 1.4 & \textbf{3.2} & \textbf{21.2} & \textbf{15.9} & \textbf{34.3} & \textbf{40.4} & \textbf{20.6} & \textbf{70.0} & 25.6 & 5.7 & 24.4 & 8.4 & 12.0 & 3.6 & \textbf{3.5} & \textbf{35.6} & \textbf{25.5} \\
			Spatter & 48.5 & 6.5 & 55.2 & \textbf{7.5} & 12.5 & 30.7 & \textbf{35.1} & 30.3 & \textbf{66.1} & 16.3 & 3.9 & 19.4 & 15.1 & 26.7 & 1.8 & 17.9 & 1.2 & 14.6 & 34.6 & 23.4 \\
			Fog & 93.2 & 60.5 & 79.2 & 18.6 & 35.3 & \textbf{46.0} & 40.4 & \textbf{63.9} & 73.8 & 7.8 & 77.9 & 69.2 & 46.7 & 85.6 & 52.3 & 68.8 & 47.1 & 45.3 & 66.4 & 56.7 \\
			Frost & \textbf{54.7} & \textbf{12.7} & 38.7 & 1.0 & \textbf{13.1} & 13.6 & 11.0 & 38.5 & 41.8 & \textbf{12.2} & 40.8 & 21.1 & 8.1 & 32.3 & 7.0 & 8.6 & 3.9 & 0.1 & 23.6 & 20.1 \\ \midrule
			\textbf{Our (ResNet-50)} &  &  &  &  &  &  &  &  &  &  &  &  &  &  &  &  &  &  &  & \\
			Snow & 59.8 & 4.3 & 47.5 & \textbf{1.5} & 2.8 & 10.5 & 11.5 & 29.7 & 26.3 & 9.8 & 67.3 & \textbf{32.7} & \textbf{8.4} & \textbf{32.1} & \textbf{12.0} & \textbf{15.7} & \textbf{7.3} & 1.6 & 30.1 & 21.6 \\
			Spatter & \textbf{79.9} & \textbf{24.9} & \textbf{69.3} & 5.9 & \textbf{27.1} & \textbf{36.8} & 33.7 & \textbf{39.4} & 61.3 & \textbf{24.7} & \textbf{42.6} & \textbf{50.7} & \textbf{23.1} & \textbf{73.3} & \textbf{20.2} & \textbf{30.1} & \textbf{7.2} & \textbf{19.6} & \textbf{50.0} & \textbf{37.9} \\
			Fog & \textbf{95.7} & \textbf{70.5} & \textbf{84.8} & \textbf{32.3} & \textbf{46.6} & 44.0 & \textbf{47.4} & 62.9 & \textbf{84.8} & \textbf{33.0} & \textbf{87.8} & \textbf{70.5} & \textbf{50.7} & \textbf{90.6} & \textbf{62.1} & \textbf{79.9} & \textbf{68.6} & \textbf{51.5} & \textbf{67.6} & \textbf{64.8} \\
			Frost & 51.6 & 10.9 & \textbf{49.8} & \textbf{2.0} & 11.8 & \textbf{15.4} & \textbf{15.4} & \textbf{42.5} & \textbf{50.9} & 8.2 & \textbf{58.3} & \textbf{24.2} & \textbf{13.7} & \textbf{40.6} & \textbf{8.1} & \textbf{15.7} & \textbf{9.3} & \textbf{1.8} & \textbf{35.4} & \textbf{24.5}
		\end{tabular}
	\end{adjustbox}
\end{table}

\section{Detailed Quantitative Results}
\label{sec:secC}

Each image corruption of Cityscapes-C is parameterized in five severity levels.
Table 1 from the main paper contains the average score for these five severity levels\footnote{Except for image noise, where we only considered severity levels that the Signal-to-Noise ratio is larger than 10}.     
For completeness, we provide in this section the mIoU score for each severity level.
Please find the mIoU scores for severity level 1$-$5 in Table~\ref{tab:sl1} $-$ Table~\ref{tab:sl5}, respectively.

\begin{table}[hbtp]
	\caption{
		Results on the Cityscapes dataset. 
		Each entry shows the mean IoU of several corrupted variants of the Cityscapes dataset for \textbf{severity level 1}.
		The higher mIoU of either the reference model or the respective model trained with Painting-by-Numbers is bold.
		Overall, we see many more bold numbers for our Painting-by-Numbers model
	}
	\label{tab:sl1}
	\begin{adjustbox}{width=\textwidth}
		\begin{tabular}{@{}ccccccccccccccccc@{}}
			\toprule
			& \multicolumn{4}{c}{\textbf{Blur}} & \multicolumn{4}{c}{\textbf{Noise}} & \multicolumn{4}{c}{\textbf{Digital}} & \multicolumn{4}{c}{\textbf{Weather}} \\ \midrule
			\multicolumn{1}{c|}{Network} & Motion & Defocus & \begin{tabular}[c]{@{}c@{}}Frosted \\ Glass\end{tabular} & \multicolumn{1}{c|}{Gaussian} & Gaussian & Impulse & Shot & \multicolumn{1}{c|}{Speckle} & Brightness & Contrast & Saturate & \multicolumn{1}{c|}{JPEG} & Snow & Spatter & Fog & Frost \\ \midrule
			\multicolumn{1}{c|}{\textbf{Reference}} &  &  &  & \multicolumn{1}{c|}{} &  &  &  & \multicolumn{1}{c|}{} &  &  &  & \multicolumn{1}{c|}{} &  &  &  &  \\
			\multicolumn{1}{c|}{MobileNet-V2} & \textbf{69.0} & \textbf{68.4} & \textbf{65.8} & \multicolumn{1}{c|}{\textbf{71.6}} & 9.6 & 14.2 & 11.1 & \multicolumn{1}{c|}{32.1} & 68.4 & 66.5 & 54.8 & \multicolumn{1}{c|}{\textbf{34.7}} & 22.5 & \textbf{71.9} & 56.0 & 36.1 \\
			\multicolumn{1}{c|}{ResNet-50} & \textbf{72.8} & \textbf{72.4} & 69.3 & \multicolumn{1}{c|}{\textbf{75.1}} & 10.7 & 13.4 & 18.8 & \multicolumn{1}{c|}{44.2} & 73.7 & 72.1 & 66.6 & \multicolumn{1}{c|}{\textbf{44.5}} & \textbf{25.5} & \textbf{76.7} & 63.9 & 43.7 \\
			\multicolumn{1}{c|}{ResNet-101} & 71.6 & 70.6 & 69.3 & \multicolumn{1}{c|}{73.1} & 22.9 & 22.9 & 30.9 & \multicolumn{1}{c|}{50.8} & 72.9 & 71.9 & 67.4 & \multicolumn{1}{c|}{\textbf{52.4}} & 24.3 & 75.9 & 64.1 & 46.9 \\
			\multicolumn{1}{c|}{Xception-41} & 73.6 & 72.5 & 70.2 & \multicolumn{1}{c|}{75.2} & 27.6 & 25.1 & 38.1 & \multicolumn{1}{c|}{57.5} & 75.7 & 73.2 & 71.3 & \multicolumn{1}{c|}{\textbf{60.2}} & \textbf{41.1} & \textbf{76.8} & 66.9 & 45.9 \\
			\multicolumn{1}{c|}{Xception-71} & 74.2 & 72.8 & 70.6 & \multicolumn{1}{c|}{75.4} & 22.0 & 11.5 & 32.3 & \multicolumn{1}{c|}{54.4} & 76.3 & 73.9 & 71.2 & \multicolumn{1}{c|}{\textbf{56.9}} & 36.7 & 76.7 & 70.2 & 46.6 \\ \midrule
			\multicolumn{1}{c|}{\textbf{Painting-by-Numbers}} &  &  &  & \multicolumn{1}{c|}{} &  &  &  & \multicolumn{1}{c|}{} &  &  &  & \multicolumn{1}{c|}{} &  &  &  &  \\
			\multicolumn{1}{c|}{MobileNet-V2} & 67.7 & 66.9 & 64.2 & \multicolumn{1}{c|}{70.2} & \textbf{17.4} & \textbf{18.4} & \textbf{21.3} & \multicolumn{1}{c|}{\textbf{40.7}} & \textbf{70.0} & \textbf{68.6} & \textbf{62.8} & \multicolumn{1}{c|}{29.8} & \textbf{25.9} & 71.0 & \textbf{66.4} & \textbf{42.9} \\
			\multicolumn{1}{c|}{ResNet-50} & 72.3 & 72.3 & \textbf{72.4} & \multicolumn{1}{c|}{74.3} & \textbf{35.7} & \textbf{34.3} & \textbf{44.4} & \multicolumn{1}{c|}{\textbf{60.7}} & \textbf{74.5} & \textbf{73.5} & \textbf{71.7} & \multicolumn{1}{c|}{38.3} & 21.6 & 75.4 & \textbf{69.6} & \textbf{47.4} \\
			\multicolumn{1}{c|}{ResNet-101} & \textbf{73.1} & \textbf{72.7} & \textbf{71.8} & \multicolumn{1}{c|}{\textbf{74.8}} & \textbf{36.5} & \textbf{39.6} & \textbf{46.2} & \multicolumn{1}{c|}{\textbf{63.1}} & \textbf{75.7} & \textbf{74.3} & \textbf{72.8} & \multicolumn{1}{c|}{48.6} & \textbf{24.5} & \textbf{76.0} & \textbf{71.0} & \textbf{50.4} \\
			\multicolumn{1}{c|}{Xception-41} & \textbf{74.2} & \textbf{73.4} & \textbf{72.1} & \multicolumn{1}{c|}{\textbf{75.2}} & \textbf{46.2} & \textbf{39.9} & \textbf{53.1} & \multicolumn{1}{c|}{\textbf{64.8}} & \textbf{76.8} & \textbf{74.6} & \textbf{73.6} & \multicolumn{1}{c|}{39.0} & 38.2 & 76.7 & \textbf{71.9} & \textbf{46.9} \\
			\multicolumn{1}{c|}{Xception-71} & \textbf{75.6} & \textbf{75.3} & \textbf{73.6} & \multicolumn{1}{c|}{\textbf{76.7}} & \textbf{35.5} & \textbf{38.4} & \textbf{45.3} & \multicolumn{1}{c|}{\textbf{61.7}} & \textbf{78.2} & \textbf{76.2} & \textbf{76.2} & \multicolumn{1}{c|}{43.0} & \textbf{41.6} & \textbf{78.2} & \textbf{74.6} & \textbf{51.8} \\ \bottomrule
		\end{tabular}
	\end{adjustbox}
\end{table}

\begin{table}[hbtp]
	\caption{
		Results on the Cityscapes dataset. 
		Each entry shows the mean IoU of several corrupted variants of the Cityscapes dataset for \textbf{severity level 2}.
		The higher mIoU of either the reference model or the respective model trained with Painting-by-Numbers is bold.
		Overall, we see many more bold numbers for our Painting-by-Numbers model    
	}
	\label{tab:sl2}
	\begin{adjustbox}{width=\textwidth}
		\begin{tabular}{@{}ccccccccccccccccc@{}}
			\toprule
			& \multicolumn{4}{c}{\textbf{Blur}} & \multicolumn{4}{c}{\textbf{Noise}} & \multicolumn{4}{c}{\textbf{Digital}} & \multicolumn{4}{c}{\textbf{Weather}} \\ \midrule
			\multicolumn{1}{c|}{Network} & Motion & Defocus & \begin{tabular}[c]{@{}c@{}}Frosted \\ Glass\end{tabular} & \multicolumn{1}{c|}{Gaussian} & Gaussian & Impulse & Shot & \multicolumn{1}{c|}{Speckle} & Brightness & Contrast & Saturate & \multicolumn{1}{c|}{JPEG} & Snow & Spatter & Fog & Frost \\ \midrule
			\multicolumn{1}{c|}{\textbf{Reference}} &  &  &  & \multicolumn{1}{c|}{} &  &  &  & \multicolumn{1}{c|}{} &  &  &  & \multicolumn{1}{c|}{} &  &  &  &  \\
			\multicolumn{1}{c|}{MobileNet-V2} & \textbf{64.9} & \textbf{64.9} & \textbf{58.7} & \multicolumn{1}{c|}{\textbf{67.1}} & 8.1 & 8.7 & 8.6 & \multicolumn{1}{c|}{19.1} & 59.6 & 61.6 & 46.3 & \multicolumn{1}{c|}{\textbf{23.4}} & 8.4 & \textbf{57.5} & 51.1 & 16.5 \\
			\multicolumn{1}{c|}{ResNet-50} & 68.7 & 69.6 & 61.0 & \multicolumn{1}{c|}{71.4} & 3.3 & 3.4 & 5.4 & \multicolumn{1}{c|}{31.2} & 67.3 & 68.5 & 62.9 & \multicolumn{1}{c|}{\textbf{29.1}} & \textbf{11.1} & 54.8 & 60.1 & 20.1 \\
			\multicolumn{1}{c|}{ResNet-101} & 68.0 & 68.2 & 62.4 & \multicolumn{1}{c|}{69.9} & 10.6 & 10.5 & 15.3 & \multicolumn{1}{c|}{40.3} & 66.8 & 69.2 & 63.7 & \multicolumn{1}{c|}{\textbf{39.0}} & 9.5 & \textbf{58.3} & 59.7 & 25.2 \\
			\multicolumn{1}{c|}{Xception-41} & 71.0 & 70.1 & 64.6 & \multicolumn{1}{c|}{71.5} & 12.6 & 11.4 & 18.1 & \multicolumn{1}{c|}{49.2} & 71.7 & 70.4 & 68.2 & \multicolumn{1}{c|}{\textbf{48.5}} & \textbf{18.0} & 60.6 & 62.4 & \textbf{24.0} \\
			\multicolumn{1}{c|}{Xception-71} & 72.3 & 70.4 & 65.8 & \multicolumn{1}{c|}{72.0} & 9.2 & 7.0 & 10.9 & \multicolumn{1}{c|}{43.0} & 73.3 & 71.3 & 67.6 & \multicolumn{1}{c|}{\textbf{43.5}} & 12.8 & \textbf{63.6} & 67.3 & 22.7 \\ \midrule
			\multicolumn{1}{c|}{\textbf{Painting-by-Numbers}} &  &  &  & \multicolumn{1}{c|}{} &  &  &  & \multicolumn{1}{c|}{} &  &  &  & \multicolumn{1}{c|}{} &  &  &  &  \\
			\multicolumn{1}{c|}{MobileNet-V2} & 63.7 & 62.6 & 57.4 & \multicolumn{1}{c|}{65.3} & \textbf{11.4} & \textbf{11.1} & \textbf{12.2} & \multicolumn{1}{c|}{\textbf{30.8}} & \textbf{66.3} & \textbf{66.0} & \textbf{57.9} & \multicolumn{1}{c|}{21.3} & \textbf{8.9} & 54.2 & \textbf{63.9} & \textbf{23.5} \\
			\multicolumn{1}{c|}{ResNet-50} & \textbf{70.0} & \textbf{70.1} & \textbf{67.9} & \multicolumn{1}{c|}{\textbf{71.8}} & \textbf{23.1} & \textbf{21.5} & \textbf{27.7} & \multicolumn{1}{c|}{\textbf{52.6}} & \textbf{71.8} & \textbf{71.8} & \textbf{69.2} & \multicolumn{1}{c|}{26.0} & 7.3 & \textbf{58.0} & \textbf{67.7} & \textbf{24.5} \\
			\multicolumn{1}{c|}{ResNet-101} & \textbf{70.8} & \textbf{71.1} & \textbf{65.6} & \multicolumn{1}{c|}{\textbf{72.5}} & \textbf{19.7} & \textbf{23.2} & \textbf{26.5} & \multicolumn{1}{c|}{\textbf{54.3}} & \textbf{73.3} & \textbf{72.9} & \textbf{71.3} & \multicolumn{1}{c|}{35.5} & \textbf{9.9} & 56.0 & \textbf{68.6} & \textbf{28.3} \\
			\multicolumn{1}{c|}{Xception-41} & \textbf{72.7} & \textbf{70.6} & \textbf{65.8} & \multicolumn{1}{c|}{\textbf{72.7}} & \textbf{30.0} & \textbf{25.1} & \textbf{35.3} & \multicolumn{1}{c|}{\textbf{58.3}} & \textbf{75.3} & \textbf{72.5} & \textbf{69.0} & \multicolumn{1}{c|}{25.8} & 12.4 & \textbf{63.7} & \textbf{69.6} & 22.6 \\
			\multicolumn{1}{c|}{Xception-71} & \textbf{74.1} & \textbf{73.8} & \textbf{68.2} & \multicolumn{1}{c|}{\textbf{75.3}} & \textbf{17.6} & \textbf{18.4} & \textbf{23.1} & \multicolumn{1}{c|}{\textbf{53.6}} & \textbf{77.0} & \textbf{74.7} & \textbf{72.7} & \multicolumn{1}{c|}{27.3} & \textbf{16.1} & 63.2 & \textbf{72.5} & \textbf{28.2} \\ \bottomrule
			
		\end{tabular}
	\end{adjustbox}
\end{table}    

\begin{table}[hbtp]
	\caption{
		Results on the Cityscapes dataset. 
		Each entry shows the mean IoU of several corrupted variants of the Cityscapes dataset for \textbf{severity level 3}.
		The higher mIoU of either the reference model or the respective model trained with Painting-by-Numbers is bold.
		Overall, we see many more bold numbers for our Painting-by-Numbers model    
	}
	\label{tab:sl3}
	\begin{adjustbox}{width=\textwidth}
		\begin{tabular}{@{}ccccccccccccccccc@{}}
			\toprule
			& \multicolumn{4}{c}{\textbf{Blur}} & \multicolumn{4}{c}{\textbf{Noise}} & \multicolumn{4}{c}{\textbf{Digital}} & \multicolumn{4}{c}{\textbf{Weather}} \\ \midrule
			\multicolumn{1}{c|}{Network} & Motion & Defocus & \begin{tabular}[c]{@{}c@{}}Frosted \\ Glass\end{tabular} & \multicolumn{1}{c|}{Gaussian} & Gaussian & Impulse & Shot & \multicolumn{1}{c|}{Speckle} & Brightness & Contrast & Saturate & \multicolumn{1}{c|}{JPEG} & Snow & Spatter & Fog & Frost \\ \midrule
			\multicolumn{1}{c|}{\textbf{Reference}} &  &  &  & \multicolumn{1}{c|}{} &  &  &  & \multicolumn{1}{c|}{} &  &  &  & \multicolumn{1}{c|}{} &  &  &  &  \\
			\multicolumn{1}{c|}{MobileNet-V2} & \textbf{55.4} & \textbf{50.4} & \textbf{39.7} & \multicolumn{1}{c|}{\textbf{54.9}} & 7.1 & 7.7 & 7.5 & \multicolumn{1}{c|}{9.7} & 50.3 & 51.9 & 51.6 & \multicolumn{1}{c|}{\textbf{18.8}} & 9.8 & \textbf{41.7} & 46.0 & 10.4 \\
			\multicolumn{1}{c|}{ResNet-50} & 59.6 & 59.5 & 37.1 & \multicolumn{1}{c|}{63.5} & 1.3 & 1.6 & 1.4 & \multicolumn{1}{c|}{9.2} & 59.8 & 61.0 & 64.9 & \multicolumn{1}{c|}{\textbf{22.4}} & \textbf{10.7} & 35.0 & 56.7 & 12.1 \\
			\multicolumn{1}{c|}{ResNet-101} & 62.4 & 60.9 & 41.3 & \multicolumn{1}{c|}{64.1} & 5.0 & 6.7 & 8.5 & \multicolumn{1}{c|}{18.9} & 58.8 & 63.9 & 62.3 & \multicolumn{1}{c|}{\textbf{32.8}} & 9.9 & 40.6 & 55.3 & 16.5 \\
			\multicolumn{1}{c|}{Xception-41} & 64.5 & \textbf{61.0} & \textbf{47.0} & \multicolumn{1}{c|}{\textbf{64.5}} & 4.6 & 6.8 & 6.6 & \multicolumn{1}{c|}{22.7} & 65.8 & 65.5 & 70.6 & \multicolumn{1}{c|}{\textbf{41.6}} & \textbf{15.7} & 42.0 & 58.5 & \textbf{16.1} \\
			\multicolumn{1}{c|}{Xception-71} & 65.5 & \textbf{63.5} & \textbf{50.8} & \multicolumn{1}{c|}{65.6} & 4.3 & 5.6 & 5.1 & \multicolumn{1}{c|}{14.5} & 69.7 & 65.9 & 73.3 & \multicolumn{1}{c|}{\textbf{38.2}} & 13.4 & 42.2 & 63.4 & 13.1 \\ \midrule
			\multicolumn{1}{c|}{\textbf{Painting-by-Numbers}} &  &  &  & \multicolumn{1}{c|}{} &  &  &  & \multicolumn{1}{c|}{} &  &  &  & \multicolumn{1}{c|}{} &  &  &  &  \\
			\multicolumn{1}{c|}{MobileNet-V2} & 52.6 & 41.7 & 36.5 & \multicolumn{1}{c|}{48.6} & \textbf{8.6} & \textbf{9.1} & \textbf{9.1} & \multicolumn{1}{c|}{\textbf{15.7}} & \textbf{62.8} & \textbf{59.9} & \textbf{66.8} & \multicolumn{1}{c|}{17.0} & \textbf{10.8} & 41.1 & \textbf{60.7} & \textbf{17.3} \\
			\multicolumn{1}{c|}{ResNet-50} & \textbf{63.2} & \textbf{61.5} & \textbf{43.9} & \multicolumn{1}{c|}{\textbf{65.7}} & \textbf{14.4} & \textbf{16.2} & \textbf{18.3} & \multicolumn{1}{c|}{\textbf{33.2}} & \textbf{68.7} & \textbf{68.9} & \textbf{71.2} & \multicolumn{1}{c|}{20.8} & 8.7 & \textbf{40.6} & \textbf{64.8} & \textbf{16.3} \\
			\multicolumn{1}{c|}{ResNet-101} & \textbf{64.0} & \textbf{64.4} & \textbf{45.7} & \multicolumn{1}{c|}{\textbf{67.5}} & \textbf{8.7} & \textbf{15.1} & \textbf{11.9} & \multicolumn{1}{c|}{\textbf{29.9}} & \textbf{70.6} & \textbf{69.6} & \textbf{74.4} & \multicolumn{1}{c|}{28.6} & \textbf{10.7} & \textbf{40.9} & \textbf{65.8} & \textbf{20.1} \\
			\multicolumn{1}{c|}{Xception-41} & \textbf{67.1} & 56.7 & 44.4 & \multicolumn{1}{c|}{61.5} & \textbf{16.5} & \textbf{17.6} & \textbf{21.2} & \multicolumn{1}{c|}{\textbf{39.3}} & \textbf{73.2} & \textbf{67.6} & \textbf{76.3} & \multicolumn{1}{c|}{19.4} & 13.5 & \textbf{48.1} & \textbf{66.1} & 13.9 \\
			\multicolumn{1}{c|}{Xception-71} & \textbf{67.9} & 63.4 & 44.4 & \multicolumn{1}{c|}{\textbf{67.1}} & \textbf{8.8} & \textbf{10.8} & \textbf{11.1} & \multicolumn{1}{c|}{\textbf{29.0}} & \textbf{75.3} & \textbf{71.2} & \textbf{77.6} & \multicolumn{1}{c|}{20.8} & \textbf{14.3} & \textbf{48.3} & \textbf{69.7} & \textbf{19.3} \\ \bottomrule

		\end{tabular}
	\end{adjustbox}
\end{table}

\begin{table}[hbtp]
	\caption{
		Results on the Cityscapes dataset. 
		Each entry shows the mean IoU of several corrupted variants of the Cityscapes dataset for \textbf{severity level 4}.
		The higher mIoU of either the reference model or the respective model trained with Painting-by-Numbers is bold.
		Overall, we see many more bold numbers for our Painting-by-Numbers model    
	}
	\label{tab:sl4}
	\begin{adjustbox}{width=\textwidth}
		\begin{tabular}{@{}ccccccccccccccccc@{}}
			\toprule
			& \multicolumn{4}{c}{\textbf{Blur}} & \multicolumn{4}{c}{\textbf{Noise}} & \multicolumn{4}{c}{\textbf{Digital}} & \multicolumn{4}{c}{\textbf{Weather}} \\ \midrule
			\multicolumn{1}{c|}{Network} & Motion & Defocus & \begin{tabular}[c]{@{}c@{}}Frosted \\ Glass\end{tabular} & \multicolumn{1}{c|}{Gaussian} & Gaussian & Impulse & Shot & \multicolumn{1}{c|}{Speckle} & Brightness & Contrast & Saturate & \multicolumn{1}{c|}{JPEG} & Snow & Spatter & Fog & Frost \\ \midrule
			\multicolumn{1}{c|}{\textbf{Reference}} &  &  &  & \multicolumn{1}{c|}{} &  &  &  & \multicolumn{1}{c|}{} &  &  &  & \multicolumn{1}{c|}{} &  &  &  &  \\
			\multicolumn{1}{c|}{MobileNet-V2} & \textbf{39.6} & \textbf{31.9} & \textbf{33.1} & \multicolumn{1}{c|}{\textbf{34.9}} & 6.1 & 6.6 & 5.8 & \multicolumn{1}{c|}{8.3} & 41.2 & 29.1 & 5.5 & \multicolumn{1}{c|}{\textbf{14.0}} & 7.6 & 32.2 & 46.1 & 9.6 \\
			\multicolumn{1}{c|}{ResNet-50} & 45.3 & \textbf{43.8} & 31.7 & \multicolumn{1}{c|}{49.1} & 1.2 & 1.2 & 1.2 & \multicolumn{1}{c|}{3.7} & 53.0 & 41.7 & 8.7 & \multicolumn{1}{c|}{\textbf{12.5}} & \textbf{8.8} & 23.4 & 54.6 & 10.8 \\
			\multicolumn{1}{c|}{ResNet-101} & \textbf{49.4} & \textbf{45.5} & 36.0 & \multicolumn{1}{c|}{51.4} & 3.6 & 4.4 & \textbf{5.3} & \multicolumn{1}{c|}{12.9} & 50.4 & 49.9 & 8.4 & \multicolumn{1}{c|}{\textbf{22.4}} & 7.8 & 32.6 & 54.3 & 15.5 \\
			\multicolumn{1}{c|}{Xception-41} & 51.4 & \textbf{42.1} & \textbf{42.0} & \multicolumn{1}{c|}{\textbf{44.8}} & 2.8 & 5.0 & 2.5 & \multicolumn{1}{c|}{12.6} & 58.7 & \textbf{50.7} & 21.6 & \multicolumn{1}{c|}{\textbf{27.0}} & \textbf{10.9} & 35.6 & 56.4 & \textbf{14.9} \\
			\multicolumn{1}{c|}{Xception-71} & \textbf{53.5} & \textbf{50.6} & \textbf{44.5} & \multicolumn{1}{c|}{\textbf{53.8}} & 3.0 & 4.0 & 2.7 & \multicolumn{1}{c|}{7.6} & 63.1 & 49.0 & 10.4 & \multicolumn{1}{c|}{\textbf{25.3}} & \textbf{9.8} & 38.8 & 62.4 & 12.1 \\ \midrule
			\multicolumn{1}{c|}{\textbf{Painting-by-Numbers}} &  &  &  & \multicolumn{1}{c|}{} &  &  &  & \multicolumn{1}{c|}{} &  &  &  & \multicolumn{1}{c|}{} &  &  &  &  \\
			\multicolumn{1}{c|}{MobileNet-V2} & 35.1 & 22.6 & 27.6 & \multicolumn{1}{c|}{23.8} & \textbf{7.0} & \textbf{7.1} & \textbf{7.1} & \multicolumn{1}{c|}{\textbf{11.1}} & \textbf{59.0} & \textbf{39.2} & \textbf{38.8} & \multicolumn{1}{c|}{10.8} & \textbf{8.2} & \textbf{41.6} & \textbf{54.1} & \textbf{15.6} \\
			\multicolumn{1}{c|}{ResNet-50} & \textbf{46.5} & 41.2 & \textbf{36.4} & \multicolumn{1}{c|}{\textbf{50.6}} & \textbf{8.9} & \textbf{9.7} & \textbf{9.8} & \multicolumn{1}{c|}{\textbf{25.7}} & \textbf{66.0} & \textbf{60.0} & \textbf{49.4} & \multicolumn{1}{c|}{12.7} & 7.5 & \textbf{37.9} & \textbf{59.5} & \textbf{14.8} \\
			\multicolumn{1}{c|}{ResNet-101} & 49.2 & 44.7 & \textbf{40.1} & \multicolumn{1}{c|}{\textbf{53.1}} & \textbf{5.1} & \textbf{6.3} & 5.1 & \multicolumn{1}{c|}{\textbf{19.2}} & \textbf{67.7} & \textbf{62.3} & \textbf{58.9} & \multicolumn{1}{c|}{17.6} & \textbf{8.7} & \textbf{36.4} & \textbf{59.5} & \textbf{17.8} \\
			\multicolumn{1}{c|}{Xception-41} & \textbf{51.5} & 31.4 & 36.4 & \multicolumn{1}{c|}{27.7} & \textbf{9.7} & \textbf{9.7} & \textbf{11.0} & \multicolumn{1}{c|}{\textbf{29.5}} & \textbf{70.9} & 50.0 & \textbf{63.8} & \multicolumn{1}{c|}{10.7} & 8.9 & \textbf{49.1} & \textbf{59.4} & 11.9 \\
			\multicolumn{1}{c|}{Xception-71} & 52.8 & 39.4 & 35.5 & \multicolumn{1}{c|}{37.7} & \textbf{5.1} & \textbf{5.3} & \textbf{5.9} & \multicolumn{1}{c|}{\textbf{18.4}} & \textbf{73.4} & \textbf{57.1} & \textbf{64.4} & \multicolumn{1}{c|}{12.1} & 9.6 & \textbf{57.3} & \textbf{62.9} & \textbf{16.3} \\ \bottomrule

		\end{tabular}
	\end{adjustbox}
\end{table}

\begin{table}[hbtp]
	\caption{
		Results on the Cityscapes dataset. 
		Each entry shows the mean IoU of several corrupted variants of the Cityscapes dataset for \textbf{severity level 5}.
		The higher mIoU of either the reference model or the respective model trained with Painting-by-Numbers is bold
	}
	\label{tab:sl5}
	\begin{adjustbox}{width=\textwidth}
		\begin{tabular}{@{}ccccccccccccccccc@{}}
			\toprule
			& \multicolumn{4}{c}{\textbf{Blur}} & \multicolumn{4}{c}{\textbf{Noise}} & \multicolumn{4}{c}{\textbf{Digital}} & \multicolumn{4}{c}{\textbf{Weather}} \\ \midrule
			\multicolumn{1}{c|}{Network} & Motion & Defocus & \begin{tabular}[c]{@{}c@{}}Frosted \\ Glass\end{tabular} & \multicolumn{1}{c|}{Gaussian} & Gaussian & Impulse & Shot & \multicolumn{1}{c|}{Speckle} & Brightness & Contrast & Saturate & \multicolumn{1}{c|}{JPEG} & Snow & Spatter & Fog & Frost \\ \midrule
			\multicolumn{1}{c|}{\textbf{Reference}} &  &  &  & \multicolumn{1}{c|}{} &  &  &  & \multicolumn{1}{c|}{} &  &  &  & \multicolumn{1}{c|}{} &  &  &  &  \\
			\multicolumn{1}{c|}{MobileNet-V2} & \textbf{33.3} & \textbf{19.2} & \textbf{26.4} & \multicolumn{1}{c|}{\textbf{12.2}} & 5.2 & 5.7 & 5.2 & \multicolumn{1}{c|}{7.0} & 32.3 & 10.0 & 4.2 & \multicolumn{1}{c|}{\textbf{10.8}} & 5.7 & 13.5 & \textbf{39.4} & 7.8 \\
			\multicolumn{1}{c|}{ResNet-50} & \textbf{39.1} & \textbf{31.0} & 27.4 & \multicolumn{1}{c|}{\textbf{23.2}} & 1.4 & 1.3 & 1.3 & \multicolumn{1}{c|}{1.4} & 45.3 & 20.3 & 5.4 & \multicolumn{1}{c|}{\textbf{8.7}} & \textbf{8.4} & 9.1 & \textbf{45.8} & 8.3 \\
			\multicolumn{1}{c|}{ResNet-101} & \textbf{43.2} & \textbf{31.0} & 30.1 & \multicolumn{1}{c|}{\textbf{23.2}} & \textbf{3.8} & \textbf{4.7} & \textbf{4.8} & \multicolumn{1}{c|}{9.6} & 39.8 & 29.0 & 6.0 & \multicolumn{1}{c|}{\textbf{16.0}} & 7.7 & \textbf{20.1} & \textbf{45.3} & 12.0 \\
			\multicolumn{1}{c|}{Xception-41} & \textbf{46.3} & \textbf{26.6} & \textbf{33.5} & \multicolumn{1}{c|}{\textbf{12.9}} & 1.8 & 3.6 & 2.7 & \multicolumn{1}{c|}{6.5} & 48.5 & \textbf{30.0} & 11.4 & \multicolumn{1}{c|}{\textbf{17.8}} & \textbf{8.4} & 18.3 & \textbf{47.9} & \textbf{11.7} \\
			\multicolumn{1}{c|}{Xception-71} & \textbf{47.1} & \textbf{35.4} & \textbf{31.3} & \multicolumn{1}{c|}{\textbf{21.8}} & 2.7 & \textbf{3.4} & 3.1 & \multicolumn{1}{c|}{4.6} & 52.4 & 26.0 & 6.0 & \multicolumn{1}{c|}{\textbf{16.8}} & 7.7 & 18.6 & 56.1 & 8.2 \\ \midrule
			\multicolumn{1}{c|}{\textbf{Painting-by-Numbers}} &  &  &  & \multicolumn{1}{c|}{} &  &  &  & \multicolumn{1}{c|}{} &  &  &  & \multicolumn{1}{c|}{} &  &  &  &  \\
			\multicolumn{1}{c|}{MobileNet-V2} & 28.2 & 13.3 & 17.9 & \multicolumn{1}{c|}{7.2} & \textbf{5.3} & \textbf{5.8} & \textbf{6.3} & \multicolumn{1}{c|}{\textbf{8.4}} & \textbf{54.5} & \textbf{20.3} & \textbf{28.6} & \multicolumn{1}{c|}{9.0} & \textbf{6.9} & \textbf{26.8} & 37.7 & \textbf{12.5} \\
			\multicolumn{1}{c|}{ResNet-50} & 38.5 & 22.2 & \textbf{30.8} & \multicolumn{1}{c|}{12.9} & \textbf{5.8} & \textbf{6.5} & \textbf{6.5} & \multicolumn{1}{c|}{\textbf{19.5}} & \textbf{63.0} & \textbf{46.9} & \textbf{41.3} & \multicolumn{1}{c|}{8.9} & 8.1 & \textbf{18.6} & 43.6 & \textbf{11.3} \\
			\multicolumn{1}{c|}{ResNet-101} & 40.8 & 27.5 & \textbf{34.5} & \multicolumn{1}{c|}{17.2} & 3.7 & 4.0 & 3.9 & \multicolumn{1}{c|}{\textbf{10.4}} & \textbf{63.5} & \textbf{50.5} & \textbf{52.1} & \multicolumn{1}{c|}{12.2} & \textbf{8.2} & 19.4 & 43.3 & \textbf{14.2} \\
			\multicolumn{1}{c|}{Xception-41} & 44.4 & 14.2 & 25.4 & \multicolumn{1}{c|}{3.5} & \textbf{6.0} & \textbf{6.2} & \textbf{7.7} & \multicolumn{1}{c|}{\textbf{20.9}} & \textbf{67.5} & 29.9 & \textbf{54.5} & \multicolumn{1}{c|}{7.0} & 7.9 & \textbf{28.7} & 41.6 & 9.2 \\
			\multicolumn{1}{c|}{Xception-71} & 45.0 & 16.1 & 21.1 & \multicolumn{1}{c|}{4.3} & \textbf{3.3} & 3.3 & \textbf{4.1} & \multicolumn{1}{c|}{\textbf{11.5}} & \textbf{70.7} & \textbf{40.4} & \textbf{54.5} & \multicolumn{1}{c|}{7.9} & \textbf{9.4} & \textbf{40.0} & \textbf{47.2} & \textbf{12.0} \\ \bottomrule

		\end{tabular}
	\end{adjustbox}
\end{table}

\section{Understanding Painting-by-Numbers}
\label{sec:secD}    

\begin{figure}[h]
	\centering
	
	\subfloat[Original image]{%
      \includegraphics[width=0.48\linewidth]{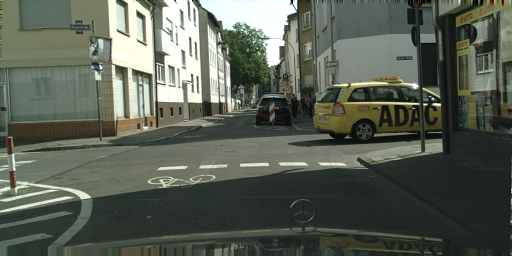}%
    }\hfil
	\subfloat[Silhouette of \textit{car}]{%
      \includegraphics[width=0.48\linewidth]{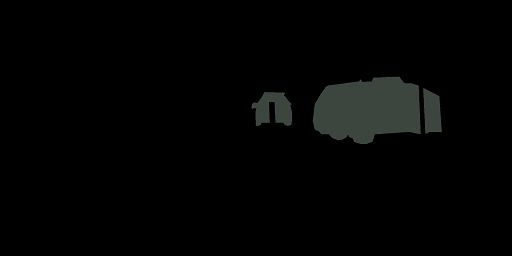}%
    }\hfil
	
	\caption{
		An original image of the Cityscapes validation image (a) and the corresponding silhouette for class \textit{car}.
		Similar to the main paper, the class texture is replaced by RGB-mean, but in addition, we also black the remaining classes.
		A network segmenting such an image is not able to rely on context or background information but has to utilize shape-based cues for the segmentation
	}
	\label{fig:silhouette}    
\end{figure}

In section 4.3 in the main paper, we conducted several experiments to verify an increased shape bias of a network which is trained with Painting-by-Numbers.
In one of these experiments, we removed the texture of a class and replaced it by the data-wide RGB-mean of the respective class. 
We provide supplementary results of an extended version of this experiment, where we further black the remaining classes, as illustrated in Fig.~\ref{fig:silhouette}.
The sensitivity score for every class is listed in Table~\ref{tab:silhouette_r50}.
As the results in the main paper indicate, the reference model is not able to segment any classes except \textit{building} and \textit{sky}, since it may rely mostly on the mean color of these classes.
Our model, on the other hand, achieves considerably higher sensitivity scores for category ``things'' such as \textit{road}, \textit{traffic light}, \textit{traffic sign}, \textit{pole}, \textit{person}, and \textit{car}.
%Fig.~\ref{fig:sil_qualresults} shows qualitative results for classes \textit{person} and \textit{car}.

\begin{table}[hbtp]
	\centering
	\caption{
		Sensitivity score per class for several corrupted variants on class-level of the Cityscapes dataset, using ResNet-50 as the network backbone.
		In these experiments, solely the silhouette of a class is present, forcing the network to rely mostly on the class-shape for the segmentation.
		Whereas the reference model is, in most cases, not able to segment the images, our model performs superior for many classes of category ``things'' with a distinct shape.
		The higher sensitivity score of a network backbone of either the reference (top) or our model (bottom) is bold.
		Overall, we see many more bold numbers for our Painting-by-Numbers model
	}
	\label{tab:silhouette_r50}
	\begin{adjustbox}{width=\textwidth}
		\begin{tabular}{@{}cccccccccccccccccccc@{}}    & \rotatebox[origin=l]{90}{road} & \rotatebox[origin=l]{90}{sidewalk} & \rotatebox[origin=l]{90}{building} & \rotatebox[origin=l]{90}{wall} & \rotatebox[origin=l]{90}{fence} & \rotatebox[origin=l]{90}{pole} & \rotatebox[origin=l]{90}{traffic light} & \rotatebox[origin=l]{90}{traffic sign} & \rotatebox[origin=l]{90}{vegetation} & \rotatebox[origin=l]{90}{terrain} & \rotatebox[origin=l]{90}{sky} & \rotatebox[origin=l]{90}{person} & \rotatebox[origin=l]{90}{rider} & \rotatebox[origin=l]{90}{car} & \rotatebox[origin=l]{90}{truck} & \rotatebox[origin=l]{90}{bus} & \rotatebox[origin=l]{90}{train} & \rotatebox[origin=l]{90}{motorcycle} & \rotatebox[origin=l]{90}{bicycle}  \\ \midrule
			\textbf{Reference (ResNet-50)} & 2.1 & 0.0 & \textbf{93.3} & 0.4 & 0.0 & 1.0 & 0.0 & 0.1 & 0.0 & 0.0 & \textbf{67.4} & 0.8 & 0.0 & 0.0 & 0.1 & \textbf{5.6} & \textbf{2.2} & 0.0 & 0.0 \\
			\textbf{Our (ResNet-50)} & \textbf{99.1} & \textbf{58.4} & 18.1 & \textbf{5.9} & \textbf{2.1} & \textbf{86.2} & \textbf{31.8} & \textbf{26.6} & \textbf{4.1} & \textbf{8.2} & 61.9 & \textbf{79.5} & \textbf{1.9} & \textbf{77.4} & \textbf{0.1} & 0.0 & 0.0 & \textbf{3.8} & \textbf{12.1} \\ \bottomrule

		\end{tabular}
	\end{adjustbox}
\end{table}

The final authenticated publication is available online at \url{https://doi.org/}

\end{document}